\pdfoutput=1 

\documentclass[pdflatex,sn-mathphys-num]{sn-jnl}%

\usepackage{graphicx}%
\usepackage{multirow}%
\usepackage{amsmath,amssymb,amsfonts}%
\usepackage{amsthm}%
\usepackage{mathrsfs}%
\usepackage[title]{appendix}%
\usepackage{xcolor}%
\usepackage{textcomp}%
\usepackage{manyfoot}%
\usepackage{booktabs}%
\usepackage{algpseudocode}%
\usepackage{listings}%

\theoremstyle{thmstyleone}%

\theoremstyle{thmstyletwo}%

\theoremstyle{thmstylethree}%

\usepackage{grffile}
\usepackage[utf8]{inputenc} %
\usepackage[T1]{fontenc}    %
\usepackage{url}            %
\usepackage{booktabs}       %
\usepackage{amsfonts}       %
\usepackage{amssymb}
\usepackage{nicefrac}       %
\usepackage{microtype}      %
\usepackage{xcolor}         %
\usepackage{mdframed}
\usepackage{caption}
\usepackage{booktabs}  
\usepackage{anyfontsize}

\usepackage[linesnumbered,ruled,vlined]{algorithm2e}
\SetKwInput{KwInput}{Input}                
\SetKwInput{KwOutput}{Output}              

\usepackage{amsmath,amsfonts,bm}

\def\eqref#1{equation~\ref{#1}}

\def\1{\bm{1}}

\DeclareMathAlphabet{\mathsfit}{\encodingdefault}{\sfdefault}{m}{sl}
\SetMathAlphabet{\mathsfit}{bold}{\encodingdefault}{\sfdefault}{bx}{n}

\def\gG{{\mathcal{G}}}

\usepackage[acronym,nohypertypes={acronym,notation}]{glossaries}
\newacronym{gns}{GNS}{Graph Network Simulator}
\newacronym{mdp}{MDP}{Markov Decision Process}
\newacronym{swarmdp}{SwarMDP}{Swarm Markov Decision Process}
\newacronym{asmdp}{ASMDP}{Adaptive Swarm Markov Decision Process}
\newacronym{mgn}{MGN}{MeshGraphNet}
\newacronym{mpc}{MPC}{Model Predictive Control}
\newacronym{mbrl}{MBRL}{Model-Based Reinforcement Learning}
\newacronym{gnn}{GNN}{Graph Neural Network}
\newacronym{sofa}{SOFA}{Simulation Open Framework Architecture}
\newacronym{mpn}{MPN}{Message Passing Network}
\newacronym{mlp}{MLP}{Multilayer Perceptron}
\newacronym{cnn}{CNN}{Convoluational Neural Network}
\newacronym{mse}{MSE}{Mean Squared Error}
\newacronym{iou}{IoU}{Intersection over Union}
\newacronym{ggns}{GGNS}{Grounding Graph Network Simulator}
\newacronym{rl}{RL}{Reinforcement Learning}
\newacronym{amr}{AMR}{Adaptive Mesh Refinement}
\newacronym{fem}{FEM}{Finite Element Method}
\newacronym{pde}{PDE}{Partial Differential Equation}
\newacronym{ppo}{PPO}{Proximal Policy Optimization}
\newacronym{iqm}{IQM}{Interquartile Mean}
\newacronym{dqn}{DQN}{Deep Q-Network}
\newacronym{gat}{GAT}{Graph Attention Network}
\newacronym{vdgn}{\textit{VDGN}}{Value Decomposition Graph Networks}
\newacronym{zz}{\textit{ZZ Error}}{Zienkiewicz-Zhu Error Estimator}
\newacronym{asmr}{\textit{ASMR}}{Adaptive Swarm Mesh Refinement}

\newacronym{method}{ASMR++}{Adaptive Swarm Mesh Refinement++}
\newacronym{rlamr}{RL-AMR}{RL-AMR}

\usepackage{xcolor} 
\newcommand{\rebuttal}[1]{#1}
\usepackage{pgfplots}
\pgfplotsset{compat=1.12,
            label style={font=\scriptsize},
            tick label style={font=\tiny},  }
\usetikzlibrary{pgfplots.groupplots}

\newcommand{\tikzsetnextfilename}[1]{}

\usepackage{adjustbox}

\begin{document}

\title[Adaptive Swarm Mesh Refinement using Deep Reinforcement Learning with Local Rewards]{Adaptive Swarm Mesh Refinement using Deep Reinforcement Learning with Local Rewards}

\author*[1]{\fnm{Niklas} \sur{Freymuth}}\email{freymuth@kit.edu}

\author[1]{\fnm{Philipp} \sur{Dahlinger}}

\author[2]{\fnm{Tobias} \sur{Würth}}

\author[1]{\fnm{Simon} \sur{Reisch}}

\author[2]{\fnm{Luise} \sur{Kärger}}

\author[1]{\fnm{Gerhard} \sur{Neumann}}

\affil*[1]{\orgname{Karlsruhe Institute of Technology (KIT)}, \orgdiv{Autonomous Learning Robots}, \orgaddress{\city{Karlsruhe}, \country{Germany}}}

\affil[2]{\orgname{Karlsruhe Institute of Technology (KIT)}, \orgdiv{Institute of Vehicle System Technology}, \orgaddress{\city{Karlsruhe}, \country{Germany}}}

\abstract{
Simulating physical systems is essential in engineering, but analytical solutions are limited to straightforward problems. 
Consequently, numerical methods like the Finite Element Method (FEM) are widely used. 
However, the FEM becomes computationally expensive as problem complexity and accuracy demands increase.
Adaptive Mesh Refinement (AMR) improves the FEM by dynamically \rebuttal{placing} mesh elements on the domain, balancing computational speed and accuracy.
Classical AMR depends on heuristics or expensive error estimators, \rebuttal{which may lead to suboptimal performance for} complex simulations.
While AMR methods \rebuttal{based on machine learning} are promising, they currently only scale to simple problems. 
In this work, we formulate AMR as a system of collaborating, homogeneous agents that iteratively split into multiple new agents.
This agent-wise perspective enables a spatial reward formulation focused on reducing the maximum mesh element error. 
Our approach, Adaptive Swarm Mesh Refinement\rebuttal{++} (ASMR\rebuttal{++}), offers efficient, stable optimization and generates highly adaptive meshes at user-defined resolution \rebuttal{at inference time}.
Extensive experiments demonstrate that ASMR \rebuttal{outperforms} heuristic approaches and learned baselines, matching the performance of expensive error-based oracle AMR strategies.
ASMR additionally generalizes to different domains during inference, and produces meshes that simulate up to 2 orders of magnitude faster than uniform refinements in more demanding settings.
}


\maketitle

\glsresetall
\glsunset{rlamr}  %
\glsunset{asmr}
\glsunset{fem}
\glsunset{method}
\section{Introduction}
\setcounter{footnote}{0}  %

\rebuttal{The numerical simulation} of fundamental \rebuttal{physical} principles like mass, momentum, and energy conservation, often expressed through complex~\glspl{pde} \rebuttal{, is a cornerstone of modern engineering}. 
Since analytical solutions of such~\glspl{pde} are limited to simple cases, numerical approximations, particularly the~\gls{fem} are commonly employed~\citep{brenner2008mathematical, reddy2019introduction, anderson2021mfem}.
\rebuttal{The~\gls{fem} provides a framework to find approximate solutions by transforming the continuous~\glspl{pde} into a discrete system of equations.}
\rebuttal{It} partitions the continuous problem domain into a mesh consisting of smaller, finite elements, allowing for an efficient numerical solution whose accuracy depends on the number used elements. 
However, as the physics becomes more complex, accurate simulations become significantly more expensive~\citep{wanner1996solving, zimmerling2022optimisation, brandstetter2022message}.

\rebuttal{
To bypass the high computational costs of the~\gls{fem}, an alternative line of research focuses on "learned simulators" that approximate physical solutions directly from data.}
Early work on these learned simulators uses simple feed-forward~\citep{um2018liquid, zimmerling2019comp} or convolutional neural networks~\citep{guo2016convolutional, zimmerling2019aipconf, zimmerling2022optimisation}\rebuttal{.}
\rebuttal{Due to the used network architectures, this line of work is} ill-suited for \rebuttal{irregular, unstructured} mesh-based representations in the~\gls{fem}.
Here,~\glspl{gns}~\citep{sanchezgonzalez2020learning, pfaff2020learning} have emerged as an alternative architecture acting on a graph encoding of the simulation state.
As an alternative to data-driven learning approach, physics-informed neural networks~\citep{raissi2019physics, wuerth2023matdes} directly optimize a neural network to predict simulations that satisfy the governing equations of a physical system.
While usually mesh-free, recent extensions combine these approaches with~\glspl{gns} and meshes to facilitate generalization to new domains during inference~\citep{wuerth2024cmame}.
All of these learned simulators share a common goal of using recent advances in deep learning to speed up the simulation of complex physical systems.
Yet, they do so by directly approximating the simulation, causing any prediction error of the learned model to directly affect the simulated quantities.

\rebuttal{Instead of replacing the solver entirely,} a more robust and risk-averse approach to speed up the~\gls{fem} is \glsfirst{amr}, which dynamically allocates more mesh elements to regions of high solution variability, striking a favorable balance between computational efficiency and accuracy~\citep{plewa2005adaptive,huang2010adaptive, fidkowski2011review}.
As~\gls{amr} allows the creation of a simulation-specific mesh, it has become increasingly important for complex problems across domains like fluid dynamics~\citep{berger1989local, baker1997mesh, zhang2020meshingnet, wallwork2022e2n}, structural mechanics~\citep{ortiz1991adaptive, stein2007adaptive, gibert20193d}, and astrophysics~\citep{cunningham2009simulating, bryan2014enzo, guillet2019high}.
Yet, general-purpose~\gls{amr} methods often rely on either simple heuristics~\citep{zienkiewicz1992superconvergent} or computationally intensive error estimation techniques, such as goal-oriented adaptive finite element methods~\citep{becker2023goal} and dual-weighted residual methods~\citep{bangerth2013adaptive}. 
While these approaches can achieve improved accuracy for specific applications, they are generally \rebuttal{limited by high computational overhead for high-fidelity estimates or a lack of robustness across different}~\citep{mukherjee1996adaptive, kita2001error, yano2012optimization, cerveny2019nonconforming, wallwork2021mesh}.
\rebuttal{The above constraints complicate} their effective use in practical scenarios.

Several recent methods apply supervised learning to improve existing or devise new \gls{amr} strategies.
Here, methods usually predict intermediary metrics for specific aspects of~\gls{amr}, such as predicting an error or refinement marking per mesh element for a subsequent refinement step~\citep{zhang2020meshingnet, bohn2021recurrent, roth2022neural, wallwork2022e2n, sluzalec2023quasi}, or predicting local mesh densities over the considered domain~\citep{huang2021machine}.
These supervised approaches typically propose a greedy and often local next-step refinement focused on minimizing the error in the next refinement step.
However, this short-term focus may come at the cost of long-term optimization, \rebuttal{particularly in problems where local refinements influence error propagation globally or where the simulation involves multiple sequential steps}~\citep{yang2023multi}.

Finally, an emerging body of work employs~\gls{rl}~\citep{sutton2018reinforcement} to \rebuttal{formulate}~\gls{amr}, and specifically \textit{h-refinement} \citep{arnold2000locally, stevenson2008completion}, i.e., element sub-division and combination, as a sequential decision-making process.
Multiple strategies for \glsfirst{rlamr} have been developed, all of which are motivated by the potential of \gls{rl} to optimize non-differentiable, long-term rewards that are formulated to correspond to an error measure on the mesh. 
When considering h-refinement, \rebuttal{the discrete mesh structure evolves at each step, resulting in a dynamic state space that requires the \gls{rl} policy to handle a varying number of inputs and outputs.}
Some work circumvents this issue by learning a global, per-mesh quantity~\citep{gillette2022learning, pan2023reinforcement}, acting on local mesh elements~\citep{huergo2024reinforcement, dzanic2024dynamo}, or sampling a fixed number of mesh edges to refine per step~\citep{wu2023learning}.
More general methods that consider the full mesh either do so by iteratively selecting an element to refine~\citep{yang2023reinforcement} or training on individual element's refinements and then inferring on the full mesh~\citep{foucart2023deep}.
Another recent method uses a value decomposition network~\citep{sunehag2017value} to learn the credit assignment of the individual elements on the mesh quality by decomposing a shared Q-function~\citep{yang2023multi}.
All of these approaches only scale to simple problems, either due to an expensive inference process~\citep{yang2023reinforcement}, misaligned objectives and high variance in the state transitions~\citep{foucart2023deep}, or noisy reward signals during training~\citep{yang2023multi}.

In this work, we instead formulate \gls{amr} via h-refinement as a Swarm \gls{rl}~\citep{vsovsic2017inverse, huttenrauch2019deep} problem, extending existing frameworks to a shared observation space and per-agent, spatial rewards.
Crucially, we allow for agents to split into new agents over time to model element sub-division in the refinement process, mapping a per-agent reward signal quantifying the reduction in simulation error over refinement steps to assign credits for individual agents throughout an episode.
We use a~\glspl{mpn}~\citep{sanchezgonzalez2020learning}, a type of \gls{gnn}~\citep{scarselli2009the, bronstein2021geometric}, for our policy due to their effectiveness in learning physical simulations~\citep{pfaff2020learning, brandstetter2022message}.
Our method, Adaptive Swarm Mesh Refinement (ASMR), consistently produces highly efficient mesh refinements with thousands of elements and can be applied to arbitrary \glspl{pde}. 
ASMR can further generalize to previously unseen domains, system dynamics and material parameters when trained appropriately,, allowing for robust and adaptive solutions across a wide range of complex scenarios.

\begin{figure}
    \centering
	\includegraphics[width=\textwidth]{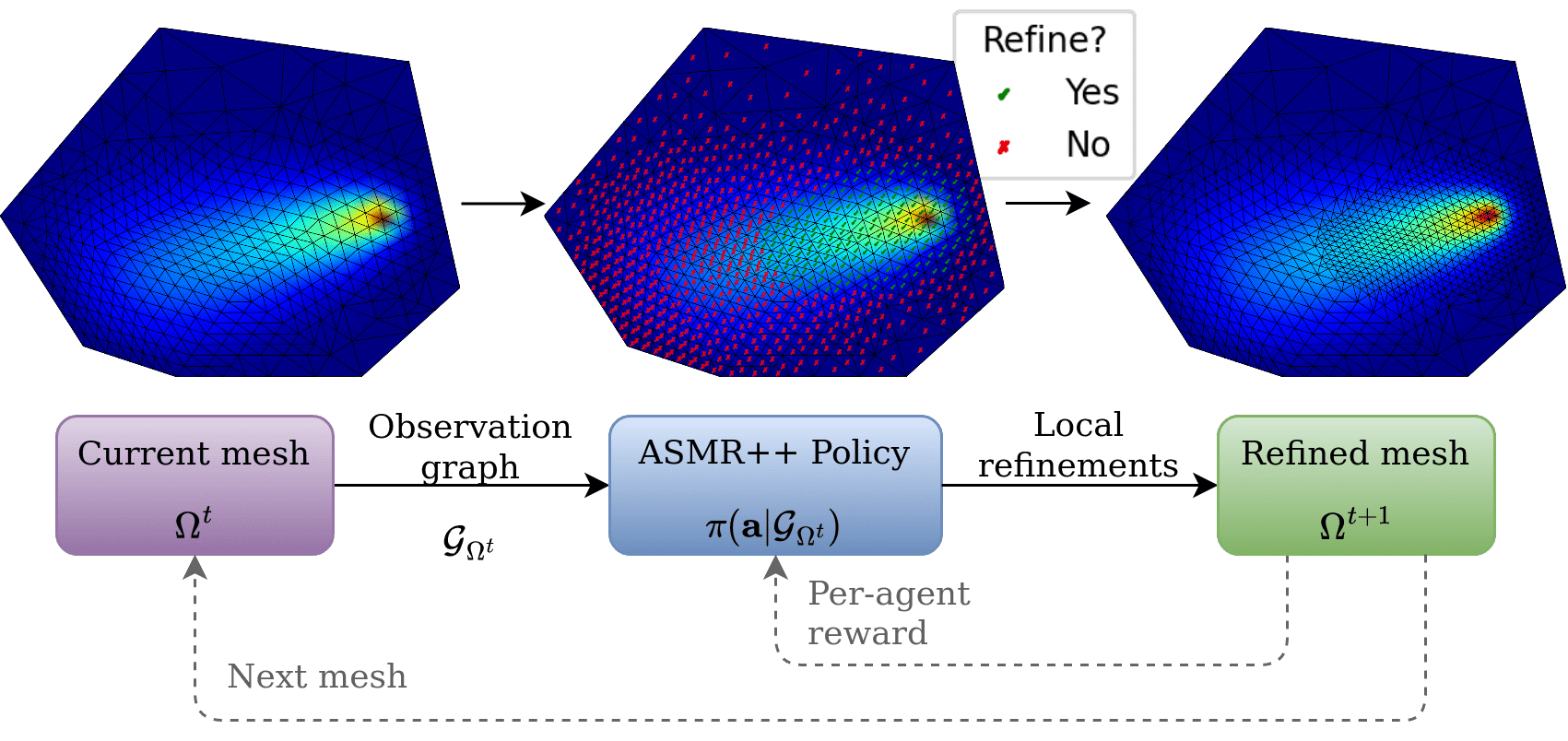}
	\caption{
        A schematic~\gls{method} refinement step. Given a current mesh, an observation graph encodes the elements as nodes and their neighborhood relationship as edges.
        A policy consumes this graph to decide on per-element markings that are given to a remesher, which subsequently produces a refined mesh.
        Based on the quality of this finer mesh, per-agent rewards are calculated.
        The process is repeated until the mesh is fully refined.
 }
    \label{fig:asmr_schematic}
    \vspace{-0.2cm}
\end{figure}

\textbf{This paper extends a previously published conference paper on ASMR}~\citep{freymuth2024swarm}. 
While the original ASMR policies produced meshes at a fixed granularity, we now \rebuttal{introduce an adaptive element penalty. This scalar value is added to the reward function to penalize over-refinement and is provided to the policy as context, allowing a single model to generate meshes of varying densities.} 
Additional extensions include an improved mapping between agents over time that acts as a regularizer and leads to better results, an improved network architecture and an improved reward formulation based on the reduction of the maximum local error of the mesh. 
While this reward formulation was already introduced and ablated in our prior work with inconclusive results, we now present a large scale study that shows its improved performance compared to the previously used reward. 
We refer to the improved version as~\gls{method}, and to the previous version as~\gls{asmr}.
Experimentally, we find that the proposed changes substantially improve the method on more difficult task setups.
We further provide additional visualizations of our method, including a schematic of the agent mapping and a qualitative error comparison to a uniform mesh.
Finally, we add two challenging tasks that utilize Neumann boundary conditions and a $3$-dimensional domain to showcase the applicability of our method to a wider range of applications.
Figure \ref{fig:asmr_schematic} provides a schematic overview of a single~\gls{method} refinement step.

We validate the effectiveness of~\gls{method} on a wide range of~\glspl{pde} that require complex refinement strategies.
Our experiments use triangular and tetrahedral meshes \rebuttal{for simplicitly, although \gls{method} is agnostic to the element type of the underlying mesh.
We employ} conforming elements and corresponding h-adaptive refinements~\citep{arnold2000locally, stevenson2008completion}, i.e., refinements via element subdivision, due to their prevalence in engineering~\citep{ho1988finite, jones1997adaptive, nagarajan2018conforming}.
Compared to non-stationary settings from related work, which only consider a low maximum refinement depth or local refinements, the challenge here is to find multiple levels of precise refinement on the full mesh.
We consider several recent \gls{rlamr} methods as baselines~\citep{yang2023reinforcement, yang2023multi, foucart2023deep}.
As these methods have been shown to work for mesh refinement and coarsening with a relatively low depth on dynamic tasks, we adapt them to our setting of stationary meshes that require multiple precise refinement steps and thousands of elements.
We further compare to a threshold-based refinement heuristic that uses the popular \gls{zz} estimate~\citep{zienkiewicz1992superconvergent}. 
Additionally, we consider different \textit{oracle error estimates}, so called because they rely on privileged information from a high-resolution uniform mesh to approximate the true solution error, to mark elements for refinement\footnote{We implement all tasks as OpenAI gym~\citep{brockman2016openai} environments
and publish our code, including all approaches and tasks presented in this paper, at \url{https://github.com/NiklasFreymuth/ASMRplusplus}.}.

\section{Adaptive Swarm Mesh Refinement++}
\label{sec:asmr} 

\gls{method} treats elements of a mesh as a swarm of homogeneous agents that collaborate to find an optimal refinement.
For this, each agent's state and observation are defined through its topological position in the mesh and includes local, rotation- and translation-invariant features of both the mesh and the~\gls{pde}.
At each step, all agents decide whether to mark their respective element for refinement. 
Agents that refine their element receive a \textit{local} reward based on how much their refinement has improved the quality of the underlying simulation.
These local rewards are aggregated into a global term for the return to ensure that each mesh element optimizes its local region while contributing to the global solution quality.
Crucially, each refinement subdivides the refined elements, introducing new agents in the process.
We introduce a~\gls{asmdp} that features a mapping of agents over time, allowing us to optimize over multiple refinement steps by propagating reward from agents of later time steps back to related earlier agents.
Figure~\ref{fig:asmr_schematic} provides a schematic overview of~\gls{method}, and the following sections describe the individual aspects in more detail.

\subsection{Adaptive Swarm Markov Decision Process.}
We view mesh refinement as a collaborative multi-agent~\gls{rl} problem with a changing number of homogeneous agents and agent-wise rewards.
For this, we adopt a swarm~\gls{rl} view, adapting the SwarMDP framework~\citep{vsovsic2017inverse, huttenrauch2019deep} to vector-valued rewards and changing state, action and observation spaces.
Formally, let an \glsfirst{asmdp} be a tuple $\langle \mathbb{S}, \mathbb{O}, \mathbb{A}, P, \mathbf{r}, \xi, \phi \rangle$. 
The state space, observation space and action space are given by $\mathbb{S}$, $\mathbb{O}$, and $\mathbb{A}$ respectively.
Since the number of agents changes over time, we denote subsets of the state, observation, and action spaces with $N$ agents as $\mathcal{S}^N\subset\mathbb{S}$, $\mathcal{O}^N\subset\mathbb{O}$, $\mathcal{A}^N\subset\mathbb{A}$.
The transition function $P : \mathcal{S}^N \times \mathcal{A}^N \rightarrow \mathcal{S}^M$ takes an action over $N$ agents and leads to a new state with $M$ agents.
The reward function $\mathbf{r} : \mathcal{S}^N \times \mathcal{A}^N \rightarrow \mathbb{R}^N$ takes the full state into account to produce a scalar reward for each agent.
Similarly, the observation function $\xi: \mathcal{S}^N \rightarrow \mathcal{O}^N$ calculates local observations for each agent from the global state.

We consider a finite-horizon setting with \rebuttal{a horizon of} $T$ steps.
In the scalar reward case, i.e., with $r(\mathbf{s},\mathbf{a})\in\mathbb{R}$, an optimal \gls{rl} policy $\pi: \mathbb{O} \times \mathbb{A} \rightarrow [ 0, 1 ]$ generally maximizes the return, i.e., the expected cumulative future reward
\begin{equation}
\label{eq:scalar_return}
J^t:=\mathbb{E}_{\pi(\mathbf{a}|\xi(\mathbf{s}))}\left[\sum_{k=0}^{T} r(\mathbf{s}^{t+k},\mathbf{a}^{t+k})\right]\text{.}
\end{equation}
A key challenge of using \gls{rl} for \gls{amr} is the changing number of agents introduced by splitting agents that are marked to be refined. 
To adapt the SwarMDP framework and its underlying Markov Decision Process to changing numbers of agents within a single episode, we thus introduce an agent mapping $\phi^t \in \mathbb{R}^{N\times M}$.
Intuitively, each entry $\phi^t_{ij}$ describes the responsibility of agent $i$ at step $t$ for agent $j$ at step $t+1$.
The responsibilities of agents at step $t$ for agents at step $t'>t$ can then be computed via the matrix multiplication 
$$\phi^{t,t'}:=\phi^{t}\phi^{t+1}\dots\phi^{t'}=\Pi_{\tau=t}^{t'}\phi^\tau\text{.}$$
Given this mapping, we can propagate the rewards of all future agents for which agent $i$ is responsible for at step $t$ to calculate a return
\begin{equation}
\label{eq:mapped_return}
J_i^t:=\mathbb{E}_{\pi(\mathbf{a}|\xi(\mathbf{s}))}\left[~\sum_{t'=t}^{T}\gamma^k(\phi^{t,t'}\mathbf{r}(\mathbf{s}^{t'},\mathbf{a}^{t'}))_i\right]\text{.}
\end{equation}
For training, e.g., a value function $V(\mathbf{s})$, this leads to a TD error~\citep{sutton2018reinforcement}
\begin{equation}
\label{eq:td_error}
\delta^t= \mathbf{r}(\mathbf{s}^t, \mathbf{a}^t)_i+\sum_j \phi^t_{ij} V_j(\mathbf{s}^{t+1}) - V_i(\mathbf{s}^t)
\end{equation}
of actions \mbox{$\mathbf{a}\sim \pi(\xi(\mathbf{s}))$}.
We derive the targets for $Q$-functions analogously.

The~\gls{asmdp} framework allows for a collaborative optimization of changing numbers of agents, each of which is equipped with its own reward.
Compared to, e.g., a centralized partially observable~\gls{mdp}~\citep{yang2023reinforcement, foucart2023deep}, it provides a more efficient learning environment, since every environment step provides a learning signal for each agent.
Similarly, the~\gls{asmdp} supports a multi-agent setting with changing numbers of homogeneous agents, circumventing the posthumous credit assignment problem~\citep{cohen2021use} without having to resort to dummy states and learned value decompositions~\citep{yang2023multi}.
The agent mapping additionally allows for an optimization of the return, i.e., the reward over time, which would not be possible with a purely local per-agent view. 
While this work focuses on~\gls{amr}, the~\gls{asmdp} framework can readily be applied to other scenarios that are characterized by localized decisions of changing numbers of entities.

\subsection{Actions and Agent Mappings.}
In the context of~\gls{amr}, a system state $s\in\mathbb{S}$ is defined by the problem domain $\Omega$, a mesh discretization $\Omega$, and a \gls{pde} formulated on $\Omega$ to be solved under specific conditions.
\gls{method} views every element $\Omega^t_i$ of a mesh $\Omega^t:=\{\Omega^t_i \subseteq \Omega |~ \dot{\bigcup}_i~\Omega^t_i = \Omega \}$ as one of many homogeneous agents. 
Each agent has a binary action space, deciding on whether it wants to mark its element for refinement or not.
These markings are provided to a remesher, yielding a finer mesh $\Omega^{t+1} = \{\Omega_{j}^{t+1}\}_j$.
The remesher subdivides each refined element $\Omega_i^t$ into a set of smaller elements $\{\Omega_{j}^{t+1}\}_j$, such that the disjoint union of all $\Omega_{j}^{t+1}$ reconstructs $\Omega_i^t$, i.e., $\dot{\bigcup}_j \Omega_{j}^{t+1} = \Omega_i^t$.
It may also refine unmarked elements to close the mesh and thus assert a conforming solution~\citep{arnold2000locally}, i.e., to make sure that elements of the mesh align with each other at the boundaries to ensure continuity of solution variables between adjacent elements.

To map between elements and their successors, \gls{asmr} uses the indicator function $\phi^t_{ij} := \mathbb{I}(\Omega^{t+1}_j\subseteq \Omega^t_i)$.
This mapping assigns each element to all elements that it subdivides into, or equivalently, lets each agent be responsible for all new agents that it spawns.
Here, we extend the mapping to include a normalization factor
\begin{equation}
    \label{eq:agent_mapping}
    \phi^t_{ij} := \frac{|\Omega^t|}{|\Omega^{t+1}|}\mathbb{I}(\Omega^{t+1}_j\subseteq \Omega^t_i)\text{.}
\end{equation}
Compared to~\gls{asmr}, the normalization factor $\frac{|\Omega^t|}{|\Omega^{t+1}|}$ scales the responsibilities such that $\sum \phi^t=\sum \phi^{t'}$.
This procedure intuitively ensures that the total `mass' of agents is preserved over time.
We find that this factor regularizes the mapping of rewards within the agent swarm, akin to, e.g., batch normalization~\citep{ioffe2015batch}.

\begin{figure}
    \centering
	\includegraphics[width=\textwidth]{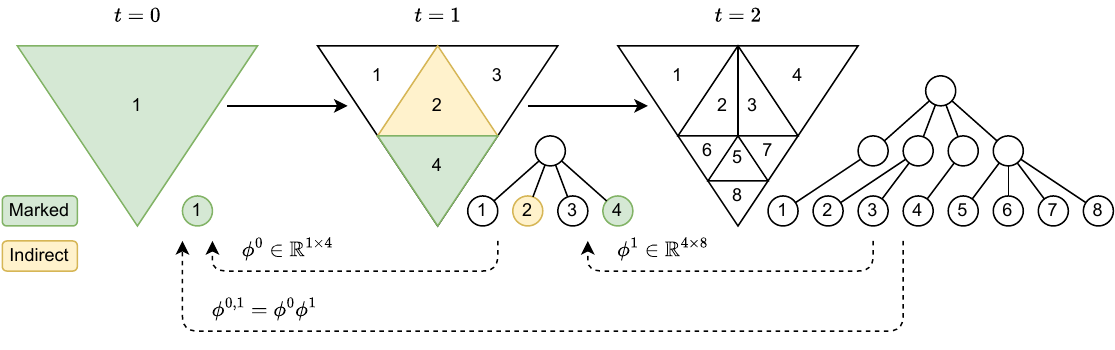}
	\caption{
 Refinement procedure and responsibility mapping of~\gls{method}. 
\textbf{Left:} An initial mesh has its single element marked for refinement.
\textbf{Middle:} The mesh element is subdivided into $4$ new elements. 
The correspondence between the old and new mesh is represented as a directed acyclic graph. 
Constructing the matrix $\phi^0$ from this graph allows us to map the new elements back to the old one. 
For the next step, Element $4$ is marked for refinement, requiring an indirect refinement of Element $2$ to ensure a conforming mesh.
\textbf{Right:} After the second refinement, the resulting mesh consists of $8$ elements. 
Using the refinement graph to construct the mapping $\phi^1$ allows us to compute responsibilities of elements in the previous mesh for this mesh. 
Chaining these mappings as $\phi^{0,1}=\phi^{0}\phi^{1}$ allows us to directly compute responsibilities between the initial mesh and the mesh after two refinement steps.
 }
    \label{fig:agent_mapping}
\end{figure}

Conceptually, we may represent each mesh in a series of refinements as a layer in a hierarchical graph, with nodes corresponding to mesh elements. 
\rebuttal{
In our case, this graph is a simple tree.
Here, each node connects to its predecessor in the prior layer and to all successors in the subsequent layer.
}
Applying an agent mapping equates to navigating through the respective level of hierarchy of this graph.
An example refinement procedure and its corresponding refinement graph for an initial mesh with a single element and $2$ refinement steps can be seen in Figure~\ref{fig:agent_mapping}
.

This work focuses on adaptive mesh refinement via iterative mesh subdivision.
For mesh coarsening, the above mapping can be extended to respect element coarsening as, e.g., 
$$\phi^t_{ij} := \frac{|\Omega^t|}{|\Omega^{t+1}|} \left[\mathbb{I}(\Omega^{t+1}_j\subseteq \Omega^t_i) + \left(\mathbb{I}(\Omega^t_i\subsetneq \Omega^{t+1}_j)/\left(\sum_k \mathbb{I}(\Omega^t_k\subsetneq \Omega^{t+1}_j)\right)\right)\right]\text{.}$$
In this case, the hierarchical graph is no longer a tree but instead a directed acyclic graph, since multiple elements can coarsen into the same new element. 
Similar modifications can be made to adapt~\gls{method} to other~\gls{amr} and adaptive mesh generation methods.

\subsection{Observations and Policy Architecture.}
\label{ssec:mpn_description}

Given a mesh $\Omega^t$ and its corresponding system of equations, we construct an observation as the bidirectional graph $\gG_{\Omega^t} = \gG = (\mathcal{V}, \mathcal{E}, \mathbf{X}_\mathcal{V}, \mathbf{X}_\mathcal{E})\in\mathcal{O}$ with mesh elements as nodes $\mathcal{V}$ and their neighborhood relation as edges $\mathcal{E} \subseteq \mathcal{V}\times \mathcal{V}$.
Mesh- and task-dependent node and edge features of dimensions $d_\mathcal{V}$ and $d_\mathcal{E}$ are provided as $\mathbf{X}_\mathcal{V}: \mathcal{V}\rightarrow \mathbb{R}^{d_\mathcal{V}}$ and $ \mathbf{X}_\mathcal{E}: \mathcal{E}\rightarrow \mathbb{R}^{d_\mathcal{E}}$.

This observation graph is consumed by an~\gls{mpn} policy.
\glspl{mpn}~\citep{pfaff2020learning, linkerhaegner2023grounding} are a popular~\gls{gnn} architecture for mesh-based physical simulation~\citep{pfaff2020learning, linkerhaegner2023grounding, wurth2024physics} as they encompass the function class of several classical~\gls{pde} solvers~\citep{brandstetter2022message}.
Given an input graph, they iteratively update \rebuttal{high-dimensional latent representations} on this graph for $L$ consecutive message passing steps. 
Each step $l$ receives the output of the previous step and updates the latent states $\mathbf{X}_\mathcal{V}$ , $\mathbf{X}_\mathcal{E}$ for all nodes $v\in \mathcal{V}$ and edges $e\in\mathcal{E}$.
Using linear embeddings $\mathbf{x}_v^0$ and $\mathbf{x}_e^0$ of the initial node and edge features, the $l$-th step is given as
$$
\mathbf{x}^{l+1}_{e} = f^{l}_{\mathcal{E}}(\mathbf{x}^{l}_v, \mathbf{x}^{l}_u, \mathbf{x}^{l}_{e}), \textrm{ with } e = (u, v)\text{,}
$$
$$
\mathbf{x}^{l+1}_{v} = f^{l}_{\mathcal{V}}(\mathbf{x}^{l}_{v}, \bigoplus_{e=(v,u)\in \mathcal{E}} \mathbf{x}^{l+1}_{e})\text{.}
$$
The operator $\oplus$ is a permutation-invariant aggregation such as a sum, mean or maximum over all aggregated elements.
Each $f^l_\cdot$ is a learned function, usually parameterized as a simple \gls{mlp}.
The~\gls{mpn}'s final output is a learned representation $\mathbf{x}^L_v$ for each node $v\in\mathcal{V}$.
This learned representation can be fed into a policy head to compute an action per agent, yielding a joint action vector~$\pi(\mathbf{a}|\mathcal{G}_{\Omega^t})$, or into a value head to compute a per-element value function estimate.
Both are parameterized as~\glspl{mlp} that are shared between the different agents.
Since~\glspl{mpn} are permutation-equivariant by design~\citep{bronstein2021geometric}, this policy architecture ensures that the amount and ordering of mesh elements does not matter, i.e., that the elements are only defined by their features and local position in the mesh.

\rebuttal{
We note recent trends in surrogate modelling for engineering applications~\citep{alkin2024universal, alkin2025ab} and \gls{rl} for continuous control~\citep{nauman2024bigger, lee2024simba} increasingly favor heavily regularized large-scale architectures, such as transformers~\citep{vaswani2017attention}.
While our \gls{mpn} architecture offers several desirable inductive biases, these larger and more complicated policy backbones are a promising extension that may enable \gls{method} to scale to larger domains and highly non-linear \glspl{pde}.
}

\subsection{Reward Definition}
\label{ssec:reward}
\glsfirst{amr} aims to generate meshes with enhanced local resolution in areas that are relevant for the underlying system of equations. 
Its objective is to achieve an optimal balance between the solution's accuracy on the mesh $\Omega^t$, and its total element count $\Omega^t_i \in \Omega^t$.

Since closed-form solutions are generally not available for most systems of equations, we compute an approximate ground truth solution $u_{\Omega^*}$ using a fine-grained uniform reference mesh $\Omega^*$~\citep{yang2023reinforcement}.
\rebuttal{While calculating such a reference solution is expensive, it only needs to be computed once per training environment and is not required during inference.}
Given $u_{\Omega^*}$, a refined mesh $\Omega^t$ and its solution $u_{\Omega^t}$, we define the \textit{error per element} as the maximum difference between the solutions on this element.
We use the maximum difference over solutions as a measure of simulation quality that is independent of, e.g., the size of the used elements, simplifying the evaluation compared to, e.g., integrated $L2$ error computations.
Given reference elements $\Omega^*_m$ with midpoints $p_{\Omega_m^*} \in \Omega^*_m$ as sampling points to query the solutions on, this yields an error
\begin{equation}
    \label{eq:err_per_element}
    \hat{\text{err}}(\Omega_i^t)\approx \max_{\Omega_m^*\subseteq \Omega_i^t} \Big|u_{\Omega^*}(p_{\Omega_m^*})-u_{\Omega^t}(p_{\Omega_m^*})\Big|\text{.}  
\end{equation}

We efficiently calculate the assignment $\Omega^*_m\in\Omega^t_i$ required for this reward using a $k$\rebuttal{-}d tree~\citep{bentley1975multidimensional}.
If a queried point $p_{\Omega_m^*}$ exactly lies on the edge between two elements in $\Omega^t$, which is possible for some refinement algorithms, we assign it to both of these elements with a weight of $0.5$. 
We normalize this error estimate with the initial mesh error to get the relative improvement of the mesh-error, i.e., 
\mbox{
$
\text{err}(\Omega_i^t) = \hat{\text{err}}(\Omega_i^t)/\sum_{\Omega_j^0\in \Omega^0}\hat{\text{err}}(\Omega_j^0)
$}.
This estimate is consistent across geometries and process conditions.
As the size of the solution $u_{\Omega^*}$ and subsequently that of the error $\hat{\text{err}}(\Omega_i^t)$ may differ significantly between systems of equations due to varied process conditions, this normalization prevents systems of equations with large solution quantities from dominating the learning process.
Given the normalized error estimate, we next construct an element-wise local reward function that estimates the benefit of a refinement and compares it to the cost of adding additional elements.
\rebuttal{In systems of equations with multiple quantities of interest, such as, e.g., elasticity problems where both the element displacement and stresses are important, the mesh must be optimized for all the quantities of interest. 
Here, we calculate rewards independently for each normalized solution dimension and use some problem-dependent average or norm to compute a scalar reward per element.}

\rebuttal{To define our reward,} we introduce a variable element penalty $\alpha$ that penalizes each added element, and compare this to the decrease in error within each element when it is refined, yielding  
\begin{equation}
\label{eq:asmr_local_reward}
\mathbf{r}(\Omega^t_i) := \left(\text{err}(\Omega^t_i)
-\max_j\phi^t_{ij}\text{err}(\Omega^{t+1}_j)\right)
-\alpha\left(\sum_j\mathbb{I}(\Omega^{t+1}_j\subseteq \Omega^t_i)-1\right)\text{.}
\end{equation}

The local reward for each element is $0$ if no refinement is made.
If an element is refined, its reward is positive if and only if the refinement decreases the highest error in this element by more than the cost of adding new elements, which is scaled with the element penalty $\alpha$.
In comparison, \gls{asmr}~\citep{freymuth2024swarm} rewards the integrated average change in error for each mesh element instead of its maximum change in error.
This average change in error is integrated over the element volume, meaning that small elements, i.e., those that have been previously refined, quickly become comparatively insignificant for the reward.
To counteract this,~\gls{asmr} scales each reward by the inverse of the respective element's volume, leading to widely varying and potentially unbound reward values that do not directly match any optimization criteria on the full mesh. 
We instead directly reward the change in maximum error within the volume of each element.
This maximum reward variant has been proposed as an ablation study in~\citep{freymuth2024swarm}, but only been explored for a single task.
We find that the maximum reward results in more well-behaved optimization, presumably because it directly targets refining elements where the reduction in error justifies adding new elements.
We provide a more detailed comparison to the~\gls{asmr} reward in~\ref{app_ssec:volume_reward}.
Since Equation~\ref{eq:asmr_local_reward} is based on an element's decrease in maximum error, it is independent of the element size, ensuring that rewards of elements of different sizes are on the same scale without the need for explicit scaling.

Equation~\ref{eq:asmr_local_reward} rewards each agent for its local refinement decision for each time step.
While this direct and explicit local credit assignment leads to a temporally and spatially dense reward, it disregards non-local effects of mesh refinement, which are common for elliptical~\glspl{pde}.
To encourage global optimization, we therefore extend the return $J_i^t$ of Equation~\ref{eq:mapped_return} to include a global term as
\begin{equation}
    \label{eq:half_half_return}
    {J_i^t}'=\frac{1}{2} J^t_i+\frac{1}{2} J^t\text{,}
\end{equation}
where $J^t$ is the global return of Equation~\ref{eq:scalar_return} using the average reward \mbox{$r=\frac{1}{N}\sum_j \mathbf{r}_j$}.

\subsection{Element Penalty}
\label{ssec:element_penalty}

We add the element penalty $\alpha$ in Equation~\ref{eq:asmr_local_reward} \rebuttal{as} an input to the policy to inform it about the penalty of adding new elements.
During training, we draw $\alpha$ from a log-uniform distribution at the start of each episode, allowing for environment samples on a wide range of different penalties.
Similar to~\gls{vdgn}~\citep{yang2023multi}, we then specify a value of $\alpha$ during inference to control the refinement level of the created mesh.
Compared to~\gls{asmr}, which trains a policy on a fixed element penalty value, we thus condition the policy on~$\alpha$ for each rollout.
This conditional policy allows for different mesh granularities during inference, whereas~\gls{asmr} needs to train a new policy when the desired number of mesh elements changes.

\begin{figure}
    \centering
    \begin{minipage}{0.02\textwidth}
            \rotatebox{90}{$\alpha=0.02$\phantom{0}}
    \end{minipage}
    \begin{minipage}{0.14\textwidth}
            \centering
            \includegraphics[width=\textwidth]{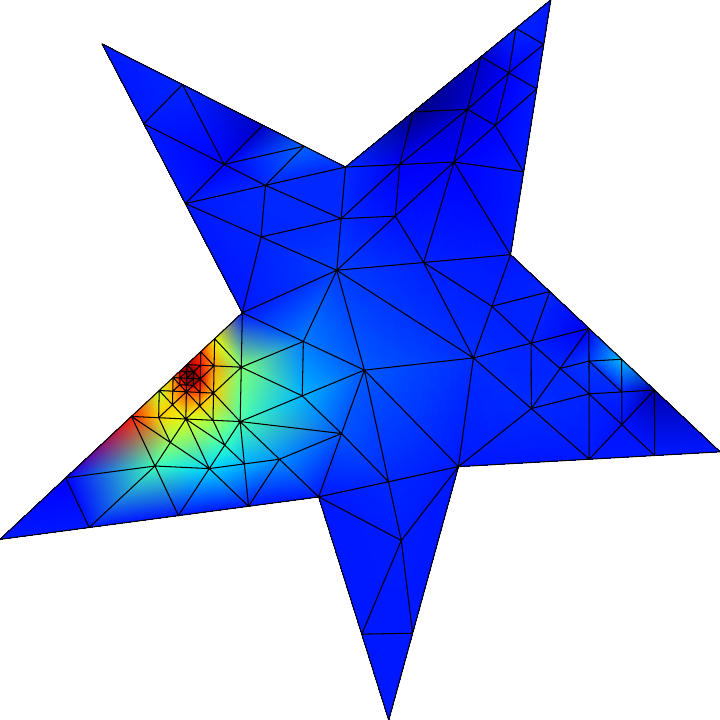}
    \end{minipage}%
    \begin{minipage}{0.02\textwidth}
    ~
    \end{minipage}
    \begin{minipage}{0.02\textwidth}
            \rotatebox{90}{$\alpha=0.01$\phantom{0}}
    \end{minipage}
    \begin{minipage}{0.14\textwidth}
            \centering
            \includegraphics[width=\textwidth]{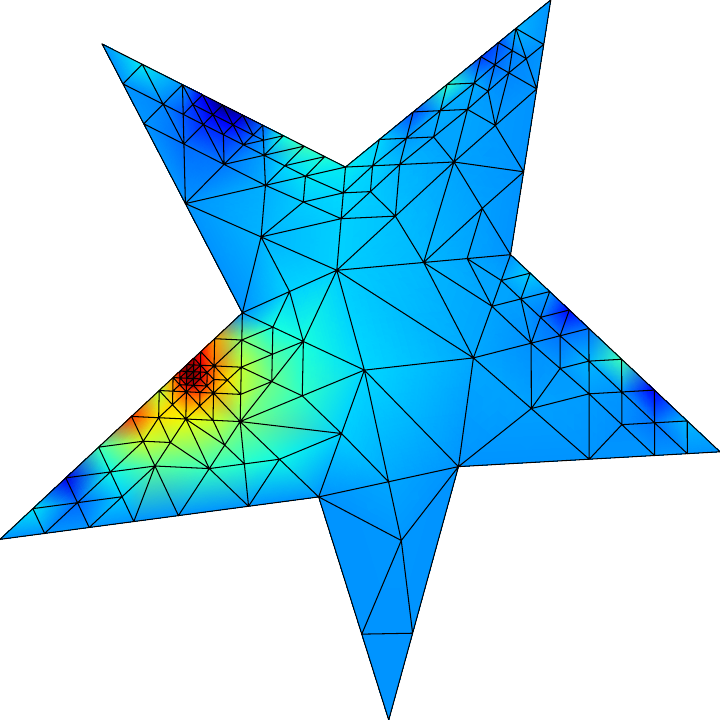}
    \end{minipage}%
    \begin{minipage}{0.02\textwidth}
    ~
    \end{minipage}
    \begin{minipage}{0.02\textwidth}
            \rotatebox{90}{$\alpha=0.005$}
    \end{minipage}
    \begin{minipage}{0.14\textwidth}
            \centering
            \includegraphics[width=\textwidth]{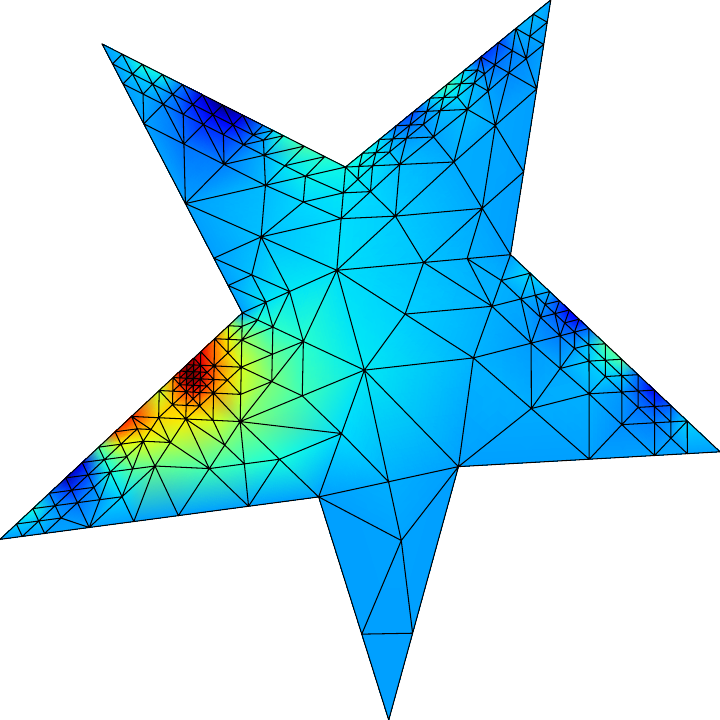}
    \end{minipage}%
    \begin{minipage}{0.02\textwidth}
    ~
    \end{minipage}
    \begin{minipage}{0.02\textwidth}
            \rotatebox{90}{$\alpha=0.003$}
    \end{minipage}
    \begin{minipage}{0.14\textwidth}
            \centering
            \includegraphics[width=\textwidth]{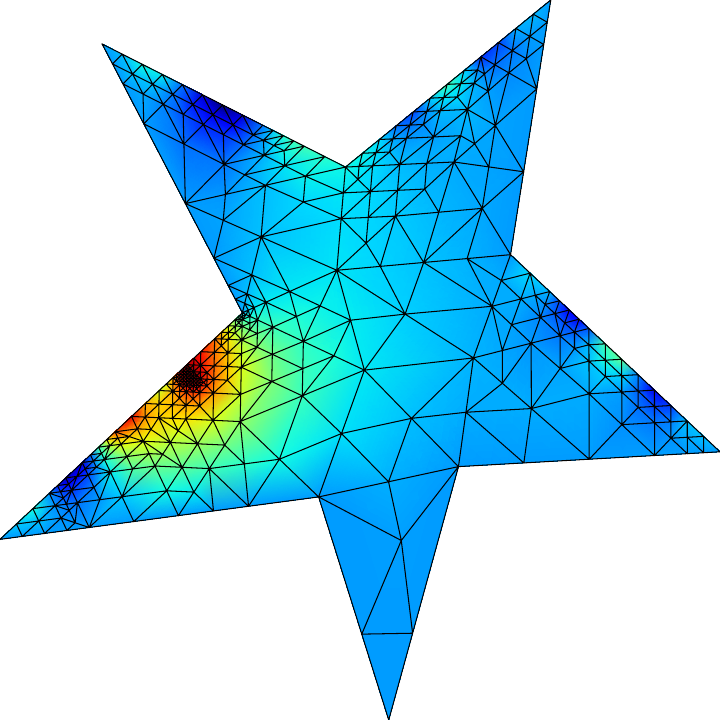}
    \end{minipage}%
    \begin{minipage}{0.02\textwidth}
    ~
    \end{minipage}
    \begin{minipage}{0.02\textwidth}
            \rotatebox{90}{$\alpha=0.002$}
    \end{minipage}
    \begin{minipage}{0.14\textwidth}
            \centering
            \includegraphics[width=\textwidth]{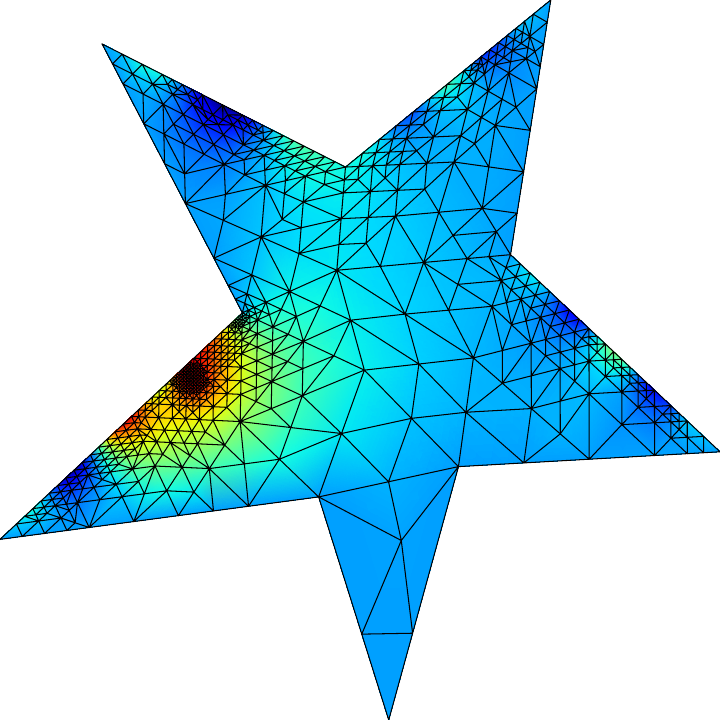}
    \end{minipage}%
    
    \begin{minipage}{0.02\textwidth}
            \rotatebox{90}{$\alpha=0.001$\phantom{0}}
    \end{minipage}
    \begin{minipage}{0.28\textwidth}
            \centering
            \includegraphics[width=\textwidth]{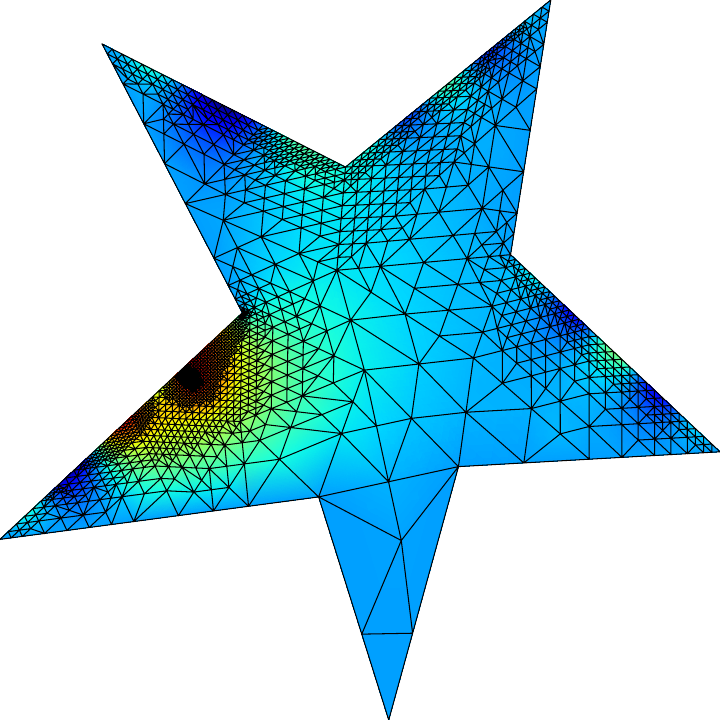}
    \end{minipage}%
    \begin{minipage}{0.02\textwidth}
    ~
    \end{minipage}
    \begin{minipage}{0.02\textwidth}
            \rotatebox{90}{$\alpha=0.0005$}
    \end{minipage}
    \begin{minipage}{0.28\textwidth}
            \centering
            \includegraphics[width=\textwidth]{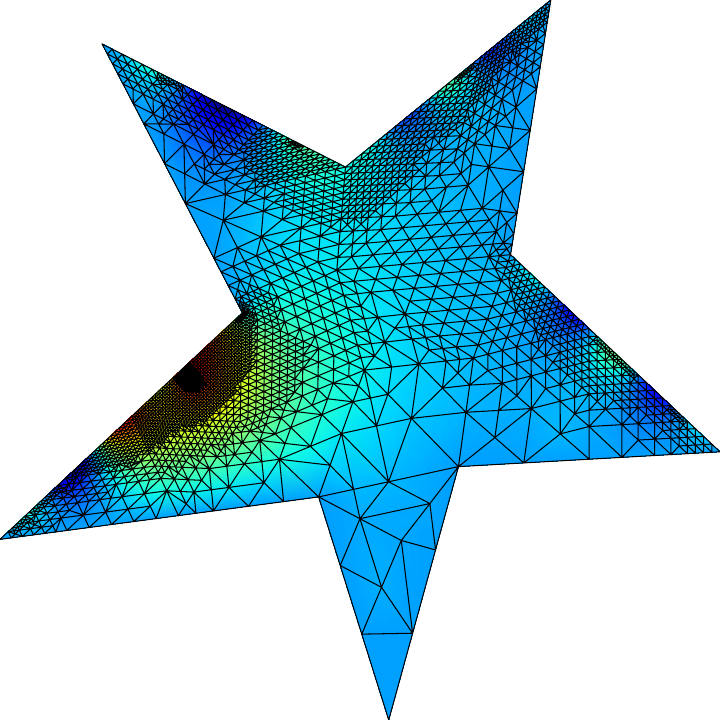}
    \end{minipage}%
    \begin{minipage}{0.02\textwidth}
    ~
    \end{minipage}
    \begin{minipage}{0.02\textwidth}
            \rotatebox{90}{$\alpha=0.0003$}
    \end{minipage}
        \begin{minipage}{0.28\textwidth}
            \centering
            \includegraphics[width=\textwidth]{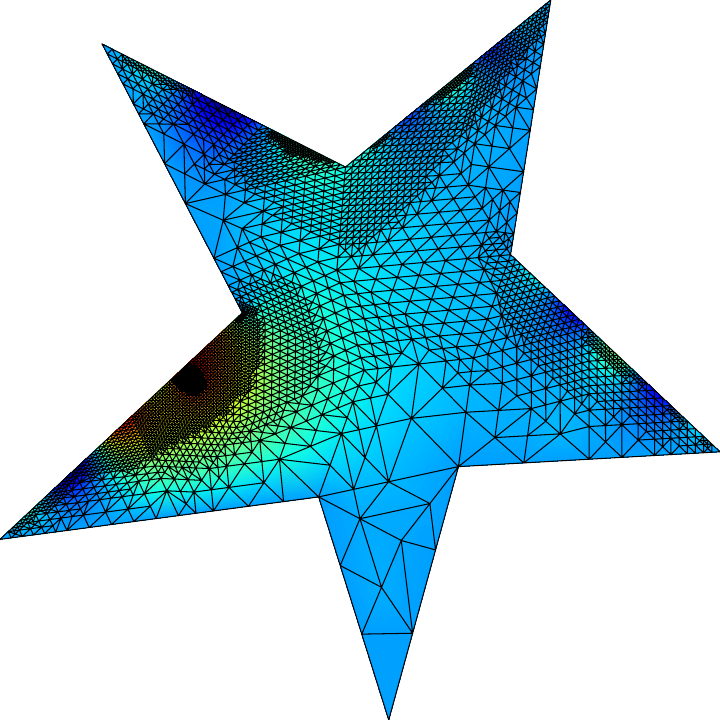}
    \end{minipage}%
    \vspace{0.01\textwidth}%

\caption{
    \gls{method} refinements for a Poisson problem with sinusoidal Neumann boundary conditions for different inputs of the element penalty $\alpha$.
    All refinements focus on the relevant parts of the problem, and lower element penalties lead to more fine-grained meshes.
    }
    \label{fig:element_penalties}
\end{figure}

Balancing the extra computational cost of adding mesh elements against the benefit of reducing simulation errors can be seen as an instance of multi-objective~\gls{rl}~\citep{van2014multi, hayes2022practical}.
The cost of adding new elements increases linearly with a penalty factor $\alpha$, and there exists some approximate ranking of the effectiveness of different mesh refinements at reducing simulation errors. 
This setup creates a roughly convex relationship between $\alpha$ and the policy's behavior.
In other words, lowering $\alpha$ generally leads to a finer mesh that includes all refinements made with a higher $\alpha$, while raising $\alpha$ results in fewer refinements throughout the mesh.
Figure~\ref{fig:element_penalties} provides an example of resulting~\gls{method} meshes for the same trained policy and Neumann boundary task for different values of the element penalty~$\alpha$.
Regardless of the penalty, the parts of the mesh with the highest potential error reduction, in this case the areas near the load function and the sinusoidal Neumann boundaries, are refined the most. 

\subsection{Pseudocode}
\label{ssec:pseudocode}

\rebuttal{
Algorithms~\ref{alg:asmr_train} and \ref{alg:asmr_inference} provide pseudocode for the training and inference procedures of \gls{method}.
\gls{method} iteratively interacts with a set of systems of equations during training, obtaining a reward by comparing the solution quality of its refined meshes to a fine-grained reference $\Omega^*$.
After training has finished, it can be applied to unseen problems without requiring access to such a fine-grained reference, providing accurate refinements from only the current coarse-mesh solution.
}

\begin{algorithm}[t]
    \SetAlgoLined
    \DontPrintSemicolon
    \caption{\rebuttal{ASMR++ Training}}
    \label{alg:asmr_train}
    \KwInput{Training environment $\mathcal{E}$ (Sampler for PDE, $\Omega^0$, $\Omega^*$)}
    \KwInput{Refinement steps $T$, Penalty range $[\alpha_{\min}, \alpha_{\max}]$}
    \KwInput{Initial policy $\pi_\theta$}
    \KwOutput{Optimized policy $\pi_\theta$}
    
    \While{not converged}{
        \textbf{Data Generation:}\;
        Initialize buffer $\mathcal{B} \leftarrow \{\}$\;
        \For{each rollout}{
            Sample $(\text{PDE}, \Omega^0, \Omega^*) \sim \mathcal{E}$\;
            Sample penalty $\alpha \sim \text{LogUniform}(\alpha_{\min}, \alpha_{\max})$\;
            \For{$t = 0 \dots T-1$}{
                Get solution $\mathbf{u}_{\Omega_t}$ from PDE and mesh $\Omega^t$\;
                Construct graph $\mathcal{G}^t$ from $\mathbf{u}_{\Omega_t}$ and $\Omega^t$ (Section~\ref{ssec:mpn_description})\;
                Sample refinement decisions $\mathbf{a}_t \sim \pi_\theta(\mathcal{G}^t, \alpha)$\;
                $\Omega^{t+1}, \phi_t \leftarrow \text{Refine}(\Omega^t, \mathbf{a}_t)$\;
                Compute local rewards $\mathbf{r}_t$ using $\Omega^*$ and penalty $\alpha$ (Equation~\ref{eq:asmr_local_reward})\;
                $\mathcal{B} \leftarrow \mathcal{B} \cup \{(\mathcal{G}^t, \mathbf{a}_t, \mathbf{r}_t, \phi_t)\}$ \;
            }
        }
        \textbf{Policy Update:}\;
        Compute global returns $J^t$ (Equation~\ref{eq:half_half_return}) from rewards in $\mathcal{B}$\;
        Calculate loss and update parameters $\theta$\;
    }
    \KwRet{$\pi_\theta$}
\end{algorithm}

\begin{algorithm}[t]
    \SetAlgoLined
    \DontPrintSemicolon
    \caption{\rebuttal{ASMR++ Inference (Deployment)}}
    \label{alg:asmr_inference}
    \KwInput{Coarse mesh $\Omega^0$, Trained policy $\pi_\theta$}
    \KwInput{Refinement steps $T$, Element penalty $\alpha$}
    \KwOutput{Final refined mesh $\Omega^T$}
    
    \For{$t = 0 \dots T-1$}{
        Get solution $\mathbf{u}_{\Omega_t}$ from PDE and mesh $\Omega^t$\;
        Construct graph $\mathcal{G}^t$ from $\mathbf{u}_{\Omega_t}$ and $\Omega^t$ (Section~\ref{ssec:mpn_description})\;
        Select refinement decisions $\mathbf{a}_t \leftarrow \text{argmax} \ \pi_\theta(\mathcal{G}^t, \alpha)$ \tcp*[r]{Deterministic}
        $\Omega^{t+1}, \_ \leftarrow \text{Refine}(\Omega^t, \mathbf{a}_t)$\;
    }
    \KwRet{$\Omega^T$}
\end{algorithm}
\section{Experiments}

\begin{figure}[ht!]
    \centering
        \begin{minipage}[b]{0.19\textwidth}
            \centering
            \includegraphics[width=\textwidth, height=2.8cm, keepaspectratio]
            {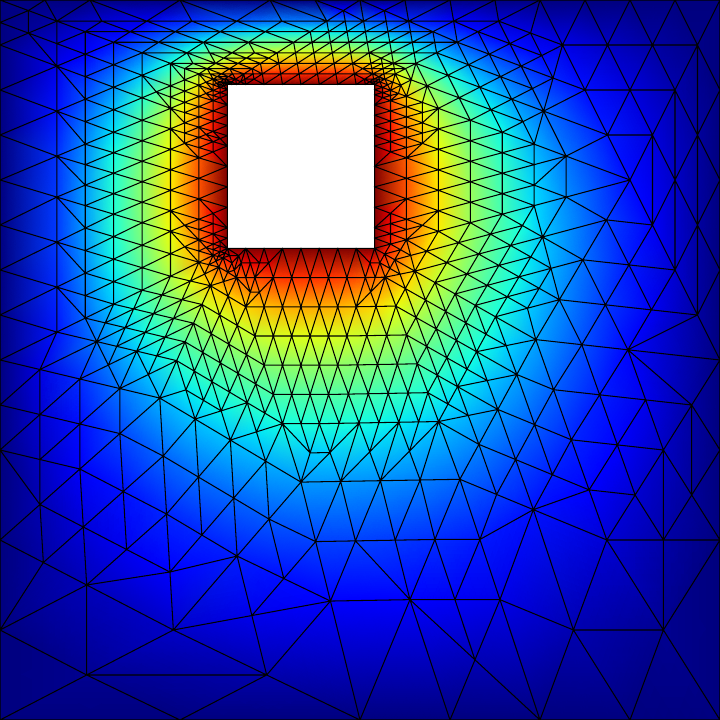}
            \caption*{Laplace}
    \end{minipage}
    \begin{minipage}[b]{0.19\textwidth}
            \centering
            \includegraphics[width=\textwidth, height=2.8cm, keepaspectratio]
            {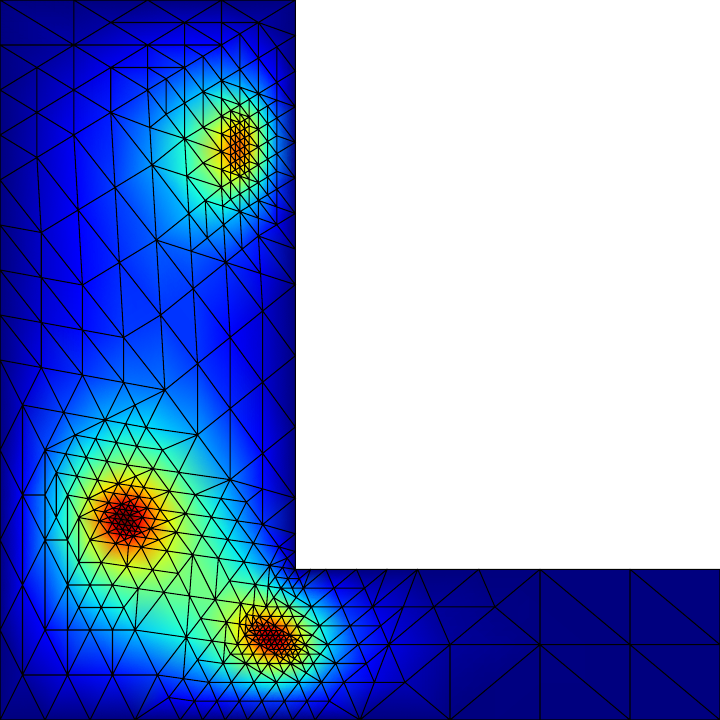}
            \caption*{Poisson}
    \end{minipage}
    \begin{minipage}[b]{0.19\textwidth}
            \centering
            \includegraphics[width=\textwidth, height=2.8cm, keepaspectratio]
            {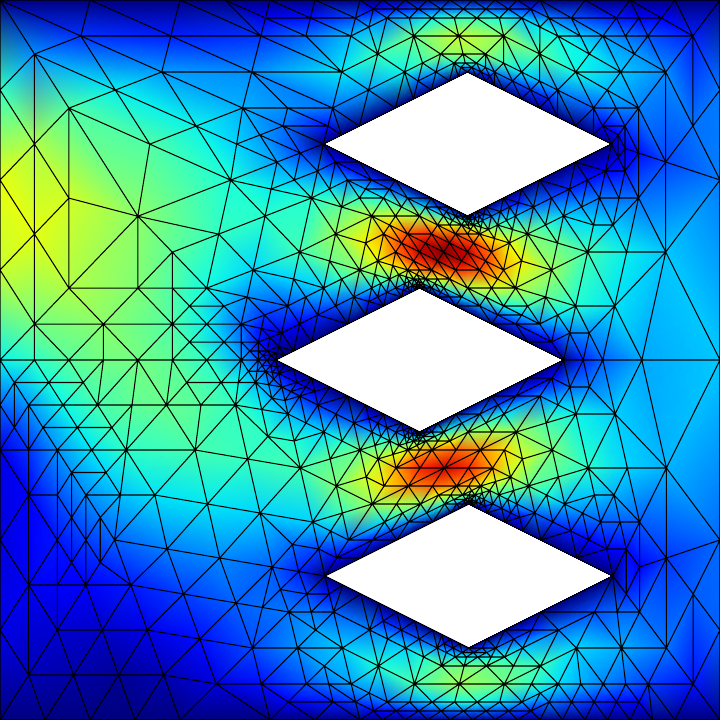}
            \caption*{Stokes Flow}
    \end{minipage}
    \begin{minipage}[b]{0.19\textwidth}
            \centering
            \includegraphics[width=\textwidth, height=2.8cm, keepaspectratio]
            {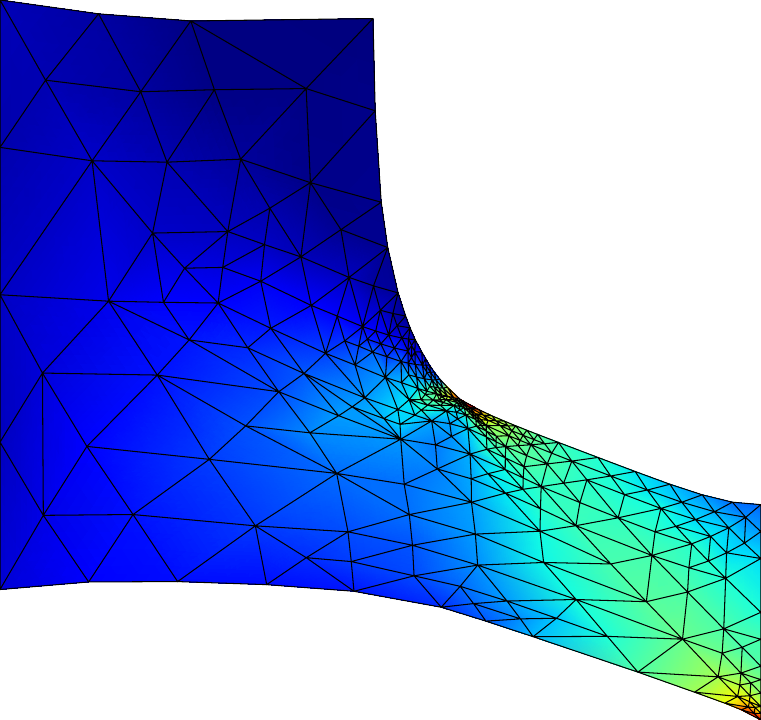}
            \caption*{Linear Elasticity}
    \end{minipage}
    \begin{minipage}[b]{0.19\textwidth}
            \centering
            \includegraphics[width=\textwidth, height=2.8cm, keepaspectratio]
            {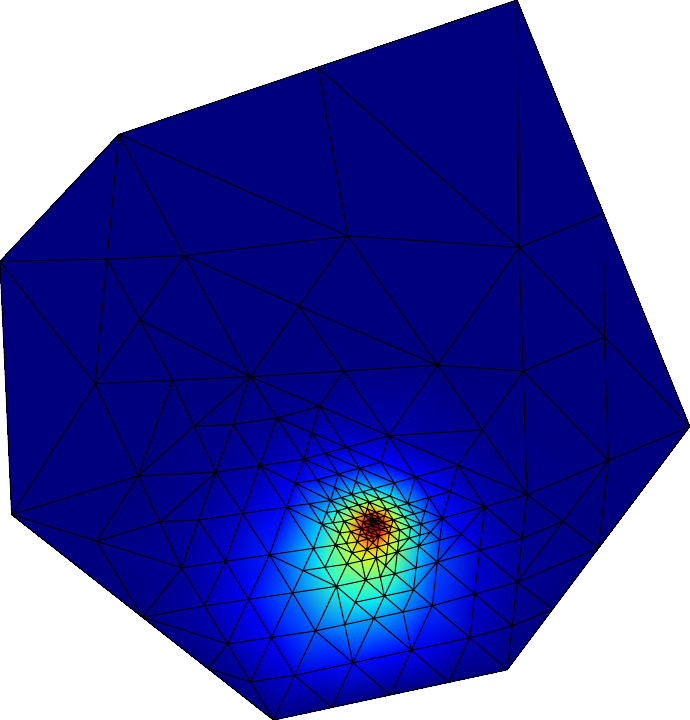}
            \caption*{Heat Diffusion}
    \end{minipage}
    \begin{minipage}[b]{0.40\textwidth}
            \centering
            \includegraphics[width=\textwidth, height=7cm, keepaspectratio]
            {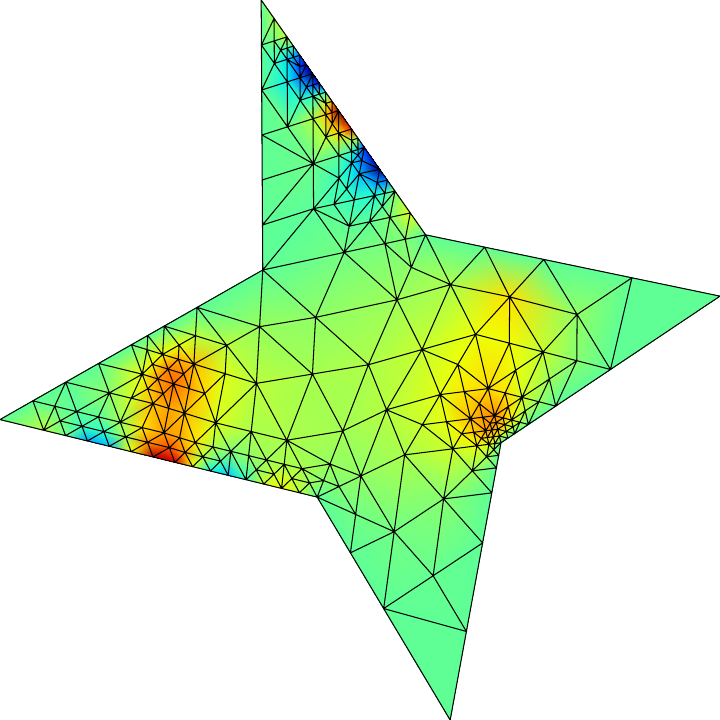}
            \caption*{Neumann Boundaries}
    \end{minipage}
    \begin{minipage}[b]{0.40\textwidth}
            \centering
            \includegraphics[width=\textwidth, height=7cm, keepaspectratio]
            {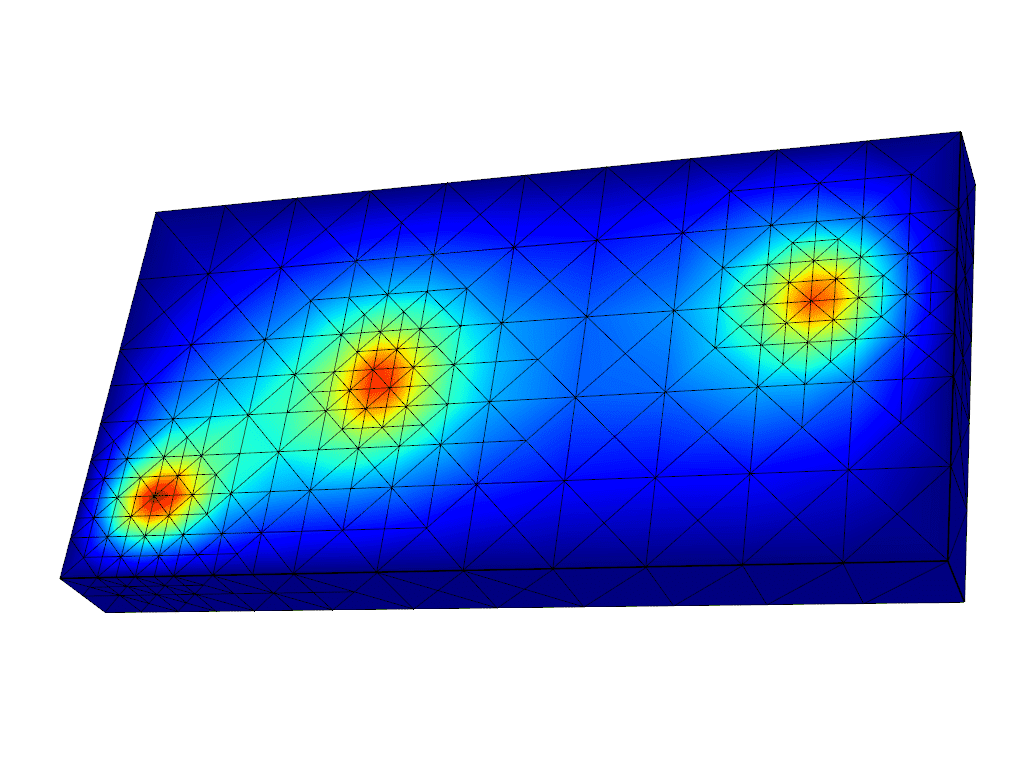}
            \caption*{$3$d Poisson}
    \end{minipage}%
    \vspace{0.01\textwidth}%
    
    \caption{
        Final~\gls{method} meshes and corresponding \gls{fem} solution for different tasks.
        \rebuttal{
        We visualize the velocity norm for Stokes Flow and a weighted average of the displacement norm and Von-Mises stress for Linear Elasticity.
        For the other tasks, we visualize their scalar solution quantity as a heatmap. 
        }
        \gls{method} provides highly adaptive meshes across a wide variety of~\glspl{pde} by accurately optimizing task-specific regions of interests.
        It additionally supports Neumann boundary conditions and volumetric $3$d meshes.
    }
    \label{fig:qualitative_all_tasks}
\end{figure}

\subsection{Systems of Equations.}
\label{ssec:pdes}
The physical behavior of a technical system can be described by a set of~\glspl{pde}.
Simulations predict the physical states of a system by solving those~\glspl{pde}. 
For non-trivial problems, the solving method is usually based on numerical approximation methods, such as the \gls{fem}, which considers the weak formulation of the~\glspl{pde}. 
Multiplying the strong \gls{pde} formulation with a test function $v(\textbf{x}) \in V$ and integrating by parts yields the weak formulation \citep{larsonFiniteElementMethod2013}, which we denote as $\Phi(u,v)$. 
We denote $V$ as the space spanned by a set of basis functions $\phi_v(\textbf{x}) \in V$ and the sought solution as $u(\textbf{x})$.
In the Galerkin \gls{fem}, the solution $u(\textbf{x}, t)$ is approximated by a weighted sum of these basis functions $u(\textbf{x}) = \sum_{v=1}^{N_\varphi} u_v(t) \phi_v(\textbf{x})$. 
Additionally, the same basis function are used to define the test function $v(\textbf{x}) = \sum_{v=1}^{N_\varphi} \phi_v(\textbf{x})$. 
In this work we consider linear elliptical or parabolic \glspl{pde}. 
For elliptical \glspl{pde}, we seek for a space-dependent solution $u=u(\textbf{x})$.
This results in $N_\varphi$ equations for $N_\varphi$ unknown degrees of freedom $u_v$. 
Due to the linearity of the \glspl{pde} and the discretization of the \gls{fem}, we can assemble a linear system of equations $\textbf{A}\textbf{u}=\textbf{b}$ out of the weak formulation $\Phi(\phi_u, u_v, \phi_v)$.
The time-dependent parabolic \glspl{pde} requires a discretization in time, e.g., using forward or backward Euler schemes. 
Subsequently, a system of equation can be assembled for each time step.
We can efficiently solve the obtained linear system of equations with numerical solvers.
For each~\gls{pde}, we create a task that consists of the weak formulations, the domain, and varying process conditions. The goal is to generate meshes that minimize the error of the~\gls{fem}, relative to a fine-grained reference mesh $\Omega^*$, which approximates the true solution of the system of equations, while keeping the number of elements in the generated meshes as low as possible.

Figure~\ref{fig:qualitative_all_tasks} shows examples for each task and visualizes final meshes created using an~\gls{method} policy.
Each task is associated with a family of domains, including L-shapes, rectangles with a square hole or multiple rhomboid holes, convex polygons, star shapes with different numbers of peaks, and a $3$-dimensional plate.
From left to right and top to bottom, these tasks are briefly characterized as follows.
\textit{Laplace's equation} task considers a rectangular heat source at the inner boundary and features high solution gradients near this source.
\textit{Poisson's equation} uses a multi-modal load function comprised of a Gaussian mixture model, which in turn causes multiple distinct regions of interest in the solution.
The \textit{Stokes flow} task solves for the velocity of a fluid streaming into the domain from inlet on the left side. 
It uses more complex shape functions and requires high precision near the corners of the rhomboid obstacles. 
The \textit{linear elasticity} task simulates the deformation and the resulting stress of a metal plate. 
The \textit{heat diffusion} task simulates a time-dependent heat source and is evaluated at the final simulation step.
Adding sinusoidal \textit{Neumann Boundaries} to Poisson's Equation models scenarios where the boundary flux is specified, resulting in high solution variability near the domain edges. 
The \textit{$3$d Poisson} task considers a $3$-dimensional variant of Poisson's equation, using tetrahedral meshes and a different underlying refinement algorithm.

We implement the~\gls{fem}, all systems of equations, meshes and refinements in \textit{scikit-fem}~\citep{gustafsson2020scikit}.
\rebuttal{All tasks are provided as OpenAI gym~\citep{brockman2016openai} environments.}
\rebuttal{While \gls{method} is designed to work with arbitrary element types, our experiments focus on conforming triangular and tetrahedral meshes.
We use} linear elements unless mentioned otherwise.
\rebuttal{Appendix \ref{app_sec:pde} specifies} the systems of equations and their process conditions.

\subsection{Adaptive Mesh Refinement.}
\label{ssec:amr}
For each system of equations, we generate an initial coarse mesh $\Omega^0$, which is then refined during an episode through iterative element subdivision. 
For simplicity, we use the default mesh refinement methods implemented in \textit{scikit-fem}~\citep{gustafsson2020scikit}.
In $2$-dimensional domains, we employ the red-green-blue refinement method~\citep{carstensen2004adaptive}, subdividing each marked element into $4$ smaller elements and then closing the mesh, as illustrated in Figure~\ref{fig:agent_mapping}. 
For $3$-dimensional meshes, we apply longest edge bisection~\citep{rivara1984algorithms, suarez2005computational} to halve each marked element. 
We emphasize that~\gls{method} is compatible with arbitrary refinement algorithms, as it learns to mark elements for refinement based on the resulting error reduction.

In $2$ dimensions, we uniformly refine the initial coarse mesh $6$ times to create the reference mesh $\Omega^*$, and use the same amount of $6$ refinement steps per episode. 
We additionally evaluate a simpler task setup with $4$ refinements for the reference and~\gls{amr} algorithm to emulate the task complexity of existing work~\citep{yang2023reinforcement, fortunato2022multiscale, yang2023multi}.
In both cases, always refining all elements precisely reconstructs the reference mesh.
In $3$-dimensional domains, we uniformly refine three times, resulting in $8^3$ times more elements in $\Omega^*$ compared to $\Omega^0$ and employ $7$ longest edge bisection \gls{amr} steps, yielding refinements that are topologically different from the reference mesh.
The number of refinements was chosen to ensure that $\Omega^*$ is a sufficiently accurate proxy for the exact solution that allows for a challenging refinement task, while remaining computationally feasible for efficient experimentation.

\subsection{Training and Evaluation.}
\label{ssec:training}

We train all \gls{rlamr} methods on $100$ randomly generated systems of equations, including randomized domains and process conditions. 
This choice reduces the number of expensive reference meshes $\Omega^*$ that must be computed during training, and is further explored in~\ref{app_ssec:ablations}.
We evaluate each converged~\gls{rlamr} policy and the heuristics on $100$ randomly sampled but fixed evaluation \glspl{pde}, training each~\gls{rlamr} algorithm for $10$ random seeds.
The systems of equations used for evaluation are disjoint from those used for training.

For all methods, we experiment with a range of target mesh resolutions to produce meshes of different refinement levels, as detailed in~\ref{app_ssec:refinement_hyperparameters}. 
\ref{app_sec:experiment_details} provides additional details on the experimental setup and additionally discusses the required computational budget.
We train the~\gls{rlamr} methods using both \gls{ppo}~\citep{schulman2017proximal} with discrete actions as an on-policy algorithm and \gls{dqn}~\citep{mnih2013playing, mnih2015human} as an off-policy variant.

We evaluate mesh quality by calculating the squared error between the solution on this mesh and the solution on the fine-grained reference $\Omega^*$.
This squared error over the domain is numerically approximated over evaluations at the midpoints $p_{\Omega_m^*}$ of the reference mesh as
\begin{equation}
\label{eq:squared_integrated_difference}
\sum_{\Omega_m^*\in \Omega^*} \text{Volume}(\Omega^*_m)\left(u_{\Omega^*}(p_{\Omega_m^*})-u_{\Omega^t}(p_{\Omega_m^*})\right)^2\text{.}
\end{equation}
We normalize the resulting value by that of the initial mesh for comparability across \glspl{pde}, and call the resulting metric the remaining squared \textit{mesh error}.
This squared error metric captures the overall error of the solution on the refined mesh, while punishing outliers.
It differs from our optimization objective in Equation~\ref{eq:asmr_local_reward}, which instead rewards minimizing the maximum error of the solution on the refined mesh.
\ref{app_ssec:alternate_metrics} discusses and evaluates a mean error and an approximation of the maximum error as alternate metrics.
We list further algorithm and network hyperparameters in~\ref{app_ssec:general_hyperparameters}.

\subsection{RL-AMR Baselines.}
\label{ssec:rlamr_baselines}
We compare to~\gls{asmr}~\citep{freymuth2024swarm} and several recent non-stationary \gls{rlamr} methods~\citep{yang2023reinforcement, yang2023multi, foucart2023deep} that have been adapted to our application, focusing on stationary refinements.
Following \gls{asmr}, we use the rewards as described in the respective papers for each method, employing the absolute integrated difference in solution compared to the reference mesh, i.e., Equation~\ref{eq:squared_integrated_difference} without the square as an error estimate.

\textit{Single Agent}~\citep{yang2023reinforcement} uses the difference in error norm as a global reward and marks a single element for refinement in each step via a categorical action over the full mesh.
\textit{Sweep}~\citep{foucart2023deep} randomly samples a single mesh element and decides on its refinement based on local features and a global resource budget during training. 
The method uses the logarithm of the change in error between consecutive steps as a reward.
During inference, each timestep considers all mesh elements in parallel.
\gls{vdgn}~\citep{yang2023multi} employs value decomposition networks~\citep{sunehag2017value} to estimate a global Q-function as the sum of element-wise local Q-functions.
For a \gls{vdgn}-like \gls{ppo} variant, we instead analogously decompose the value function.
As our method is an extension of~\gls{asmr}~\citep{freymuth2024swarm}, this baselines largely follows our method. However, it instead uses the volume-scaled reward in Equation~\ref{eq:volume_reward}, omits the normalization factor in the agent mapping in Equation~\ref{eq:agent_mapping}, considers an infinite horizon~\gls{rl} setting, and uses less regularization in the neural network architecture. 

We use an \gls{mpn} policy for all~\gls{rlamr} methods except for \textit{Sweep}, which utilizes a simple \gls{mlp}.
The~\gls{rlamr} methods handle the trade-off between mesh resolution and solution accuracy slightly differently.
\textit{Single Agent} simply refines for a number of steps, \textit{Sweep} uses a target number of elements, and the other methods penalize each added element.
Here,~\gls{asmr} trains a new policy for each penalty, while only~\gls{method} and the \gls{vdgn}-like baseline use a conditional policy that can produce meshes of different resolutions during inference.
\ref{app_ssec:baseline_hyperparameters} lists all baseline-specific hyperparameters, while~\ref{app_ssec:refinement_hyperparameters} provides details on the mesh resolution parameters for all methods.

\subsection{Heuristic Baselines.}
\label{ssec:heuristic_baselines}
In addition to the~\gls{rlamr} methods above, we compare to a traditional error-based heuristics acting on a refinement threshold $\theta$~\citep{binev2004adaptive, bangerth2012algorithms, foucart2023deep}.
Given an error estimate $\text{err}(\Omega^t_i)$ for element $i$, the heuristic marks all elements for refinement for which 
$\text{err}(\Omega^t_i)>\theta\cdot\max_j \text{err}(\Omega^t_j)$.
We consider an \textit{Oracle Error Heuristic} that assumes knowledge about the fine-grained reference solution $\Omega^*$ and uses the absolute integrated difference to this solution as its error metric.
Similarly, we compare to a \textit{Maximum Oracle Error Heuristic}, for which the error estimate of Equation~\ref{eq:err_per_element} is used.
Both variants require the reference mesh $\Omega^*$ to estimate an error, which is expensive to compute and thus may be unavailable during inference.
We additionally consider the commonly used \gls{zz}, which instead uses a superconvergent patch recovery process for its error estimate~\citep{zienkiewicz1992superconvergent} and does not require access to $\Omega^*$.
As the recovery process averages over neighboring mesh elements, the \gls{zz} generally produces smooth error estimates. 
This property can lead to more coherent refinements when compared to the oracle heuristics, especially when considering the closing operations needed to ensure a conforming mesh after every refinement step.

All heuristics greedily refine based on local error estimates, in the sense that they are based purely on the current error rather than the potential reduction in error. Thus, they implicitly assume that refining elements with high error is the best way to minimize the error on the mesh.
This assumption does not always hold, as, e.g., elliptical~\glspl{pde} often feature globally propagating errors~\citep{strauss2007partial, foucart2023deep}.
The~\gls{rlamr} methods instead learn to optimize an objective function, which rewards the error reduction over time.
This difference between greedily marking elements with a high local error, and learning to anticipate the long-term error reduction for any given refinement allows the \gls{rlamr} methods to find more global strategies that take multiple refinement steps and non-local information of the mesh elements into account.

\subsection{Observation Graph.}
\label{ssec:observations}
As node features for the observation graph of Section~\ref{ssec:mpn_description}, we use the mean and standard deviation of the~\gls{fem} solution on the element's vertices, the element volume and the current environment timestep.
For~\gls{method} and the~\gls{vdgn}-like baseline, we also add the current element penality to condition the refinement process of the policy on a given target resolution.
Some tasks employ additional node features to encode their process conditions, which we describe in~\ref{app_sec:pde}.
We use the Euclidean distance between the element midpoints as the single edge feature.
Compared to using either a distance vector between element midpoints, or providing absolute positions, this ensures that the predicted sizing fields are invariant under Euclidean group operations~\citep{pfaff2020learning, bronstein2021geometric}.

\section{Results}
\label{sec:results}

\begin{figure}[ht]
    \centering
    \input{figures/marking_strategy_asmr/marking_strategy_asmr}
    \caption{
    Visualization of a full~\gls{method} rollout for a heat diffusion example, including the policy action that marks individual elements for refinement.
    \rebuttal{The mesh's background color denotes solution magnitude.}
    The odd columns show which elements the policy marks for refinement (designated by a green tick), while the even columns show the resulting refined meshes.
    \gls{method} produces a sequence of refinements that improves solution accuracy while using few total mesh elements. 
    }
    \label{fig:full_rollout_visualization}
\end{figure}

\subsection{Qualitative Results.}
\label{ssec:qualitative_results}

Figure \ref{fig:full_rollout_visualization} visualizes a full rollout of~\gls{method} on an instance of the heat diffusion task, including the per-element refinement markings after every policy step.
\rebuttal{
For this and other qualitative figures, the background color represents the magnitude of the problem-dependent scalar solution.
}
\gls{method} is trained to produce a sequence of refinements that improve the solution accuracy enough to justify the additional mesh elements, which here corresponds to more refinements along the path of the heat source. 
Figure~\ref{fig:element_penalties} shows final refined meshes for the same Poisson problem for a fixed~\gls{method} policy that is conditioned on different values of the element penalty $\alpha$.
Here, decreasing the element penalty results in more mesh elements and thus a more accurate solution. 
Our method is able to focus on the interesting parts of the mesh regardless of the final mesh resolution, thus providing a favorable trade-off between solution accuracy and computational cost.
Figure~\ref{fig:qualitative_all_tasks} provides final refinements of \gls{method} on exemplary systems of equations for all considered tasks.
The refinement strategy individually adapts to each task, choosing to refine elements that decrease the simulation error regardless of the underlying system of equations.
We further explore this error reduction property in Figure~\ref{fig:qualitative_error}.
The visualization shows that our method achieves a much more uniform distribution of errors across the mesh when compared to simple uniform refinement, while also acquiring a much smaller simulation error for the same number of mesh elements.

Appendix~\ref{app_sec:baseline_visualizations} visualizes all methods on the Stokes flow task to showcase the different refinement behaviors  for different target mesh resolution parameters.
Appendix~\ref{app_sec:asmr_visualizations} provides additional \gls{method} visualizations for the remaining tasks, including different camera angles for the $3$-dimensional variant of Poisson's equation that uses tetrahedral meshes.

\subsection{Quantitative Results.}
\label{ssec:quantitative_results}
Quantitatively, we measure the mesh quality using a Pareto plot of the number of elements and the normalized \rebuttal{squared} error of the mesh, \rebuttal{as defined in Equation~\ref{eq:squared_integrated_difference}}. 
We evaluate each \gls{rlamr} policy on $100$ evaluation environments per mesh resolution parameter. 
We then calculate the interquartile mean~\citep{agarwal2021deep} of the resulting $100$ used elements and normalized errors of the final produced meshes, and plot the resulting values.
This process is done for $10$ independently trained policy seeds and repeated across a range of mesh resolution parameters.
For methods that require a fixed target mesh resolution, we use $10$ different parameters, for the adaptive methods we instead use $20$ parameters, as detailed in~\ref{app_ssec:refinement_hyperparameters}.
We further provide a log-log quadratic regression over the aggregated results of each method as a general trend-line. 
To improve clarity and focus on representative behavior,  we omit data points with abnormally high element counts, which are sometimes produced as outliers by the~\gls{vdgn}-like baseline.

{
\captionsetup{skip=-0.01cm} %

\begin{figure}[ht!]
    \centering
    
    \begin{minipage}[b]{0.48\textwidth}
            \centering
            \includegraphics[width=\textwidth, height=6cm, keepaspectratio]
            {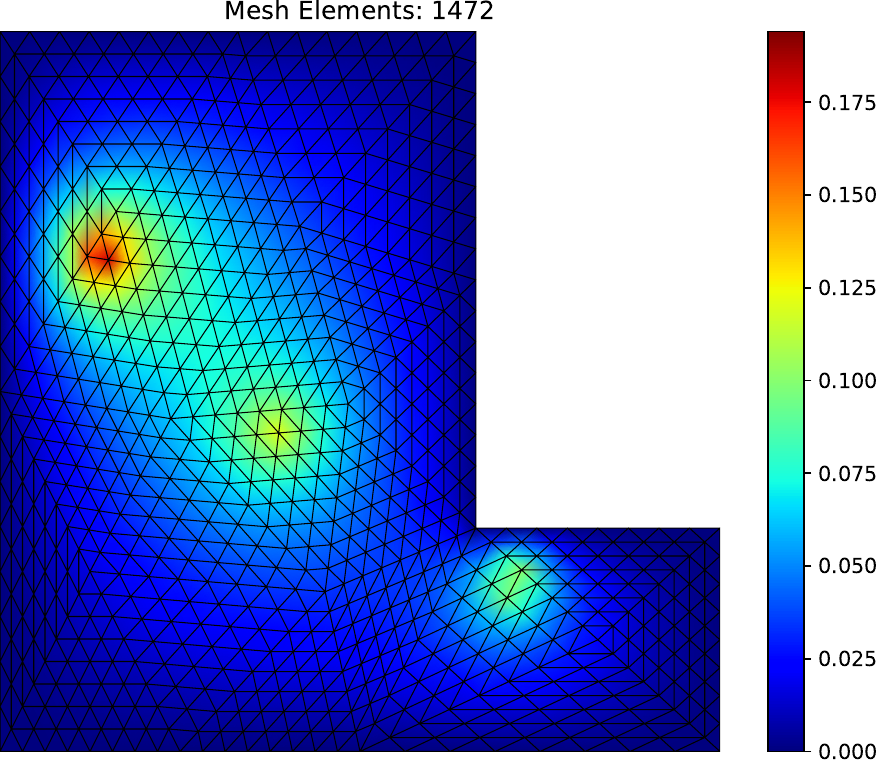}
            \caption*{Uniform solution}
    \end{minipage} \hfill %
    \begin{minipage}[b]{0.48\textwidth}
            \centering
            \includegraphics[width=\textwidth, height=6cm, keepaspectratio]
            {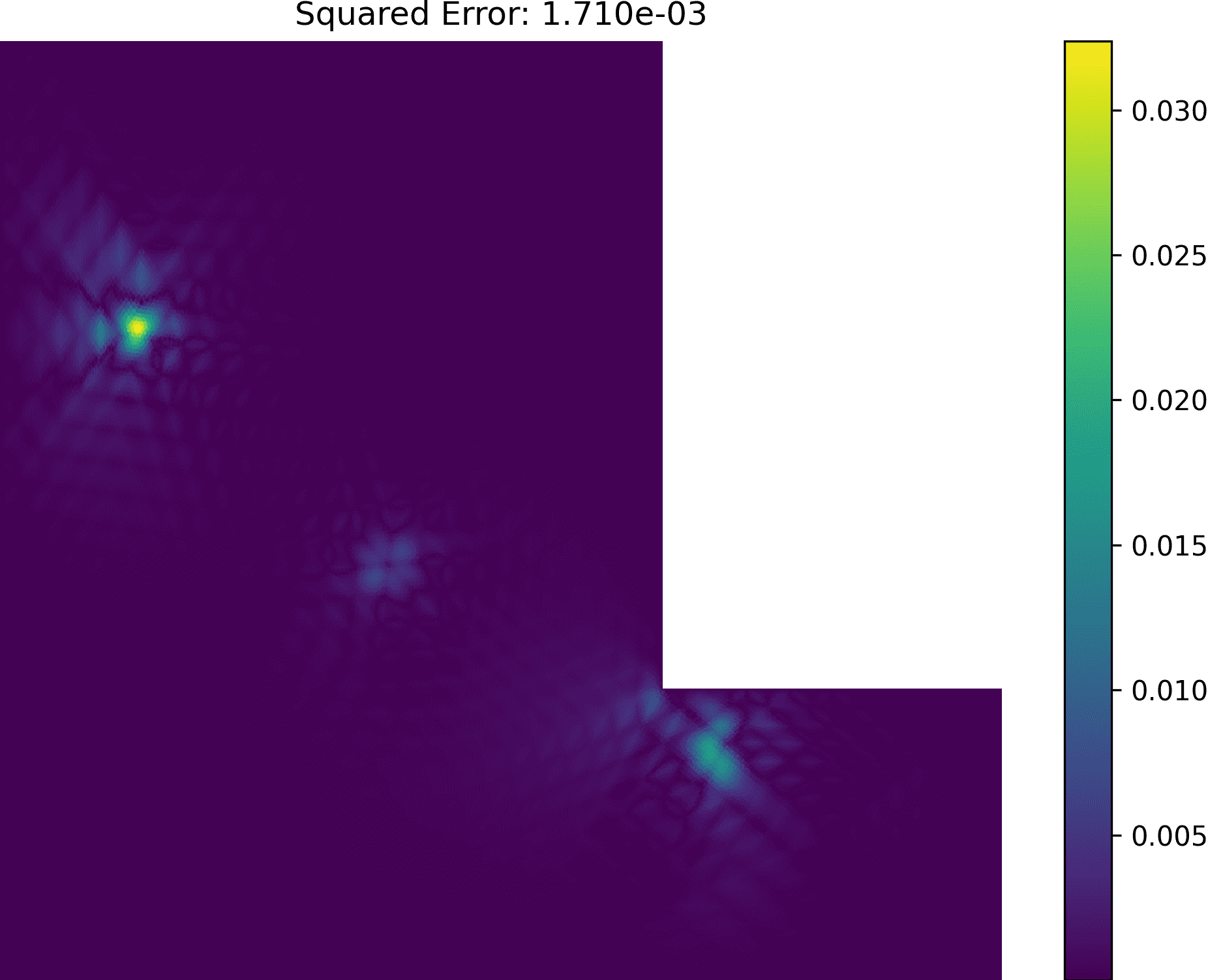}
            \caption*{Error of uniform mesh}
    \end{minipage}
    \vspace{0.5cm}

        \begin{minipage}[b]{0.48\textwidth}
            \centering
            \includegraphics[width=\textwidth, height=6cm, keepaspectratio]
            {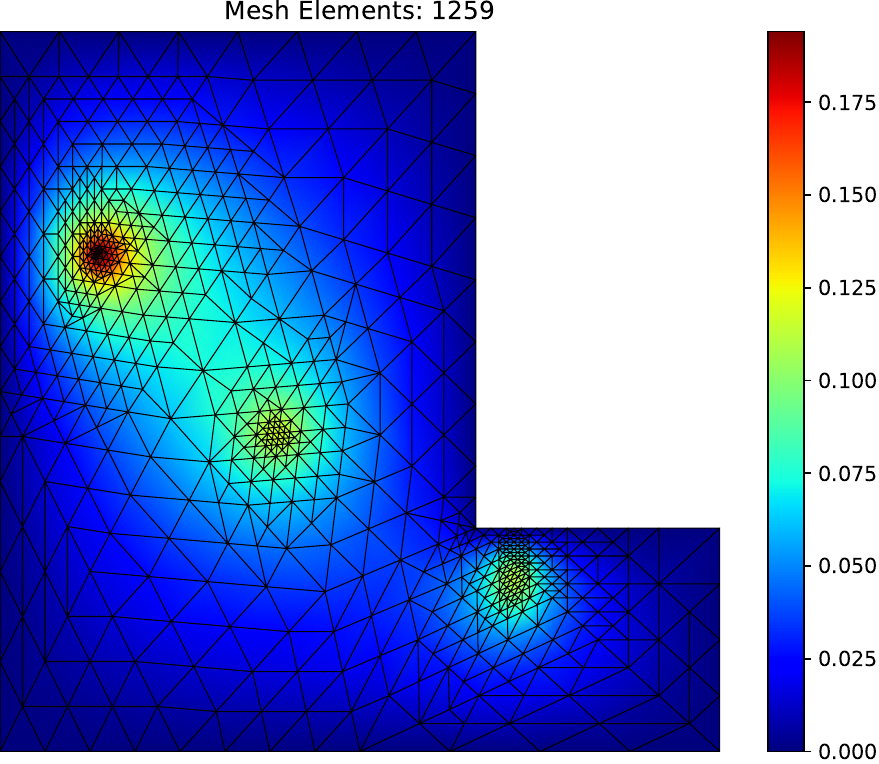}
            \caption*{\gls{method} solution}
    \end{minipage} \hfill %
    \begin{minipage}[b]{0.48\textwidth}
            \centering
            \includegraphics[width=\textwidth, height=6cm, keepaspectratio]
            {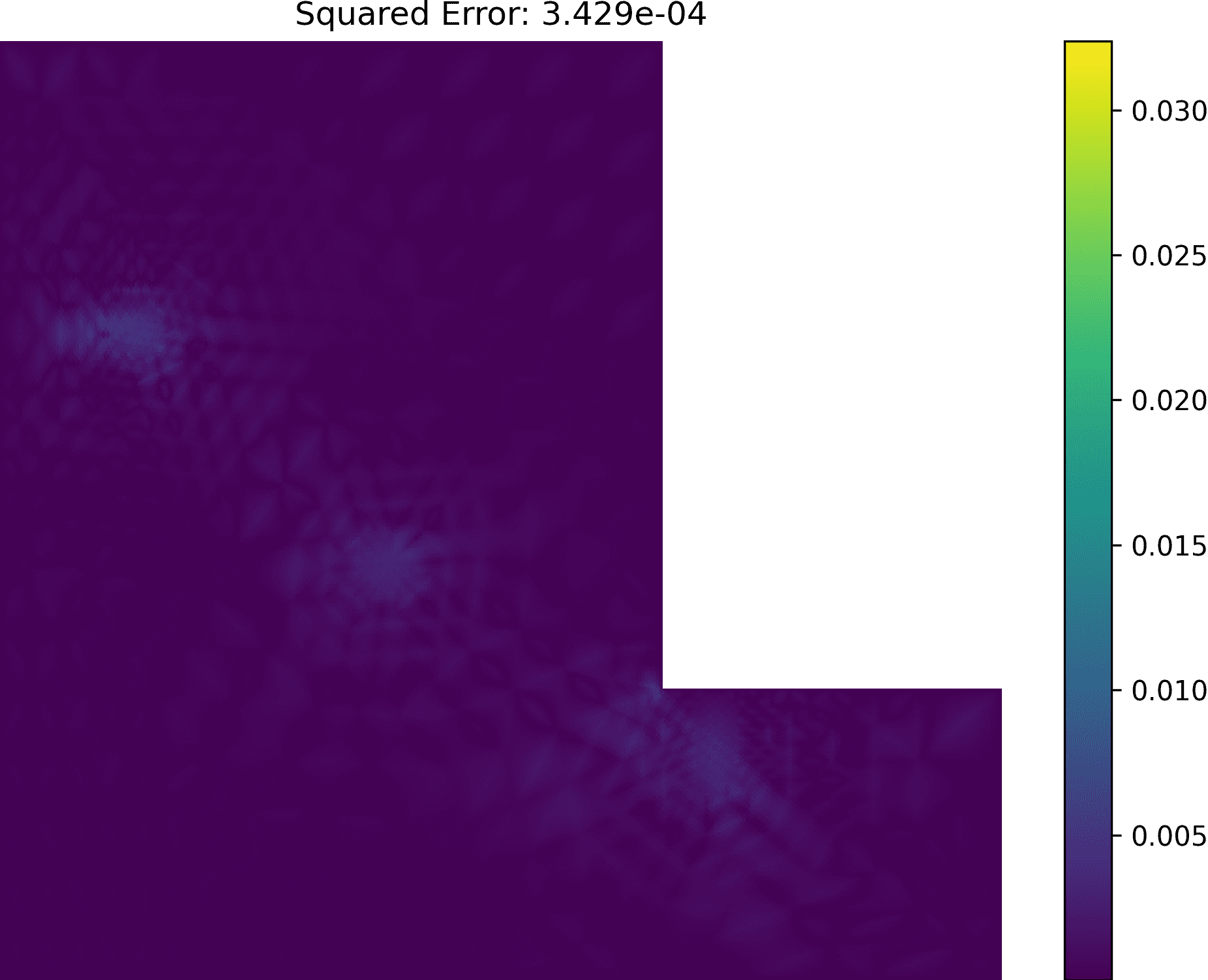}
            \caption*{Error for \gls{method} mesh}
    \end{minipage}
    \vspace{0.01\textwidth}%
    
    \caption{
        Comparison of exemplary~\gls{method} refinement with a uniform mesh on Poisson's equation. 
        (\textbf{Left}) Mesh and~\gls{pde} solution for a uniform mesh (\textbf{Top}) and our approach (\textbf{Bottom}). Both meshes have a similar number of total elements, but different spatial resolutions.
        (\textbf{Right}) Normalized squared difference in solution to the ground truth reference $\Omega^*$, evaluated at each reference element's midpoint. \gls{method} has less than a third of the error of the uniform mesh, and a much more even distribution of errors across the mesh.
    }
    \label{fig:qualitative_error}
\end{figure}
}

We experiment with \gls{ppo} as an on-policy algorithm and \gls{dqn} as an off-policy variant of the \gls{rl} backbone for all~\gls{rlamr} methods on the Stokes flow task in~\ref{app_ssec:ppo_vs_dqn}.
All~\gls{rlamr} methods, including \gls{method}, perform better with \gls{ppo}, indicating that an on-policy algorithm is favorable when dealing with action and observation spaces of varying size.
Further, we compare our \glspl{mpn} to the \gls{gat}-like \citep{velickovic2018graph} architecture proposed by \gls{vdgn}~\citep{yang2023multi} in~\ref{app_ssec:mpn_vs_gat}.
For both our method and the~\gls{vdgn}-like baseline, the~\gls{mpn} shows minor but consistent benefits in performance.
We thus use \gls{ppo} and \glspl{mpn} in all other experiments.

\ref{app_ssec:zz_error_ablation} compares the performance of the \gls{zz} \textit{Heuristic} when applied directly to the initial mesh to variants that instead uniformly refine the mesh once or twice before using the estimator.
We find that the method benefits from $2$ uniform initial refinements, and use this variant for all experiments. 
This tuning of the initial meshes in not required by the~\gls{rlamr} methods and prevents coarse refinements.

\begin{figure}[ht!]
    \centering
    \includegraphics{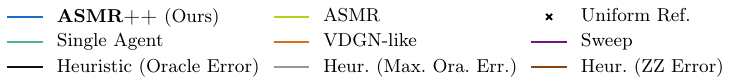}
    \includegraphics{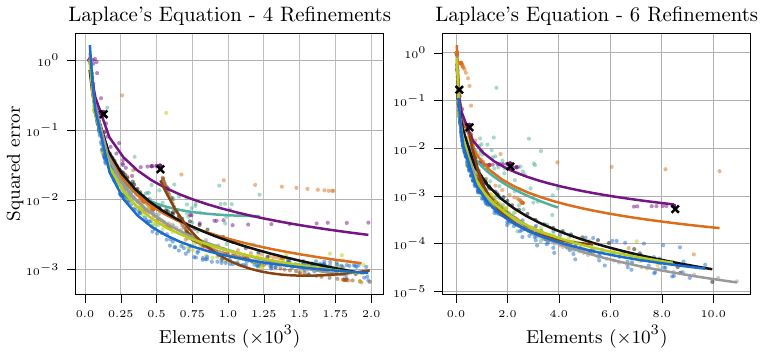}

    \caption{
    Pareto plot of normalized squared errors and number of final mesh elements on Laplace's equation.
    (\textbf{Left}) All \gls{rlamr} methods produce better-than-uniform meshes for shallow refinements and reference meshes with only $4$ refinement steps.
    (\textbf{Right}) When increasing the complexity to $6$ refinement steps, most~\gls{rlamr} methods exhibit high variance in mesh quality and sometimes fail to provide useful meshes.
    \gls{method} consistently produces high-quality meshes for both task complexities, performing on par with or better than the heuristics and providing a slight but consistent benefit over~\gls{asmr}.
    }
    \label{fig:quantitative_laplace}
\end{figure}

\begin{figure}
    \centering
    \includegraphics{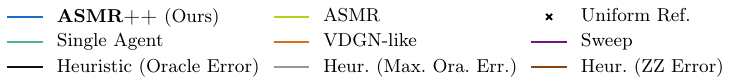}
    \includegraphics{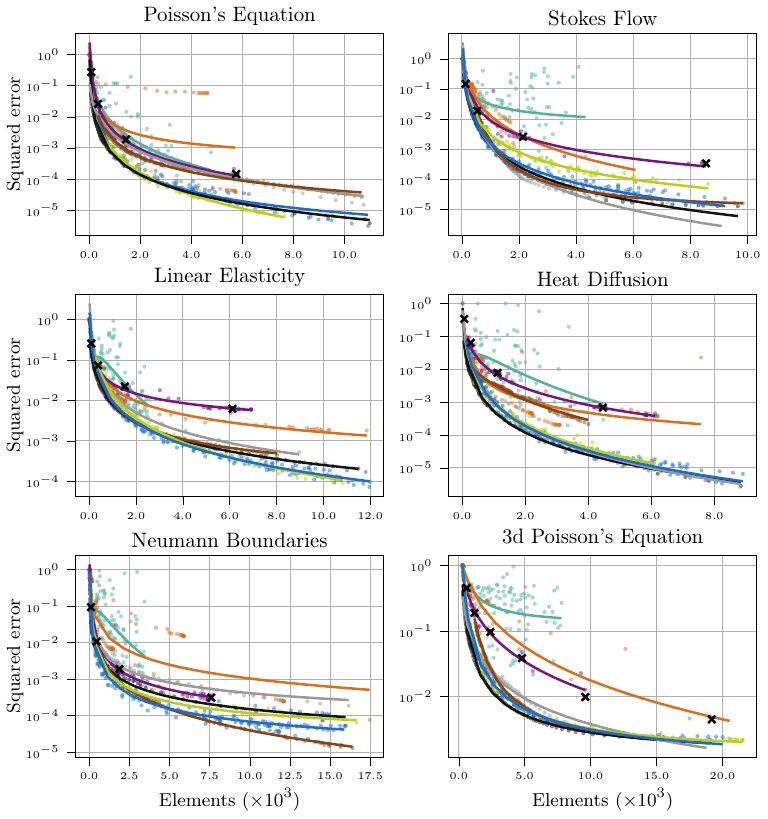}
    \caption{
        Pareto plot of normalized squared errors and number of final mesh elements across different tasks.
        On all tasks, the error decreases in a log-linear relation to the number of used elements.
        \gls{method} gracefully scales to meshes of more than ten thousand elements, significantly outperforming the learned \textit{Single Agent}, \textit{VDGN}-like and \textit{Sweep} in terms of consistency and mesh quality. 
        \gls{method} also slightly improves over~\gls{asmr} on most tasks and generally compares favorably to the \textit{Oracle}, \textit{Maximum Oracle} and \gls{zz} \textit{Heuristics}.
    }
    \label{fig:quantitative_others}
\end{figure}

Using these initial results, we compare the different approaches on Laplace's equation.
On the left of Figure~\ref{fig:quantitative_laplace}, we evaluate a simplified setup that uses $4$ refinement steps for both the reference mesh and the~\gls{amr} algorithms, except for \textit{Single Agent}, which instead refines for up to $100$ time steps. 
All methods clearly outperform uniform refinements. 
The~\gls{zz} \textit{Heuristic} performs well for larger meshes, but can not produce meshes with few elements as it starts with $2$ uniform initial refinements.

The right of Figure~\ref{fig:quantitative_laplace} increases the complexity to $6$ refinement steps, and $400$ time steps for \textit{Single Agent}.
Similarly, Figure~\ref{fig:quantitative_others} provides results on the remaining $6$ tasks.
Across all tasks, only \gls{asmr} and~\gls{method} can handle larger meshes while the other~\gls{rlamr} methods fail to provide better-than-uniform meshes or exhibit high variance in the quality of the proposed refinements.
Our method additionally outperforms all heuristics, including the \textit{Oracle Error} and \textit{Maximum Oracle Error Heuristics}, in several instances.
These results indicate that our optimization method facilitates learning non-local long-horizon refinement strategies that minimize the mesh error instead of simple targeting elements that currently have a high error.
\gls{method} further improves over~\gls{asmr} in most settings, especially in the complex Stokes flow task where it compares well to the different \textit{Heuristics}.
In the tasks where~\gls{method} does not perform better than~\gls{asmr}, both perform similar to or better than all heuristics, potentially indicating that both methods offer near-optimal refinements for these tasks.
On the $3$d Poisson's Equation, the different methods quickly converge to a relatively high final mesh error, likely due to the difference in mesh topology between the refined meshes and the uniform reference that is caused by the longest edge bisection refinement strategy.

\ref{app_ssec:alternate_metrics} presents additional results using a mean error and an adapted maximum error metric.
On both metrics, \gls{method} is generally competitive with or better than the \textit{Heuristics} and significantly outperforms the \textit{Single Agent}, \textit{Sweep} and \gls{vdgn}-like baselines.
Comparing the different metrics, the \textit{Maximum Oracle Error Heuristic} improves over the \textit{Oracle Error Heuristic} on the maximum error metric and vice versa.
Likewise,~\gls{asmr}, which uses a scaled variant of the mean error instead of the maximum objective of Equation~\ref{eq:asmr_local_reward}, outperforms~\gls{method} on half of the tasks in terms of the remaining mean error, while~\gls{method} is superior for every setup when evaluating the maximum error metric.

\subsection{Ablations and Additional Experiments.}
\label{ssec:additional_experiments}

We conduct several ablations and additional experiments on the challenging Stokes flow task to determine what makes our approach uniquely effective.
These experiments individually change a single aspect of~\gls{method} while leaving the rest of the approach unchanged.

\subsubsection{Reward.}

\begin{figure}[ht!]
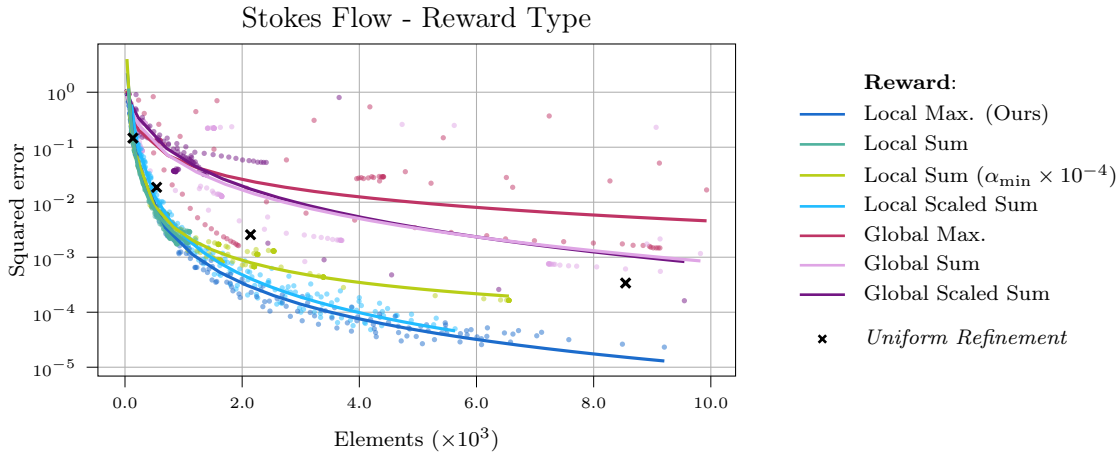

    \centering 
    \begin{minipage}{0.62\textwidth}
        \tikzsetnextfilename{abl_reward}
        \input{appendix/figures/quantitative/ablations/tikz/stokes_rewards__squared_error}
    \end{minipage}%
    \begin{minipage}{0.01\textwidth}
    ~
    \end{minipage}%
    \begin{minipage}{0.33\textwidth}
        \tikzsetnextfilename{abl_reward_legend}
        \begin{tikzpicture}
\tikzstyle{every node}'=[font=\scriptsize]
\input{tikz_colors}
\begin{axis}[%
                    hide axis,
                    xmin=10,
                    xmax=50,
                    ymin=0,
                    ymax=0.1,
                    legend style={
                        draw=white!15!black,
                        legend cell align=left,
                        legend columns=1,
                        legend style={
                            draw=none,
                            column sep=1ex,
                            line width=1pt
                        }
                    },
                    ]

\addlegendimage{empty legend}
\addlegendentry{\textbf{Reward}:}

\addlegendimage{royalblue28108204}
\addlegendentry{Local Max. (Ours)}

\addlegendimage{cadetblue80178158}
\addlegendentry{Local Sum}
\addlegendimage{yellowgreen18420623}
\addlegendentry{Local Sum ($\alpha_{\text{min}}\times10^{-4}$)}
\addlegendimage{deepskyblue33188255}
\addlegendentry{Local Scaled Sum}

\addlegendimage{indianred19152101}
\addlegendentry{Global Max.}
\addlegendimage{plum223165229}
\addlegendentry{Global Sum}
\addlegendimage{purple11522131}
\addlegendentry{Global Scaled Sum}

\addlegendimage{empty legend}
\addlegendentry{~}
\addlegendimage{only marks, mark=x, black}
\addlegendentry{\textit{Uniform Refinement}}
\end{axis}
\end{tikzpicture}    
    \end{minipage}
    \caption{
        Pareto plot of normalized squared errors and number of final mesh elements on the Stokes flow task for different reward functions for~\gls{method}. 
        In general, a using a local reward function, i.e., individual rewards for each agent, is crucial for the performance of the method.
        Here, optimizing the reduction in total error per mesh element leads to sub-optimal meshes for large numbers of elements, even when significantly decreasing the minimum sampled element penalty $\alpha_{\text{min}}$.
        Scaling this optimization with the inverse area of each mesh element as done by~\gls{asmr} offsets this issue, but performs worse than the simpler maximum reward of~\gls{method}. 
    }
    \label{app_fig:abl_reward}
\end{figure}

We compare the local maximum error reward of Equation~\ref{eq:asmr_local_reward} to different local and global variants.
For the global variants, we obtain a scalar reward function by averaging over the local reward functions. Here, we compute the return using Equation~\ref{eq:scalar_return}, i.e., without mapping between agents over time.
In addition to the maximum reward of Equation~\ref{eq:asmr_local_reward}, we compare to the volume-scaled reward proposed by~\gls{asmr} in Equation~\ref{eq:volume_reward}, and a variant that simple rewards a decrease in integrated error, i.e., Equation~\ref{eq:volume_reward} without the volume scaling term.

Figure~\ref{app_fig:abl_reward} shows the results.
Using any type of global reward function leads to unstable refinements, especially for large meshes, likely because the credit for each reward cannot be properly assigned to the large number of agents in the system.
For the local reward variants, the local maximum reward performs best, closely followed by the volume-scaled variant.
Simply integrating the error of each mesh element and rewarding a reduction in this integrated error performs poorly for finer meshes.
This result is likely the case because smaller elements generally contain less integration points of the reference mesh, and thus have less integrated error to reduce. 
Optimizing this reward produces meshes that have an even distribution of integrated element errors, rather than refinements that effectively decrease the average error on the mesh.

\subsubsection{Return and Agent Mapping.}

\begin{figure}[ht!]
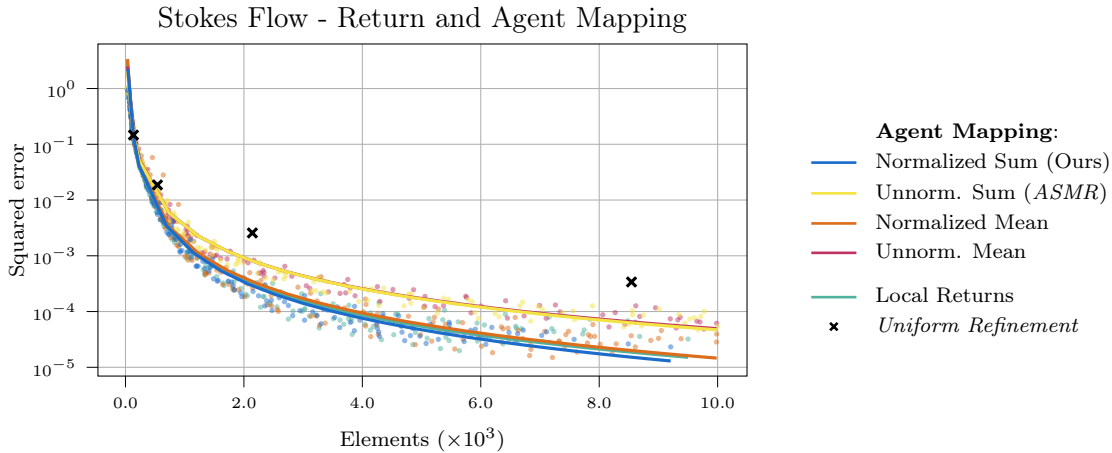

    \centering 
    \begin{minipage}{0.62\textwidth}
        \tikzsetnextfilename{abl_return}
        \input{appendix/figures/quantitative/ablations/tikz/stokes_return_mapping__squared_error}
    \end{minipage}%
    \begin{minipage}{0.01\textwidth}
    ~
    \end{minipage}%
    \begin{minipage}{0.33\textwidth}
        \tikzsetnextfilename{abl_return_legend}
        \begin{tikzpicture}
\tikzstyle{every node}'=[font=\scriptsize]
\input{tikz_colors}
\begin{axis}[%
                    hide axis,
                    xmin=10,
                    xmax=50,
                    ymin=0,
                    ymax=0.1,
                    legend style={
                        draw=white!15!black,
                        legend cell align=left,
                        legend columns=1,
                        legend style={
                            draw=none,
                            column sep=1ex,
                            line width=1pt
                        }
                    },
                    ]

\addlegendimage{empty legend}
\addlegendentry{\textbf{Agent Mapping}:}
\addlegendimage{royalblue28108204}
\addlegendentry{Normalized Sum (Ours)}
\addlegendimage{sandybrown24422769}
\addlegendentry{Unnorm. Sum (\textit{ASMR})}
\addlegendimage{chocolate21910927}
\addlegendentry{Normalized Mean}
\addlegendimage{indianred19152101}
\addlegendentry{Unnorm. Mean}

\addlegendimage{empty legend}
\addlegendentry{~}
\addlegendimage{cadetblue80178158}
\addlegendentry{Local Returns}
\addlegendimage{only marks, mark=x, black}
\addlegendentry{\textit{Uniform Refinement}}
\end{axis}
\end{tikzpicture}    
    \end{minipage}
    \caption{
        Pareto plot of normalized squared errors and number of final mesh elements on the Stokes flow task for different variants of the agent mapping $\phi$ and local returns of~\gls{method}.
        The normalization factor in the mapping of Equation~\ref{eq:agent_mapping} greatly improves performance.
        Using a sum instead of a mean mapping additionally improves performance.
        Adding a global term to the returns, as done in Equation~\ref{eq:half_half_return} further improves performance, indicating that a partially global objective improves global decision making of the individual agents.
    }
    \label{app_fig:abl_return}
\end{figure}

Equation~\ref{eq:agent_mapping} enables the computation of a TD error across agents at successive time steps by mapping each agent at time step $t$ to every agent it creates at time step $t+1$. 
In this setup, each spawned agent is fully attributed to its predecessor, meaning the originating agent is considered wholly responsible for any agent it generates. 
Compared to~\gls{asmr}, we additionally apply a regularization to this mapping through the ratio $\Omega^t/\Omega^{t+1}$.
An alternative approach would be to instead limit the total responsibility per agent to $1$, achieved by averaging the mapping as
$$
\phi'^t_{ij} :=\frac{\mathbb{I}(\Omega^{t+1}_j\subseteq \Omega^t_i)}{\sum_{j'}\mathbb{I}(\Omega^{t+1}_{j'}\subseteq \Omega^t_i)}\text{.}
$$

We assess these two methods in Figure~\ref{app_fig:abl_return}, finding that the normalization factor $\frac{\Omega^t}{\Omega^{t+1}}$ significantly boosts performance, likely due to its role as a regularizer during training. 
Furthermore, using a sum mapping, i.e., assigning each new agent a weight of $1$ from its creator, proves more effective than the mean mapping, which averages the total mapping weights per old agent.
We additionally explore optimizing a return that omits the global term in Equation~\ref{eq:half_half_return}, i.e., that only optimizes the local return of each agent. 
Here, we find that the partially global objective improves performance, likely because it aids in global decision making.

\subsubsection{Target Mesh Resolutions.}
Each \gls{rlamr} method uses a single parameter to control the number of target elements of the final refined mesh. 
\gls{method} and \gls{vdgn} condition their policy on an element penalty $\alpha$, while \gls{asmr} specifies a fixed element penalty as a hyperparameter. 
Similarly, \textit{Sweep} uses a fixed budget $N_{\text{max}}$ per policy, and \textit{Single Agent} trains on a specified number of rollout steps $T$.
\rebuttal{
Figure~\ref{app_fig:abl_element_penalty_range} evaluates the impact of these parameters for \gls{method} and the \gls{vdgn}-like baseline.
\gls{method} remains stable for a wide range of penalty parameters.
The penalty can be tuned by identifying the order of magnitude that corresponds to the desired number of mesh elements. 
This process allows users to easily control the trade-off between solution accuracy and computational cost without requiring sensitive hyperparameter tuning.
}
Additional details are provided in~\ref{app_ssec:refinement_hyperparameters} and Table~\ref{app_tab:resolutions_rl}.

\begin{figure}[ht!]
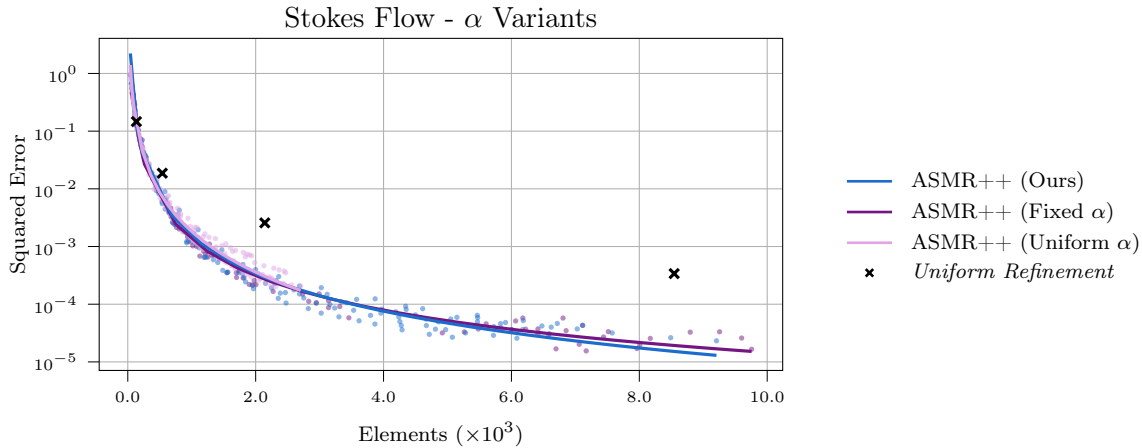

    \begin{minipage}{0.62\textwidth}
        \tikzsetnextfilename{abl_element_penalty_variants}
        \input{appendix/figures/quantitative/ablations/tikz/stokes_element_penalty_sampling__squared_error} 
    \end{minipage}%
    \begin{minipage}{0.01\textwidth}
    ~
    \end{minipage}%
    \begin{minipage}{0.33\textwidth}
        \tikzsetnextfilename{abl_element_penalty_variants_legend}
        \begin{tikzpicture}
\tikzstyle{every node}'=[font=\scriptsize]
\input{tikz_colors}
\begin{axis}[%
                    hide axis,
                    xmin=10,
                    xmax=50,
                    ymin=0,
                    ymax=0.1,
                    legend style={
                        draw=white!15!black,
                        legend cell align=left,
                        legend columns=1,
                        legend style={
                            draw=none,
                            column sep=1ex,
                            line width=1pt
                        }
                    },
                    ]

\addlegendimage{royalblue28108204}
\addlegendentry{ASMR++ (Ours)}

\addlegendimage{purple11522131}
\addlegendentry{ASMR++ (Fixed $\alpha$)}
\addlegendimage{plum223165229}
\addlegendentry{ASMR++ (Uniform $\alpha$)}
\addlegendimage{only marks, mark=x, black}
\addlegendentry{\textit{Uniform Refinement}}

\end{axis}
\end{tikzpicture}   
    \end{minipage}
     
    \caption{
        Pareto plot of normalized squared errors and number of final mesh elements on the Stokes flow task for different configurations of the element penalty $\alpha$ for~\gls{method}.
        Training a new policy for each value of $\alpha$, as done by~\gls{asmr}~\citep{freymuth2024swarm}, does not significantly improve performance, which indicates the utility and benefit of training a conditional policy on a range of $\alpha$ values. 
        Sampling $\alpha$ uniformly instead of log-uniformly during training leads to meshes with significantly fewer elements, as those values that would create large meshes are sampled less often.
    }
    \label{app_fig:abl_element_penalty_variants}
    
\end{figure}

Figure~\ref{app_fig:abl_element_penalty_variants} compares training~\gls{method} on a fixed element penalty, as done in~\gls{asmr}~\citep{freymuth2024swarm}, to training on an adaptive penalty that the policy is conditioned on for each rollout. 
It additionally evaluates sampling the penalty uniformly instead of log-uniformly during training.
Interestingly, training~\gls{method} on a fixed element penalty does not improve performance, suggesting that the~\gls{method} policy is able to learn to condition on a concrete $\alpha$ value.
These results imply that~\gls{method} can successfully learn optimal tradeoffs between element cost and error reduction and apply it during inference. 
Thus, we can train a single policy for a range of mesh granularities, omitting the need for~\gls{asmr}'s expensive re-training for each target granularity.
If we train on uniformly instead of log-uniformly sampled values of $\alpha$, the method predominantly provides meshes with low numbers of elements, likely because there is a logarithmic relationship between the element penalty and the number of mesh elements.

\begin{figure}[ht!]
    \centering
    \includegraphics{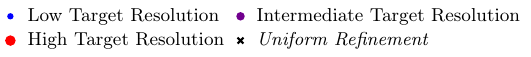}
    \includegraphics{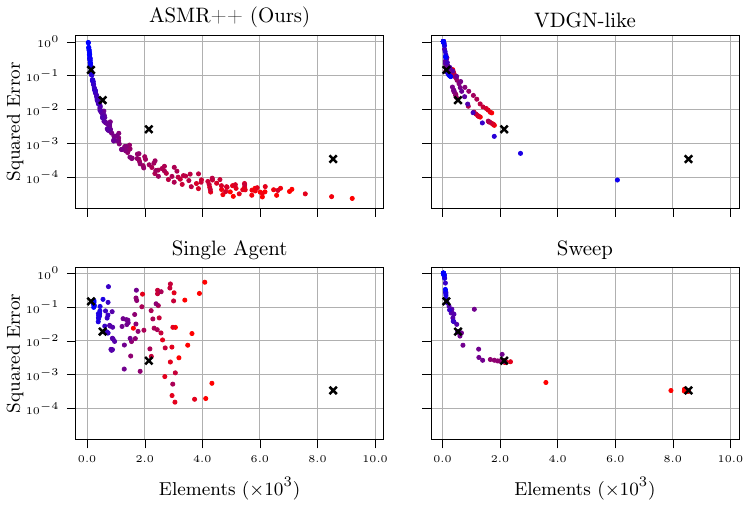}

    \caption{
        Pareto plot of normalized squared errors and number of final mesh elements for the Stokes flow task for all \gls{rlamr} methods.
        Each plot represents a single method for different evaluations of its respective mesh resolution parameters, as detailed in Appendix~\ref{app_ssec:refinement_hyperparameters} and Table~\ref{app_tab:resolutions_rl}.
        Small blue dots indicate meshes target with few elements, while large red dots correspond to finer target meshes.
        While the different \gls{rlamr} baselines provide poor-performing or inconsistent policies,~\gls{method} yields accurate refinements and consistent numbers of final mesh elements when the policy is conditioned on any given target resolution.
    }
    \label{fig:target_mesh_resolutions}
\end{figure}

Figure \ref{fig:target_mesh_resolutions} visualizes the methods' performance for different target resolutions.
Here, \gls{method} provides meshes with consistent numbers of elements for a given target resolution, while the other \gls{rlamr} baselines produce poorly refined meshes or inconsistent refinement granularities over target resolutions.

\subsubsection{Additional Ablations.}
\ref{app_ssec:ablations} provides additional ablations over changes in the network architecture, varying numbers of training~\glspl{pde} and different scales of the element penalty parameter in the reward.
The results indicate that as few as $10$ training~\glspl{pde} are sufficient to provide accurate refinements, with performance increasing for up to $100$~\glspl{pde}. 
In terms of architecture, edge dropout~\citep{rong2019dropedge} of $0.1$, i.e., randomly omitting $10\,\%$ of edges in the mesh topology during training, slightly improves performance.
Linearly increasing or decreasing the range of the element penalty $\alpha$ leads to meshes with fewer or more elements, respectively, yet~\gls{method} always provides stable and efficient refinements for the respective element penalty ranges.

\subsection{Runtime Comparison.}
\label{ssec:runtime}

\begin{figure}[ht!]
    \centering
    \includegraphics{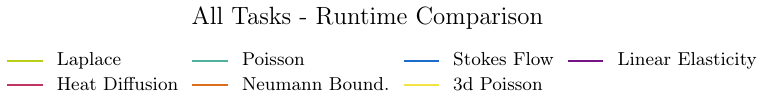}
    \includegraphics{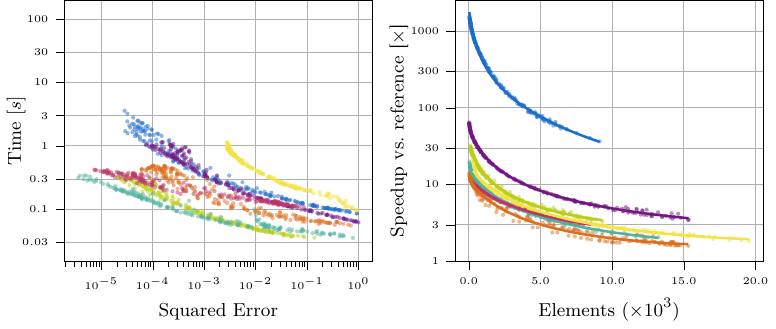}
    \caption{
    Wallclock-time evaluation of~\gls{method} for all tasks.
    Each color corresponds to a different task.
    (\textbf{Left)} Runtime of~\gls{method} and the squared error when compared to the uniform solution $\Omega^*$, which has an error of $0$ by definition.
    (\textbf{Right}) Speed-up of~\gls{method} compared to calculating the uniform solution $\Omega^*$, which contains roughly $10^5$ elements, with concrete numbers varying depending on the domain.
    Our approach is able to quickly produce high-quality meshes, achieving speedups of $1$ to $2$ orders of magnitude, depending on the task and target resolution of the mesh.
    }
    \label{app_fig:runtime_comparison}
\end{figure}

Figure \ref{app_fig:runtime_comparison} compares the runtime of \rebuttal{\gls{method} with directly} computing the uniform reference $\Omega^*$ across all tasks. 
Our method's timing includes the creation of the initial mesh, and iteratively solving the system of equations on the mesh, creating an observation graph, querying the policy for a refinement strategy, and finally refining a total of $6$ times.
For the uniform mesh, we simply measure the time it takes to refine the initial mesh $6$ times and to subsequently solve the problem on the resulting mesh.
We use a single 8-Core AMD Ryzen 7 3700X Processor for all measurements.

Our approach is always significantly faster than computing the fine-grained reference solution $\Omega^*$ despite the computational overhead of computing the observation graph and executing the policy.
Notably, higher squared error tolerances lead to reduced computation time, showing that \gls{method} can adapt to task-specific computational budgets.
The Stokes flow task uses comparatively expensive $P_2/P_1$ Taylor-Hood-elements~\citep{john2016finite}, causing our method to be more than $30$ times faster than $\Omega^*$ even for highly refined and accurate final meshes.

\rebuttal{
\gls{asmr} uses the same inference procedure as \gls{method} and thus shows the same general runtime behavior as \gls{method} when controlling for network size.}
Similarly, the other~\gls{rlamr} methods also feature an iterative refinement procedure and thus comparable runtime, but produce meshes of worse quality\rebuttal{ than~\gls{method}.}
In contrast, the \textit{Oracle Error Heuristic} and \textit{Maximum Oracle Error Heuristic} produce high-quality meshes, but require computing $\Omega^*$ for an error estimate. As such, these heuristics are impractical in terms of runtime, as they depend on the fine-grained solution that~\gls{amr} aims to approximate.

\begin{figure}[ht!]
    \centering
    \includegraphics{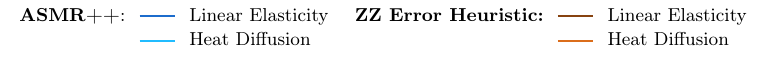}
    \includegraphics{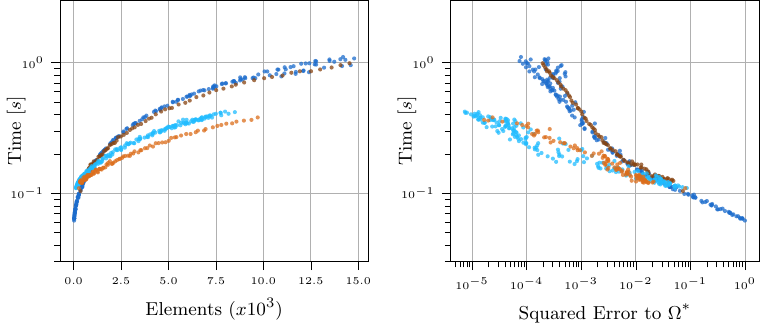}
    \caption{
        Wall-clock time comparison of~\gls{method} and the~\gls{zz} \textit{Heuristic} for the Linear Elasticity and Heat Diffusion tasks.
        For the~\gls{zz} \textit{Heuristic}, we measure the time required to solve the system of equations, execute the heuristic, and refine the mesh.
        For~\gls{method}, the timing also accounts for generating the observation graph and executing the trained policy.
        (\textbf{Left}) \gls{method} is slower for the same number of mesh elements due to additional computational steps.
        (\textbf{Right}) For a given target error, \gls{method} achieves comparable or better runtime on these tasks by requiring fewer mesh elements, thanks to its higher refinement quality.   
    }
    \label{fig:runtime_asmr_vs_zz}
\end{figure}

The~\gls{zz} \textit{Heuristic}, which relies on element gradient information, is comparatively fast. 
Figure~\ref{fig:runtime_asmr_vs_zz} compares it to~\gls{method} for both a fixed number of elements and for achieving a target error. 
While~\gls{method} is slightly slower per element due to the additional cost of generating an observation graph and executing a trained policy, its superior refinement quality leads to comparable or better runtimes for achieving a given target error on the Linear Elasticity and Heat Diffusion tasks. 
Additionally, the~\gls{zz} \textit{Heuristic} relies on initial mesh tuning (compare Figure \ref{app_fig:zz_error_ablation}) and exhibits significant variability in solution quality across tasks. 
While it can be faster than~\gls{method} on tasks where its heuristic happens to align well, it lacks the principled, error-focused optimization objective provided by~\gls{method} and its underlying~\gls{rl} framework.
Prior work~\citep{yang2023reinforcement} has also demonstrated that even simpler~\gls{rl}-based~\gls{amr} approaches outperform it on various problem classes, particularly those involving dynamic or time-varying behavior.

\subsection{Generalization Capabilities.}
\label{ssec:generalization}
{
\captionsetup{skip=-0.01cm} %

\begin{figure}
    \centering
    \begin{minipage}[b]{0.75\textwidth}
            \centering
            \includegraphics[width=\textwidth]
            {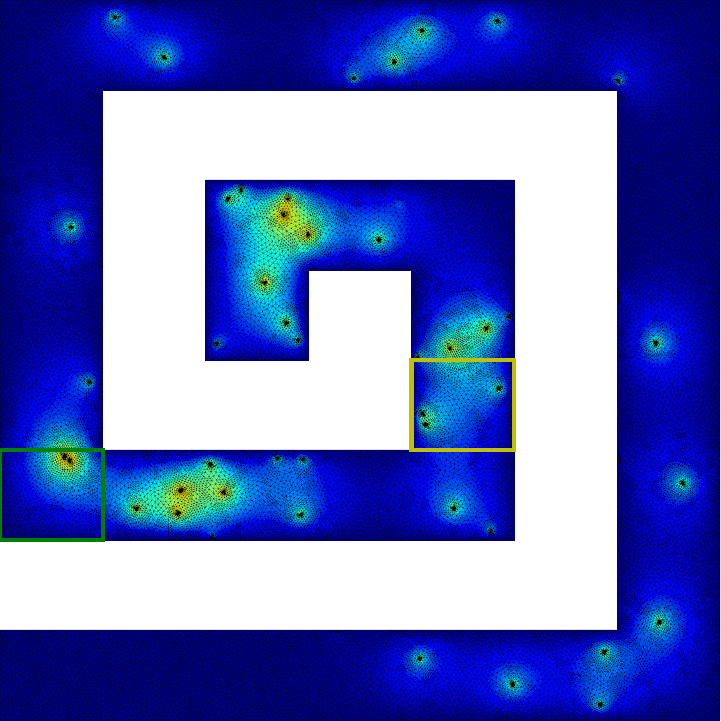}
            \caption*{Full mesh}
    \end{minipage}
    
    \vspace{0.5cm}
    \begin{minipage}[b]{0.45\textwidth}
            \centering
            \includegraphics[width=\textwidth]
            {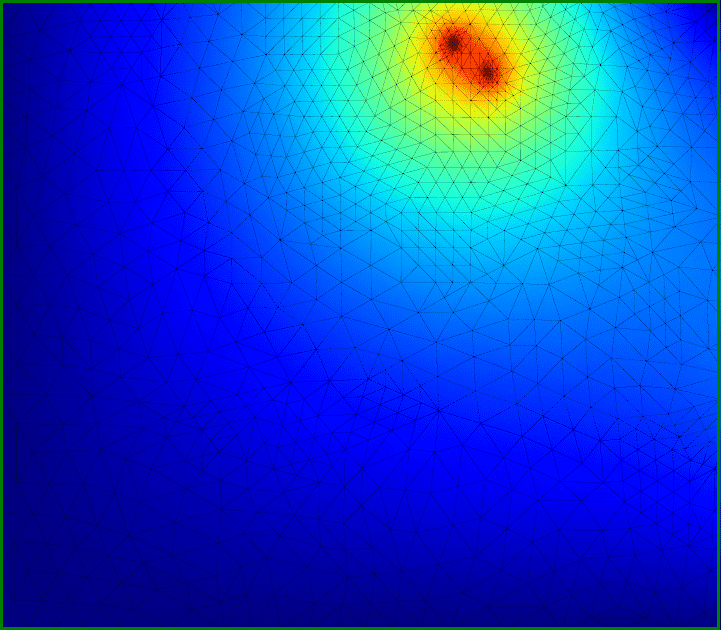}
            \caption*{Left zoom}
    \end{minipage} \hfill %
    \begin{minipage}[b]{0.45\textwidth}
            \centering
            \includegraphics[width=\textwidth]
            {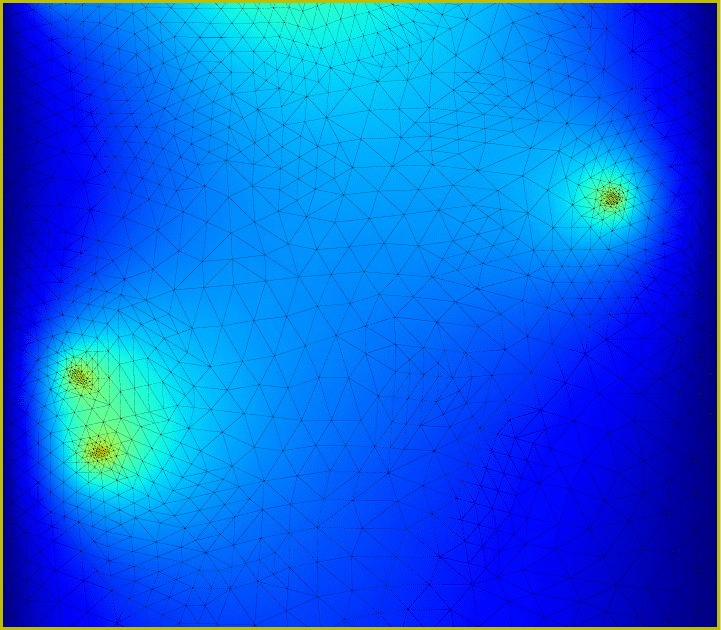}
            \caption*{Right zoom}
    \end{minipage}
    \vspace{0.01\textwidth}%
    
    \caption{
        (\textbf{Top}): Visualization of a final~\gls{method} mesh \rebuttal{and corresponding solution magnitude} on a $20\times20$ spiral domain. 
        The mesh consists of $52\,223$ elements and the full refinement procedure takes about $12.2$ seconds on a regular CPU.
        In contrast, computing the uniform refinement $\Omega^*$ takes roughly $20$ minutes. 
        (\textbf{Bottom}): Close-ups of the full mesh.
    }
    \label{fig:size20_generalization}
\end{figure}
}

\begin{figure}[ht!]
    \centering
        \includegraphics{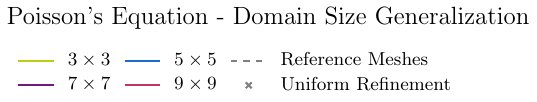}
        \includegraphics{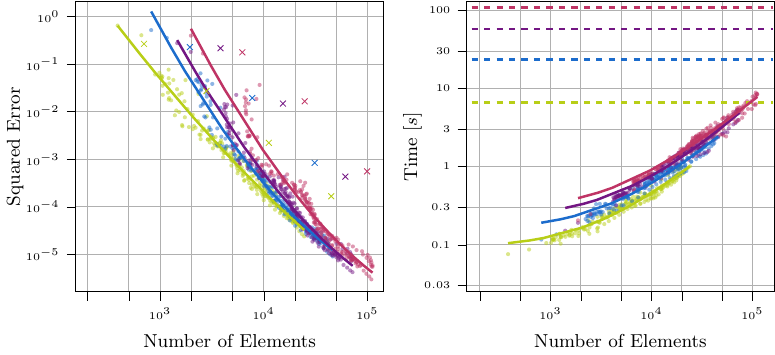}

    \caption{
        \gls{method} mesh error and inference speed over different sizes $N\times N$ of generalization domains after training on augmented $1\times 1$ domains.
        Each evaluated domain size is denoted by a different color.
        (\textbf{Left}) Pareto plot of squared errors and number of final mesh elements for \gls{method} and uniform refinements.
        Our method efficiently up-scales to larger domains during inference.
        (\textbf{Right}) Wallclock-time in seconds of~\gls{method} for different numbers of elements compared to the uniform reference $\Omega^*$ (dashed lines, color corresponding to domain size).
        While evaluating the uniform reference quickly becomes expensive,~\gls{method} provides efficient and comparatively cheap refinements even for larger domains.
    }
    \label{app_fig:poisson_generalization_runtime}
\end{figure}

We further experiment with the generalization capabilities of~\gls{method} to unseen and larger domains during inference.
\rebuttal{Appendix}~\ref{app_ssec:poisson_generalizing_1x1} shows the generalization capabilities of our approach when evaluated for different domains and load functions for Poisson's equation, indicating that it strongly generalizes to arbitrary domains and process conditions during inference.

\subsubsection{Generalization to Larger Domains.}
We additionally train on a variant of Poisson's equation that may sample load functions modes outside of the considered domain and randomizes solution values on the domain's boundaries.
\rebuttal{Appendix~\ref{app_ssec:domain_size_generalization} provides training details.}
These changes effectively modify the training~\glspl{pde} to mimic small patches of a larger mesh, acting as a data augmentation technique that facilitates up-scaling to larger domains and meshes during inference. 
An exemplary refinement on a $20\times20$ from a policy trained on $1\times1$ domains can be seen in Figure~\ref{fig:size20_generalization}.
The refined mesh has more than $50\,000$ elements, which is significantly larger than any meshes seen during training, yet accurately captures the interesting parts of the solution.
The full refinement procedure is roughly $100$ times faster than solving the fine-grained reference $\Omega^*$.

Figure~\ref{app_fig:poisson_generalization_runtime} shows the mesh error and inference speed for~\gls{method} policies trained on the augmented setup for Poisson's equation when evaluated on domains of different sizes.
For the $9\times9$ domain we used $10$ instead of $100$ evaluation~\glspl{pde} due to their increased runtime and memory requirements.
While the number of elements for uniform refinements is linear in the domain's volume, meshes created by our method grow less quickly to achieve the same error threshold.
These results suggest that there are fewer elements with a significant error in larger domains.
Accordingly, evaluating the reference mesh $\Omega^*$ quickly becomes expensive as the domain size gets larger, while~\gls{method} provides fast and efficient refinements.
For large meshes, our method is about $100$ times faster than evaluating $\Omega^*$ while maintaining a mesh error of less than $0.001$. 
\rebuttal{Appendix \ref{app_ssec:domain_size_generalization} shows} additional visualizations for adapted training~\gls{pde} and final refined meshes for different domain sizes.

The generalization capabilities of our method are likely a result of the agent-wise optimization and the utilized~\gls{mpn}, which facilitate refinement strategies that are largely based on local element neighborhoods.
Especially the ability of~\gls{method} to efficiently up-scale to larger meshes during inference opens up promising applications for practical engineering applications.
A policy can be trained on small and cheap environments, and then used in larger and more challenging setups during inference.

\subsubsection{Generalization to Different Materials.}

\begin{figure}
    \centering
    \includegraphics{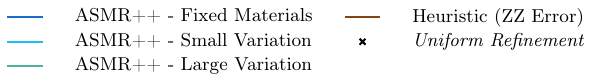}
    \includegraphics{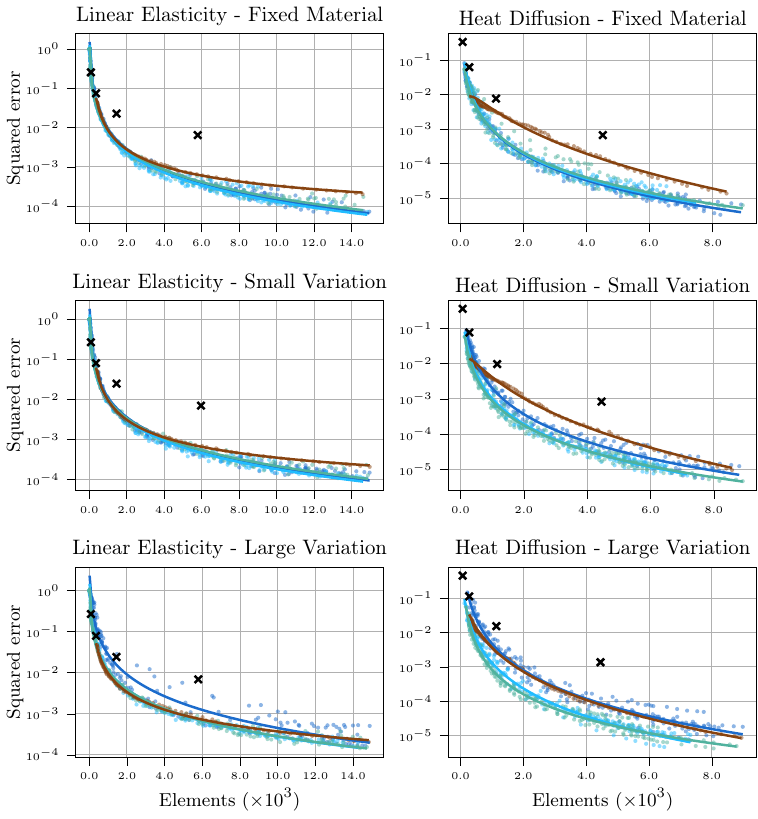}
    \caption{
         Pareto plot of normalized squared errors and number of final mesh elements for the (\textbf{Left}) Linear Elasticity and (\textbf{Right}) Heat Diffusion tasks. 
         Each sub-plot compares~\gls{method} trained on different material parameters and the~\gls{zz} \textit{Heuristic} on one set of evaluation environments.
         More precisely, we evaluate (\textbf{Top}) fixed parameters, (\textbf{Middle}) small variations and (\textbf{Bottom}) large variations in parameters. 
        While~\gls{method} trained on fixed parameters struggles to generalize to unseen parameter ranges during evaluation, training on a modest range of parameters enables generalization to unseen setups while maintaining refinement accuracy on the previously seen parameter ranges.
    }
    \label{fig:material_generalization}
\end{figure}

An essential aspect of~\gls{amr} strategies is their ability to generalize across a wide range of systems of equations. To evaluate this for~\gls{asmr}, we introduce varying material parameters to the Heat Diffusion and Linear Elasticity tasks. We define three levels of variability for both tasks, namely \textit{fixed materials} that keep the constant parameters of the main experiments, \textit{small variation} were the material parameters are varied within a narrow range, and \textit{large variation} where the parameters span a wide range.  
Table~\ref{tab:material_variants} summarizes the specific parameter ranges for each task and variation level. 
\begin{table}[h]
\centering
\caption{
Parameter ranges for the material generalization experiments by task. Parameter units are specified in parentheses. Meshes are normalized to unit length, so physical units correspond to this scale.}
\label{tab:material_variants}
    \begin{tabular}{llcc}
    \toprule
    \textbf{Task} & Parameter (Unit) & \textbf{Level} & \textbf{Value}  \\
    \midrule
    Heat Diffusion & Diffusivity ($m^2/s$) & Fixed materials & $1.0$e$-3$\\
                   & & Small variation & $[1.0$e$-4, 1.0$e$-2]$\\
                   &  & Large variation & $[1.0$e$-5, 1.0$e$-1]$\\
    \midrule
    Linear Elasticity & Poisson Ratio & Fixed materials &  $0.3$ \\
                      & & Small variation & $[0.2, 0.4]$ \\
                      & & Large variation & $[0.0, 0.499]$ \\
                      & Young's Modulus (GPa) & Fixed materials & $1.0$ \\
                      & & Small variation & $[0.3, 3.0]$ \\
                      & & Large variation & $[0.1, 10.0]$ \\
    \bottomrule
    \end{tabular}
\end{table}

For training and evaluation, we follow the methodology and setup of the main experiments. 
Material parameters are sampled log-uniformly for diffusivity and Young's modulus, and uniformly for the Poisson ratio. 
The log of diffusivity is added as an input feature to the heat diffusion~\gls{method} policy, while the log of Young's modulus and the Poisson ratio are added for linear elasticity.
We evaluate policies trained on each of the three variability levels on all levels, and also compare to the~\gls{zz} \textit{Heuristic} on all setups. 
Figure~\ref{fig:material_generalization} presents the quantitative results. 
Training~\gls{method} on a limited range of parameter variations enables it to generalize effectively to a broader set of out-of-distribution parameters, while still maintaining high refinement accuracy on the fixed material parameters used in the main experiments.
\rebuttal{This generalization ability comes at no additional computational cost, since the difference between these policies is the variation in material parameters in their training data.}
Although~\gls{method} consistently outperforms the~\gls{zz} \textit{Heuristic}, refinement quality declines slightly as parameter variability increases. 
This decline is likely due to the limited training data of only $100$ systems of equations, and the comparatively limited capacity of the~\gls{mpn} backbone used for the policy.
\rebuttal{For even more complex \glspl{pde} and generalization settings, we thus expect \gls{method} to require a larger number of training problems and an increased latent dimensionality of the policy architecture to ensure robust and accurate refinements.}

\begin{figure}[ht!]
    \centering

    \begin{minipage}[b]{0.32\textwidth}
        \centering
        \includegraphics[height=2.8cm, keepaspectratio]
        {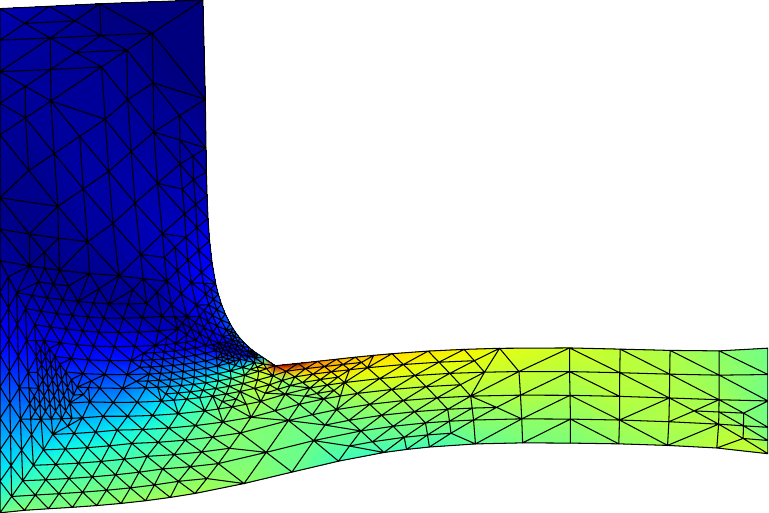}
        \caption*{Concrete}
    \end{minipage}
    \begin{minipage}[b]{0.32\textwidth}
        \centering
        \includegraphics[height=2.8cm, keepaspectratio]
        {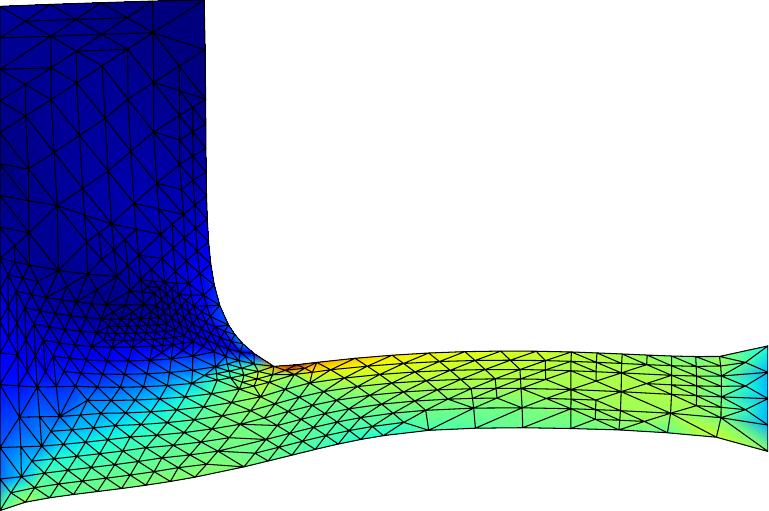}
        \caption*{Aluminium}
    \end{minipage}   
    \begin{minipage}[b]{0.32\textwidth}
        \centering
        \includegraphics[height=2.8cm, keepaspectratio]
        {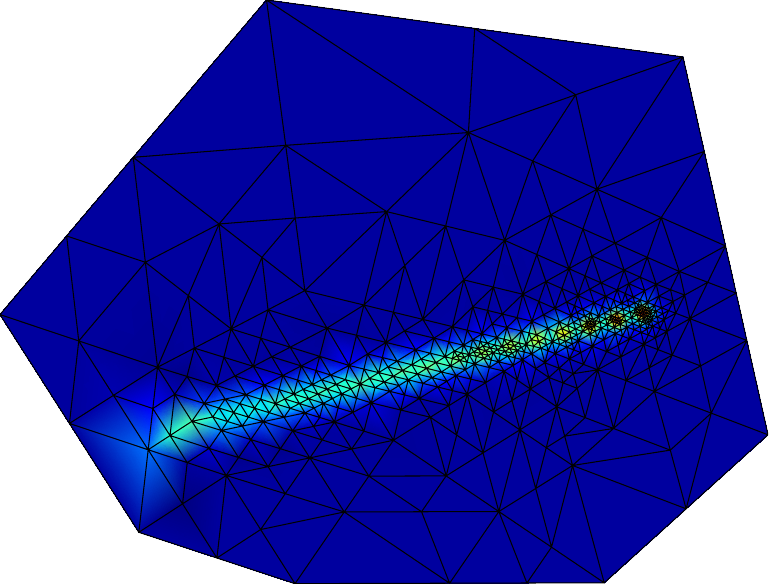}
        \caption*{Diffusivity~$=1.0$e$-5$}
    \end{minipage}
    
    \begin{minipage}[b]{0.32\textwidth}
            \centering
            \includegraphics[height=2.8cm, keepaspectratio]
            {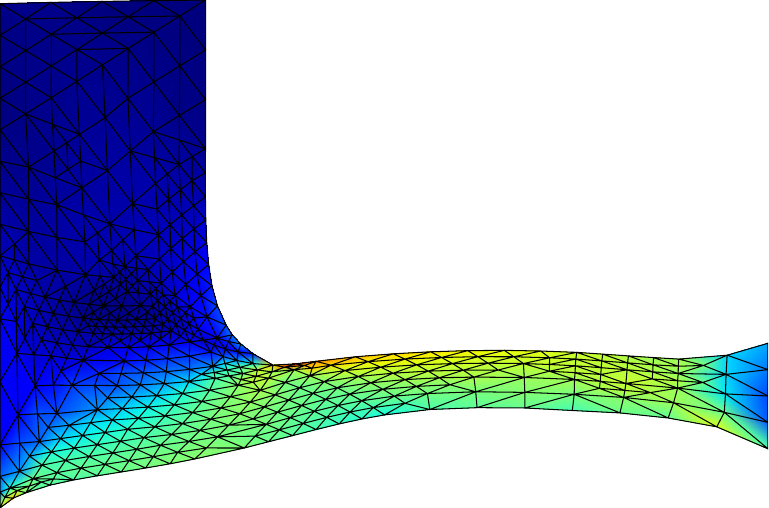}
            \caption*{Gold}
    \end{minipage}    
    \begin{minipage}[b]{0.32\textwidth}
            \centering
            \includegraphics[height=2.8cm, keepaspectratio]
            {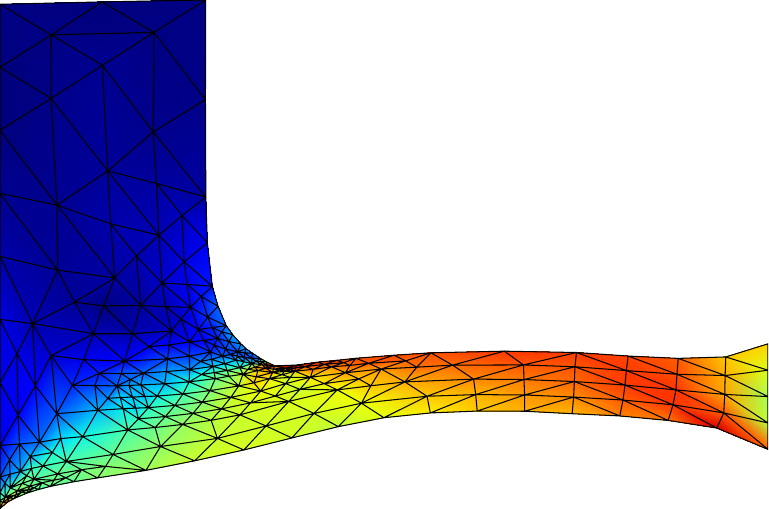}
            \caption*{Polypropylen}
    \end{minipage}
    \begin{minipage}[b]{0.32\textwidth}
            \centering
            \includegraphics[height=2.8cm, keepaspectratio]
            {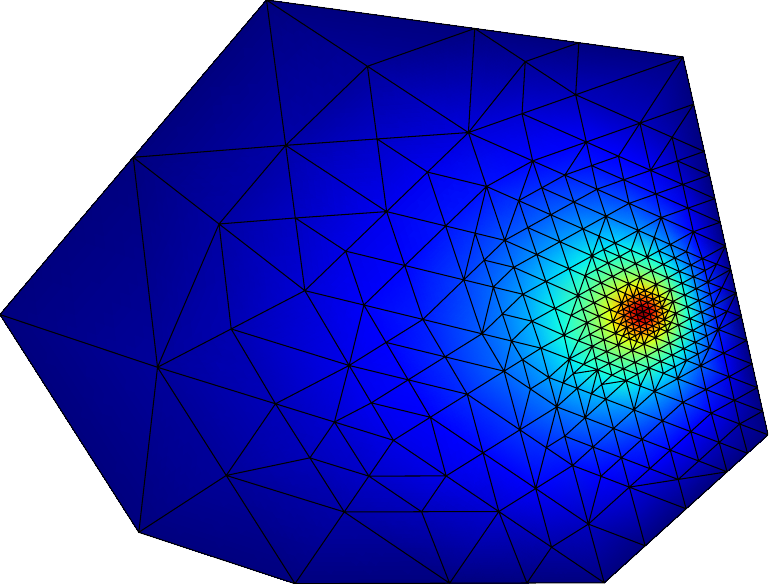}
            \caption*{Diffusivity~$=1.0$e$-1$}
    \end{minipage}
    \vspace{0.01\textwidth}%
    
    \caption{
        Final~\gls{method} meshes and corresponding \gls{fem} solutions for the same Linear Elasticity and Stokes Flow problems with varying material parameters.
        \textbf{Left, Middle}: Training on systems of equations with randomly sampled Poisson's ratio and Young's modulus enables the same~\gls{method} policy to deliver precise refinements for simulations on several different real-world materials, such as
        concrete (Young's modulus 32 GPa, Poisson ratio 0.15), 
        gold (77.2 GPa, 0.43),
        aluminum (68 GPa, 0.32), and
        polypropylene (\mbox{1.2 GPa}, 0.42).
        Note that the displacements in the figures have been intentionally chosen to be extreme to highlight different refinement behaviors.
        \textbf{Right}: Similarly, training with diverse diffusivity parameters allows for refinements for diffusivities spanning several orders of magnitude.
    }
    \label{fig:material_generalization_qualitative}
\end{figure}

Figure~\ref{fig:material_generalization_qualitative} provides final refinements of \gls{method} on exemplary systems of equations for different material parameters. 
For each task, we use a single policy trained on the "large variation" environments to generate all refinements.
We find that training on a range of material parameters allows~\gls{method} to provide tailored refinements on a wide range of material parameters.
\rebuttal{
For example, \gls{method} picks up on the high solution gradient in the lower left corner of the L-shape for Polypropylene, while finding that refinement is not necessary for the more evenly deforming Concrete.
Additionally, since the model is trained to minimize global error, it may occasionally keep secondary refinements that have a negligible impact on the total computational budget.
}

\section{Conclusion}
We introduce \glsfirst{method}, an adaptive mesh refinement method leveraging swarm reinforcement learning that increases the efficiency of numerical simulations with the finite element method.
Our method iteratively evaluates a system of equations on a mesh, generates an observation graph from the mesh and solution, processes this graph with a message passing graph neural network, and subsequently marks mesh elements for subdivision. 
We treat each mesh as a homogeneous agent in a collaborative multi-agent system, training all agents with a shared policy.
Crucially, we provide a per-agent reward that balances the local reduction in simulation error against the cost of adding additional mesh elements.
This cost can be tuned during inference to generate meshes of different resolution using the same policy.
Accommodating different mesh sizes within an episode due to subdivisions, we propose a location-based mapping that assigns each new mesh element to the element in the previous mesh that created it.

Experimental results on stationary, conforming $2$d and $3$d meshes demonstrate that this agent-wise approach creates high-quality refinements for meshes with thousands of elements and without requiring any expensive error estimate during inference.
\gls{method} surpasses both existing reinforcement learning-based methods and traditional refinement techniques, achieving a mesh quality comparable to expensive oracle-based error heuristics.
We conduct additional experiments that highlight the benefits of coupling our per-agent reward function with our temporal agent mapping. 
Further, the method directly generalizes to different process conditions and domains during inference.
When trained on a range of material parameters, it generalizes well to new, out-of-distribution material parameters, and when combined with simulation-specific data augmentation during training, it scales to significantly larger domains and meshes during inference.
For the considered systems of equations, our method is up to $30$ times faster than using a uniform refinement in small domains and can exceed a speed-up of factor $100$ when scaling to larger evaluation setups during inference.

\textbf{Limitations and Future Work}
Our method requires solving the system of equations on the mesh following each refinement, which quickly becomes computationally expensive for finer meshes.
To mitigate this, we aim to integrate auxiliary physics-based losses or policy distillation methods to implicitly learn to predict which regions to refine from just the problem's process conditions. 
Additionally, our reward function is currently based on an expensive reference mesh and solution.
While this reference is only required during training and~\gls{method} generalizes well from as few as $10$ different training systems of equations, the maximum mesh resolution is limited by this reference.
Another promising direction is to apply~\gls{method} to other numerical discretization methods such as the Finite Volume Method.
Lastly, we aim to extend our method from stationary to time-dependent refinement strategies and thus include both mesh refinement and coarsening operations\rebuttal{, as well as scale \gls{method} to more complex non-linear problems, such as anisotropic diffusion}.

\backmatter

\bmhead{Acknowledgements}
NF was supported by the BMBF project Davis (Datengetriebene Vernetzung für die ingenieurtechnische Simulation).
This work is also part of the DFG AI Resarch Unit 5339 regarding the combination of physics-based simulation with AI-based methodologies for the fast maturation of manufacturing processes. The financial support by German Research Foundation (DFG, Deutsche Forschungsgemeinschaft) is gratefully acknowledged. The authors acknowledge support by the state of Baden-Württemberg through bwHPC, as well as the HoreKa supercomputer funded by the Ministry of Science, Research and the Arts Baden-Württemberg and by the German Federal Ministry of Education and Research.

\section*{Declarations}

\paragraph{Funding}
NF was supported by the BMBF project Davis (Datengetriebene Vernetzung für die ingenieurtechnische Simulation). This work is also part of the DFG AI Research Unit 5339 on the combination of physics-based simulation with AI-based methodologies for accelerating the maturation of manufacturing processes. Financial support by the German Research Foundation (DFG, Deutsche Forschungsgemeinschaft) is gratefully acknowledged. Computational resources were provided by the state of Baden-Württemberg through bwHPC, and the HoreKa supercomputer funded by the Ministry of Science, Research and the Arts Baden-Württemberg and the German Federal Ministry of Education and Research.

\paragraph{Conflict of interest}
The authors declare that they have no competing interests.

\paragraph{Ethics approval and consent to participate}
Not applicable.

\paragraph{Consent for publication}
Not applicable.

\paragraph{Data availability}
Our method uses Reinforcement Learning, which generates data on the fly.
Thus, no explicit dataset exists.
The Reinforcement Learning environments are made publicly available together with the rest of the code.

\paragraph{Materials availability}
Not applicable.

\paragraph{Code availability}
We publish code under \url{github.com/NiklasFreymuth/ASMRplusplus}

\paragraph{Author contribution}
Niklas Freymuth: Conceptualization, Methodology, Software, Validation, Formal analysis, Investigation, Data Curation, Writing - Original Draft, Visualization.\\
Philipp Dahlinger: Software, Validation, Writing - Review \& Editing.\\
Tobias Würth: Conceptualization, Data Curation, Writing - Review \& Editing.\\
Simon Reisch: Software, Visualization.\\
Luise Kärger: Writing - Review \& Editing, Supervision.\\
Gerhard Neumann: Conceptualization, Methodology, Writing - Review \& Editing, Supervision, Project administration, Funding acquisition.

\begin{appendices}

\section{Systems of Equations}
\label{app_sec:pde}
The following briefly describes the system of equations used for our experiments, including weak~\glspl{pde}, domains and process conditions.
Each task uses an underlying system of equations and varies either the process conditions, the used domain, or both.
For some tasks, we add additional information about the task as node features to the observation graph of the~\gls{mpn} policy.

\subsection{Laplace's Equation.}
Consider a domain $\Omega$ bounded internally by $\partial\Omega_{\text{in}}$ and externally by $\partial\Omega_{\text{out}}$. 
We seek a function $u(\boldsymbol{x})$ satisfying Laplace's Equation
\begin{equation*}
    \int_\Omega \nabla u \cdot \nabla v ~\text{d}\boldsymbol{x} = 0 \quad \forall v \in V \text{,}
\end{equation*}
where $v(\boldsymbol{x})$ denotes the test function and the solution $u(\boldsymbol{x})$ has to satisfy the Dirichlet boundary conditions
\begin{equation*}
u(\boldsymbol{x}) = 0 \text{,}~ \boldsymbol{x} \in \partial\Omega_{\text{in}} \quad \text{and} \quad
u(\boldsymbol{x}) = 1 , \boldsymbol{x} \in \partial\Omega_{\text{out}}\text{.}
\end{equation*}
The domain is a unit square $(0,1)^2$ with a squared hole defined by the inner boundary $\partial\Omega_{\text{in}}$.
For each system of equations, the hole's size is uniformly sampled from $U(0.05, 0.25)$ in both directions, while its center is placed randomly in $U(0.2, 0.8)^2$.
The minimum distance to $\partial\Omega_{\text{in}}$ of each face midpoint is included as a node feature.

\subsection{Poisson's Equation.}
\label{app_sec:poisson_equation}

We address Poisson's Equation
\begin{align*}
\int_\Omega \nabla u \cdot \nabla v\text{d}\boldsymbol{x} = \int_\Omega f v\text{d}\boldsymbol{x} \quad \forall v \in V \text{,}
\end{align*}
where $f(\boldsymbol{x})$ is the load function, and $v(\boldsymbol{x})$ is the test function. 
The solution $u(\boldsymbol{x})$ must satisfy $u(\boldsymbol{x}) = 0$ on $\partial \Omega$.
We use L-shaped domain $\Omega$, defined as $(0,1)^2\backslash (p_0\times(1,1))$, where the lower left corner $p_0$ is sampled from $U(0.2, 0.95)$ in $x$ and $y$ direction. 

We employ a Gaussian Mixture Model with three components for each problem's load function. 
The mean of each component is drawn from $U(0.1, 0.9)^2$, with rejection sampling ensuring all means reside within $\Omega$. 
We then independently draw diagonal covariances from a log-uniform distribution $\exp(U(\log(0.0001, 0.001)))$ and randomly rotate them to yield a full covariance matrix. 
Component weights are generated from $\exp(N(0,1))+1$ and subsequently normalized.
The load function evaluation at each face's midpoint is used as a node feature.

\subsection{Stokes Flow.}
\label{app_sec:stokes_flow}

Let $\boldsymbol{u}(\boldsymbol{x})$ and $p(\boldsymbol{x})$ represent the velocity and pressure fields, respectively, in a channel flow. 
We analyze the Stokes flow, aiming to solve for $u$ and $p$ without a forcing term, i.e.,
$$
\nu \int_\Omega\nabla \boldsymbol{v} \cdot \nabla \boldsymbol{u} ~\text{d}\boldsymbol{x} -\int_\Omega (\nabla \cdot \boldsymbol{v}) p ~\text{d} 
\boldsymbol{x} = 0 \quad \forall  \boldsymbol{v} \in \boldsymbol{V} \text{,}
$$
$$
\int_\Omega (\nabla \cdot \boldsymbol{u}) q ~\text{d}\boldsymbol{x} = 0 \quad  \forall q \in V \text{,}
$$
with test functions $\boldsymbol{v}(\boldsymbol{x})$ and $q(\boldsymbol{x})$~\citep{quarteroni2009numerical}.
The velocity field at the inlet is defined as
$$
\boldsymbol{u}(x = 0,y) = u_\text{P}  y (1 - y) + \sin{ \left( \varphi + 2 \pi y  \right)} \text{.}
$$
At the outlet, we impose $\nabla \boldsymbol{u}(x=1, y) = \boldsymbol{0}$, and assume a no-slip condition $\boldsymbol{u} = \boldsymbol{0}$ at all other boundaries.
For numerical stability, we utilize $P_2/P_1$ Taylor-Hood elements, with quadratic velocity and linear pressure shape functions~\citep{john2016finite}. 

The inlet profile parameter $u_\text{P}$ is sampled from a log-uniform distribution $\exp(U(\log(0.5, 2)))$. 
The domain is structured as a unit square with three rhomboid holes of length $0.4$ and height $0.2$.
The holes are centered at $y\in\{0.2, 0.5, 0.8\}$ and we randomly sample their x-coordinate from $U(0.3, 0.7)$.
We optimize the meshes w.r.t. the velocity vector and present a scalar error as the norm of the vectorized velocity error.

\subsection{Linear Elasticity.}
\label{app_sec:linear_elasticity}

We investigate the steady-state deformation of a solid under stress caused by displacements at the boundary $\partial\Omega$ of the domain $\Omega$.
Let $\boldsymbol{u}(\boldsymbol{x})$ be the displacement field, $\boldsymbol{v}(\boldsymbol{x})$ the test function, and
$$\boldsymbol{\varepsilon}\left(\boldsymbol{u}\right) = \frac{1}{2}(\nabla \boldsymbol{u} + (\nabla \boldsymbol{u})^\top)$$
the strain tensor.
The linear-elastic and isotropic stress tensor is given as
$$
\boldsymbol{\sigma}\left(\boldsymbol{\varepsilon}\right) = 2\mu \boldsymbol{\varepsilon} + \lambda \text{tr}(\boldsymbol{\varepsilon}) \boldsymbol{I}\text{,}
$$
using Lamé parameters
$$
\lambda = \frac{E\nu}{(1 + \nu)(1 - 2\nu)}
\quad \text{and} \quad
\mu = \frac{E}{2(1 + \nu)}
$$
with a problem specific Young's modulus $E = 1$ and Poisson ratio $\nu = 0.3$.
Without body forces, the problem is given as~\citep{zienkiewicz2005finite}
$$
\int_{\Omega} \boldsymbol{\sigma} \left(\boldsymbol{\varepsilon}\left(\boldsymbol{u}\right)\right) : \boldsymbol{\varepsilon}\left(\boldsymbol{v}\right) ~\text{d}\boldsymbol{x} = 0 \quad \forall  \boldsymbol{v} \in \boldsymbol{V} \text{.}
$$

We use the same class of L-shaped domains as in the Poisson problem in Section \ref{app_sec:poisson_equation}.
We fix the displacement of the left boundary to $0$, i.e., $\boldsymbol{u} (x = 0, y) = 0$, and randomly sample a displacement direction from $U[0,\pi]$ and magnitude from $U(0.2, 0.8)$ to displace the right boundary as
$\boldsymbol{u} (x = 1, y) = \boldsymbol{u}_{P}$.
The stress $\boldsymbol{\sigma} \cdot \boldsymbol{n} = \boldsymbol{0}$ is zero normal to the boundary at both the top and bottom of the part. 
The task-dependent displacement vector $\boldsymbol{u}_{P}$ is added as a globally shared node feature to all elements. 
Our objective includes the norm of the displacement field $\boldsymbol{u}$ and the resulting Von-Mises stress, using an equal weighting between the two.

\subsection{Non-Stationary Heat Diffusion.}
\label{app_sec:heat_diffusion}
Using temperature $u(\boldsymbol{x})$, a test function $v(\boldsymbol{x})$, and a thermal diffusivity constant $a=0.001$, we address a non-stationary thermal diffusion problem defined by
$$
\int_\Omega \frac{\partial u }{\partial t} ~\text{d}\boldsymbol{x} +\int_\Omega a \nabla u \cdot \nabla v ~\text{d}\boldsymbol{x} = \int_\Omega f v ~\text{d}\boldsymbol{x} \quad \forall v \in V \text{.}
$$
Here, $f(x,y)$ is a position-dependent heat distribution 
$$
f(x,y) = 1\,000  ~ \text{exp} \left(-100 ~ ((x-x_p(\tau)) + (y-y_p(\tau)))\right)\text{.}
$$
The position of the heat source's path $\boldsymbol{p}_\tau (\tau) = \left( x_p(\tau), y_p(\tau) \right)$ is linearly interpolated over time as 
$$
\boldsymbol{p}_\tau=\boldsymbol{p}_0+\frac{\tau}{\tau_{\text{max}}}(\boldsymbol{p}_{\tau_{\text{max}}}-\boldsymbol{p}_0)\text{.}
$$
The start and goal positions $\boldsymbol{p}_0$ and $\boldsymbol{p}_{\tau_{\text{max}}}$ are randomly drawn from the entire domain.
The temperature $u \in \partial\Omega$ is set to zero on all boundaries. 
We use a total of $\tau_{\text{max}}=20$ uniform time steps in $\{0.5, \dots, 10\}$ and an implicit Euler method for the time-integration.
Domain geometry is derived from $10$ equidistant points on a circle centered at $(0.5, 0.5)$ with a radius of $0.4$. 
Each point is randomly distorted by a value $U(-0.2, 0.2)^2$ before we proceed to calculate the points convex hull as our domain.
We finally normalize the convex hull to lie in $(0,1)^2$.
The result is a family of convex polygons with up to $10$ vertices.
We provide and measure the error and solution of the final simulation step, and provide the distance to the start and end position of the heat source as additional node features for each element.

\subsection{Neumann Boundaries.}
\label{app_sec:neumann_boundaries}

We adapt Poisson's Equation in Section~\ref{app_sec:poisson_equation} to include Neumann boundary conditions.
Using star-shaped domains, we randomly select half the line segments $\partial\Omega_l$ that make up the boundary of the star as Neumann boundaries, and use zero Dirichlet boundary conditions for the remaining segments.
Each selected line segment is assigned a random sinusoidal Neumann boundary condition
$$
\nabla u \cdot \boldsymbol{n} = 10\cdot A \cdot p_{\boldsymbol{x}} \cdot (1-p_{\boldsymbol{x}}) \cdot \sin (\pi \nu p_{\boldsymbol{x}}) \text{,} \quad \boldsymbol{x}\in\partial\Omega_l \text{,}
$$
where $\boldsymbol{n}(\boldsymbol{x})$ denotes the normal vector of the boundary segment, $p_{\boldsymbol{x}}\in[0,1]$ is the normalized position along the line segment, $A\in U(1, 3)$ the amplitude and $\nu\in\{3, 5, 7\}$ the frequency.

To create the star-shaped domains, we define an outer radius of $0.5$ and an inner radius of $0.2$ from a point centered at $(0.5, 0.5)$.
We then sample a number of star points in $U(3, 5)$ and equidistantly interleave them between the two boundaries.
We randomly distort each point by $U(-0.05, 0.05)^2$ before applying a random rotation and normalizing the full domain to be within $(0,1)^2$.
Initial meshes are created with a target volume of $0.02$ instead of the $0.05$ used for the other tasks to compensate for the lower total volume of the star-shaped domains, and the diagonal covariances for the Gaussian mixture model load function are drawn from $U(0.00005, 0.0005)$.
Each element gets the distance to the closest Neumann and Dirichlet boundaries as additional node features.

\subsection[3-Dimensional Poisson's Equation.]{$3$-Dimensional Poisson's Equation.}

Finally, we extend Poisson's Equation in Section~\ref{app_sec:poisson_equation} to a $3$-dimensional domain.
We use a rectangular plate of length $l_x=1$, width $l_y=0.5$ and height $l_z=0.1$ as our domain, and create an initial mesh from by dividing it into $0.1^3$ cubes and triangulating each cube.
We assign Dirichlet boundary conditions 
$$
u(x = 0, y, z) = u(x = 1, y, z) = u(x, y=0, z) = u(x = 1, y=0.6, z) = 0 
$$
on the sides of the plate and define natural boundary conditions on the lower $z=0$ and upper side $z=0.1$  of the plate.
We use tetrahedral elements for this task, allowing for a fully volumetric mesh, and solve with an iterative instead of a direct solver to accommodate the larger number of elements.
The remesher uses longest edge bisection~\citep{rivara1984algorithms, suarez2005computational} to halve marked elements, whereas the $2$-dimensional domains employ the red-green-blue refinement method~\citep{carstensen2004adaptive}.

\section{Further Experiments}

\subsection{Experiment Setup.}
\label{app_sec:experiment_details}
We repeat each experiment for $10$ different random seeds, randomizing the parameters of the neural network as well as the~\glspl{pde}, domains and process conditions.
We normalize all domains to unit size unless mentioned otherwise.
We create initial meshes with meshpy\footnote{\url{https://github.com/inducer/meshpy}} with a target element size of $0.05$ for all tasks except for the Neumann boundary task, which uses a target element size of $0.02$.
For all experiments, we terminate an episode with a large negative reward of $-1\,000$ when a threshold of $20\,000$ elements, respectively $50\,000$ elements for the $3$d Poisson task, is exceeded.
Unless mentioned otherwise, we train all policies on $100$ randomized systems of equations, but evaluate on the same set of $100$ fixed systems of equations across different methods and seeds for better comparability.
All experiments are run for up to $3$ days on $8$ cores of an Intel Xeon Platinum $8358$ CPU, but usually terminate within a single day.

We train a total of $5$~\gls{rlamr} methods on $7$ separate tasks for $10$ repetitions each.
Since $3$ of these methods train on fixed number of target mesh elements, this leads to $(2+(3\cdot10))\cdot7\cdot10=2240$ core experiments.
A similar combined number of experiments is needed for the heuristics, ablations and preliminary experiments.

\subsection{Volume Reward.}
\label{app_ssec:volume_reward}
Equation~\ref{eq:asmr_local_reward} rewards the reduction in maximum element error for a given refinement.
While this has been proposed as an ablation by~\gls{asmr}~\citep{freymuth2024swarm}, it instead uses a different reward for all experiments.
This reward considers a numerically integrated error per element, i.e., replaces Equation~\ref{eq:err_per_element} with
\begin{equation*}
  \hat{\text{err}}(\Omega_i^t)\approx \sum_{\Omega_m^*\subseteq \Omega_i^t} \text{Volume}(\Omega^*_m)\left|u_{\Omega^*}(p_{\Omega_m^*})-u_{\Omega^t}(p_{\Omega_m^*})\right|
\end{equation*}
and changes Equation~\ref{eq:asmr_local_reward} to a volume-scaled reduction in this integrated error, i.e., 
\begin{equation}
\label{eq:volume_reward}
    \mathbf{r}(\Omega^t_i) := 
    \frac{1}{\text{Volume}(\Omega^t_i)} \left(\text{err}(\Omega^t_i)
    -\sum_j\mathbf{M}^t_{ij}\text{err}(\Omega^{t+1}_j)\right)
    -\alpha\left(\sum_j\mathbf{M}^t_{ij}-1\right)\text{.}
\end{equation}
Both reward formulations evaluates if for any refinement, the reduction in error is greater than the cost of adding additional elements.
The volume scaling component in Equation~\ref{eq:volume_reward} encourages the policy to give priority to refining smaller mesh elements, roughly canceling out the volume of the integration points in Equation~\ref{eq:err_per_element}, aiming to reduce the average error per unit volume for each element. 
Optimizing this reward lowers the errors for each mesh element, adjusted by the inverse of its volume, which promotes a uniform distribution of error in relation to the size of each element across the mesh.
Opposed to this, the maximum reward of Equation~\ref{eq:asmr_local_reward} minimizes the maximum error across any integration point in the mesh, posing a more direct and clear objective.
We compare to this reward in Section~\ref{app_ssec:ablations} and show the difference in performance in Figure~\ref{app_fig:abl_reward}, finding that both variants perform similarly.

\section{Extended Results}

The following sections evaluate various algorithmic design choices for the different~\gls{rlamr} methods and the~\gls{zz} heuristic.
In particular, we evaluate different~\gls{rl} backbones and \gls{gnn} architectures to ensure a fair comparison between the different approaches.
We also conduct experiments on various small changes in the setup and training of~\gls{method} to find and explain what makes it effective.
All experiments are conducted on the Stokes flow task unless mentioned otherwise, as the task requires challenging refinements at various parts of the domain.

\subsection{Initial Meshes for the ZZ Error Heuristic.}

\begin{figure}[ht!]
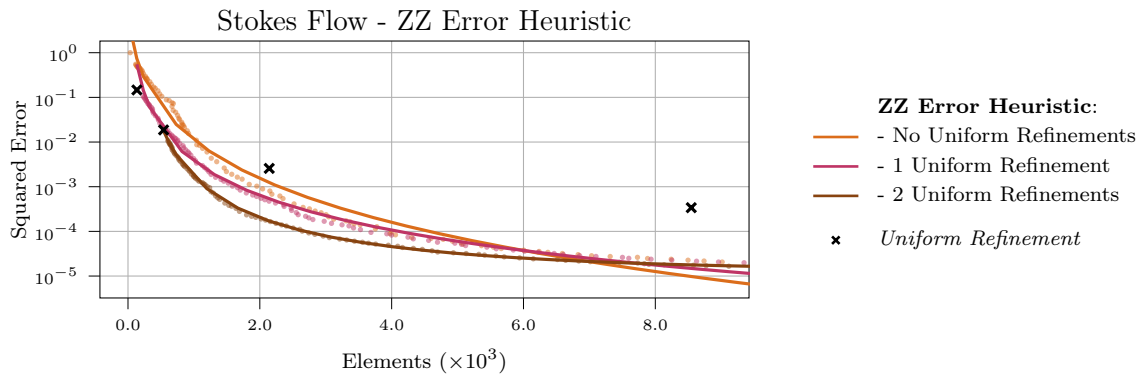

    \centering 
    \begin{minipage}{0.62\textwidth}
        \tikzsetnextfilename{app_zz_error_ablation}
        \input{appendix/figures/quantitative/ablations/tikz/zz_error_uniform_refinements_plot}
    \end{minipage}%
    \begin{minipage}{0.01\textwidth}
    ~
    \end{minipage}%
    \begin{minipage}{0.33\textwidth}
        \tikzsetnextfilename{app_zz_error_ablation_legend}
        \begin{tikzpicture}
\tikzstyle{every node}'=[font=\scriptsize]
\input{tikz_colors}
\begin{axis}[%
hide axis,
xmin=10,
xmax=50,
ymin=0,
ymax=0.1,
legend style={
    draw=white!15!black,
    legend cell align=left,
    legend columns=1,
    legend style={
        draw=none,
        column sep=1ex,
        line width=1pt
    }
},
]
\addlegendimage{empty legend}
\addlegendentry{\textbf{ZZ Error Heuristic}:}
\addlegendimage{chocolate21910927}
\addlegendentry{- No Uniform Refs.}
\addlegendimage{indianred19152101}
\addlegendentry{- $1$ Uniform Ref.}
\addlegendimage{saddlebrown1356716}
\addlegendentry{- $2$ Uniform Refs.}
\addlegendimage{empty legend}
\addlegendentry{~}
\addlegendimage{only marks, mark=x, black}
\addlegendentry{\textit{Uniform Refinement}}

\end{axis}
\end{tikzpicture}    
    \end{minipage}
    \caption{
        Pareto plot of normalized squared errors and number of final mesh elements on the Stokes flow task for the \glsfirst{zz} \textit{Heuristic} for $0$, $1$ and $2$ initial uniform mesh refinements.
        A finer initial mesh leads to improved performance for the \gls{zz} \textit{Heuristic} but prevents the creation of very coarse meshes, indicating that the initial element size should be tuned for the~\gls{zz} \textit{Heuristic}.
    }
    \label{app_fig:zz_error_ablation}
\end{figure}

\label{app_ssec:zz_error_ablation}
Figure~\ref{app_fig:zz_error_ablation} demonstrates the performance of the \gls{zz} \textit{Heuristic} applied to the initial mesh, in comparison to its performance when the process begins with either one or two uniform mesh refinements. 
The results indicate that initial uniform refinements significantly improve the heuristic's effectiveness, likely because the uniform refinements make it easier to identify gradients in key areas, which may be overlooked if the mesh is too coarse.
Consequently we opt for the twice-refined mesh approach for all experiments. 
In comparison, the~\gls{rlamr} methods do not need to tune their initial mesh, as the methods are trained to produce optimal refinement sequences.

\subsection{Proximal Policy Optimization and Deep Q-Networks.}
\begin{figure}
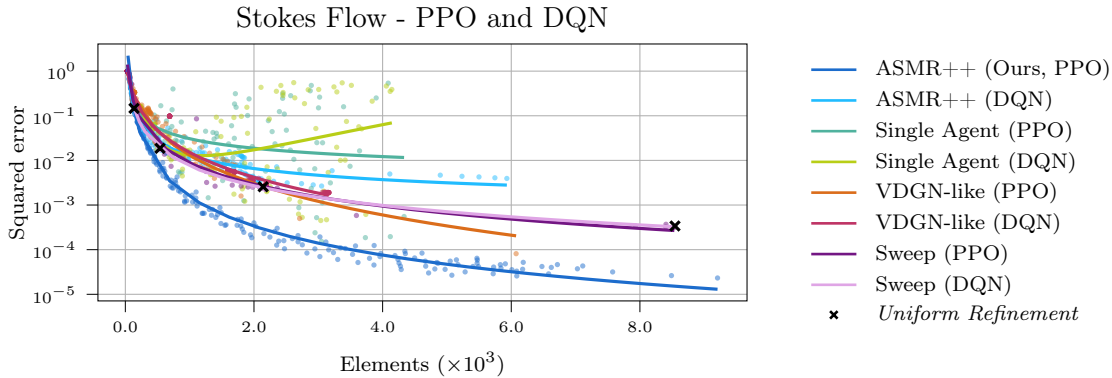

    \centering 
    \begin{minipage}{0.62\textwidth}
        \tikzsetnextfilename{app_abl_ppo_dqn}
        \input{appendix/figures/quantitative/ablations/tikz/stokes_ppo_dqn__squared_error}   
    \end{minipage}%
    \begin{minipage}{0.01\textwidth}
    ~
    \end{minipage}%
    \begin{minipage}{0.33\textwidth}
        \tikzsetnextfilename{app_abl_ppo_dqn_legend}
        \begin{tikzpicture}
\tikzstyle{every node}'=[font=\scriptsize]
\input{tikz_colors}
\begin{axis}[%
                    hide axis,
                    xmin=10,
                    xmax=50,
                    ymin=0,
                    ymax=0.1,
                    legend style={
                        draw=white!15!black,
                        legend cell align=left,
                        legend columns=1,
                        legend style={
                            draw=none,
                            column sep=1ex,
                            line width=1pt
                        }
                    },
                    ]
                    
\addlegendimage{royalblue28108204}
\addlegendentry{ASMR++ (Ours, PPO)}
\addlegendimage{deepskyblue33188255}
\addlegendentry{ASMR++ (DQN)}
\addlegendimage{cadetblue80178158}
\addlegendentry{Single Agent (PPO)}
\addlegendimage{yellowgreen18420623}
\addlegendentry{Single Agent (DQN)}
\addlegendimage{chocolate21910927}
\addlegendentry{VDGN-like (PPO)}
\addlegendimage{indianred19152101}
\addlegendentry{VDGN-like (DQN)}
\addlegendimage{purple11522131}
\addlegendentry{Sweep (PPO)}
\addlegendimage{plum223165229}
\addlegendentry{Sweep (DQN)}
\addlegendimage{only marks, mark=x, black}
\addlegendentry{\textit{Uniform Refinement}}

\end{axis}
\end{tikzpicture}    
    \end{minipage}
    \caption{
        Pareto plot of normalized squared errors and number of final mesh elements on the Stokes flow task for the different~\gls{rl}-\gls{amr} methods using \gls{ppo} and \gls{dqn} backends.
        The on-policy~\gls{ppo} outperforms the off-policy~\gls{dqn} for all methods. 
        Only~\gls{method} consistently performs better than a uniform mesh.
    }
    \label{app_fig:abl_ppo_dqn}
\end{figure}

\label{app_ssec:ppo_vs_dqn}
Figure~\ref{app_fig:abl_ppo_dqn} shows results on the Stokes flow task for \glsfirst{ppo}~\citep{schulman2017proximal} and \glsfirst{dqn}~\citep{mnih2013playing, mnih2015human} as the \gls{rl} backbone for all~\gls{rlamr} algorithms.
For the \gls{ppo} version of the \gls{vdgn}-like baseline, we apply the value decomposition~\citep{sunehag2017value} to the value function instead of the Q-function, i.e., we define the value function of the full mesh as the sum of value functions of the individual mesh elements.
Further, we use a mean instead of a sum for the agent mapping of the TD error in Equation~\ref{eq:td_error} for training the $Q$-value of the \gls{dqn} experiments of \gls{method}, as this experimentally increases training stability.
\gls{ppo} generally outperforms \gls{dqn}, suggesting that on-policy optimization is favorable for the changing observation and action spaces of \gls{amr}.
We thus choose \gls{ppo} for all further experiments as it leads to better performance for all methods.

\subsection{Design Choices and Ablations.}
\label{app_ssec:ablations}

\subsubsection{Graph Attention Networks.}

\label{app_ssec:mpn_vs_gat}
\begin{figure}
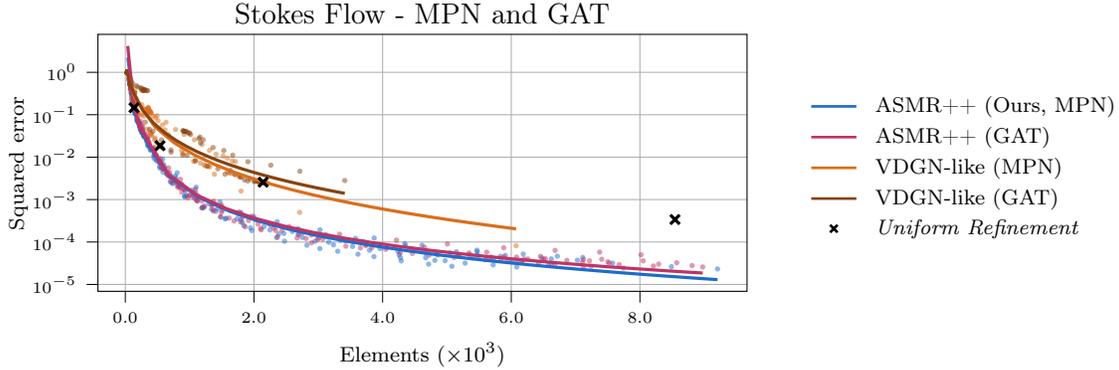

    \centering 
    \begin{minipage}{0.62\textwidth}
        \tikzsetnextfilename{app_abl_gat}
        \input{appendix/figures/quantitative/ablations/tikz/stokes_gat__squared_error}   
    \end{minipage}%
    \begin{minipage}{0.01\textwidth}
    ~
    \end{minipage}%
    \begin{minipage}{0.33\textwidth}
        \tikzsetnextfilename{app_abl_gat_legend}
        \begin{tikzpicture}
\tikzstyle{every node}'=[font=\scriptsize]
\input{tikz_colors}
\begin{axis}[%
                    hide axis,
                    xmin=10,
                    xmax=50,
                    ymin=0,
                    ymax=0.1,
                    legend style={
                        draw=white!15!black,
                        legend cell align=left,
                        legend columns=1,
                        legend style={
                            draw=none,
                            column sep=1ex,
                            line width=1pt
                        }
                    },
                    ]
                    \addlegendimage{royalblue28108204}
\addlegendentry{ASMR++ (Ours, MPN)}
\addlegendimage{indianred19152101}
\addlegendentry{ASMR++ (GAT)}
\addlegendimage{chocolate21910927}
\addlegendentry{VDGN-like (MPN)}
\addlegendimage{saddlebrown1356716}
\addlegendentry{VDGN-like (GAT)}
\addlegendimage{only marks, mark=x, black}
\addlegendentry{\textit{Uniform Refinement}}
\end{axis}
\end{tikzpicture}    
    \end{minipage}
    \caption{
        Pareto plot of normalized squared errors and number of final mesh elements on the Stokes flow task for \gls{method} and the \gls{vdgn}-like baseline for~\gls{mpn} and \gls{gat} network architectures.
        For both methods, the~\gls{mpn} shows slightly better performance, though the impact of the architecture is comparatively minor.
    }
    \label{app_fig:abl_gat}
\end{figure}

Figure~\ref{app_fig:abl_gat} compares the \glspl{mpn} of Section \ref{ssec:mpn_description} with the \glsfirst{gat} architecture proposed by \gls{vdgn}~\citep{yang2023multi}.
The \gls{mpn} slightly improves over the \gls{gat} for both methods, leading us to choose this architecture for all further experiments.

\subsubsection{Element Penalty.}

\begin{figure}[ht!]
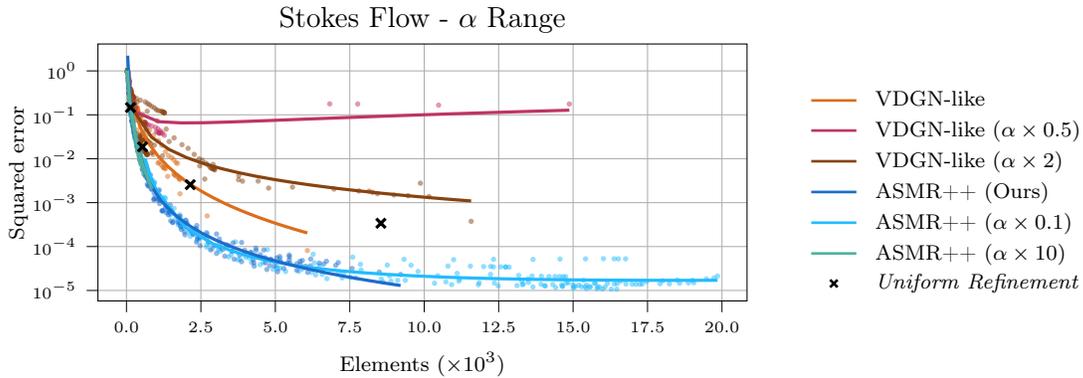

    \centering
    \begin{minipage}{0.62\textwidth}
        \tikzsetnextfilename{app_abl_element_penalty_range}
        \input{appendix/figures/quantitative/ablations/tikz/stokes_element_penalty_range__squared_error} 
    \end{minipage}%
    \begin{minipage}{0.01\textwidth}
    ~
    \end{minipage}%
    \begin{minipage}{0.33\textwidth}
        \tikzsetnextfilename{app_abl_element_penalty_range_legend}
        \begin{tikzpicture}
\tikzstyle{every node}'=[font=\scriptsize]
\input{tikz_colors}
\begin{axis}[%
                    hide axis,
                    xmin=10,
                    xmax=50,
                    ymin=0,
                    ymax=0.1,
                    legend style={
                        draw=white!15!black,
                        legend cell align=left,
                        legend columns=1,
                        legend style={
                            draw=none,
                            column sep=1ex,
                            line width=1pt
                        }
                    },
                    ]
\addlegendimage{chocolate21910927}
\addlegendentry{VDGN-like}
\addlegendimage{indianred19152101}
\addlegendentry{VDGN-like ($\alpha\times 0.5$)}
\addlegendimage{saddlebrown1356716}
\addlegendentry{VDGN-like ($\alpha\times 2$)}

\addlegendimage{royalblue28108204}
\addlegendentry{ASMR++ (Ours)}
\addlegendimage{deepskyblue33188255}
\addlegendentry{ASMR++ ($\alpha\times 0.1$)}
\addlegendimage{cadetblue80178158}
\addlegendentry{ASMR++ ($\alpha\times 10$)}

\addlegendimage{only marks, mark=x, black}
\addlegendentry{\textit{Uniform Refinement}}

\end{axis}
\end{tikzpicture}   
    \end{minipage}

    \caption{
        Pareto plot of normalized squared errors and number of final mesh elements on the Stokes flow task for different ranges of the element penalty $\alpha$ for~\gls{method} and the~\gls{vdgn}-like baseline.
        Increasing or decreasing the element penalty of~\gls{method} by a factor of $10$ leads to finer meshes for larger penalties and vice versa.
        Regardless of the scale of the element penalty,~\gls{method} produces high-quality meshes. 
        The~\gls{vdgn}-like baseline is comparatively unstable, yielding worse results when adapting the element penalty.
    }
    \label{app_fig:abl_element_penalty_range}
    
\end{figure}

Figure~\ref{app_fig:abl_element_penalty_range} shows the effect of different ranges for the element penalty $\alpha$ for~\gls{method} and the~\gls{vdgn}-like baseline.
Here, we scale the minimum and maximum sampled penalty by a constant factor during training and inference.
While the~\gls{vdgn}-like baseline works well for the chosen penalty, it is generally unstable and fails to produce consistent refinements for finer meshes, regardless of whether the range of $\alpha$ is scaled by a factor of $2$.
In comparison, \gls{method} produces high-quality meshes for different element penalty ranges.
When decreasing all penalties by a factor of $10$,~\gls{method} produces finer meshes with significantly more elements compared to the regular penalty. 
Similarly, increasing the penalty makes refinements less attractive, reducing the number of produced elements.

\subsubsection[Number of Training PDEs]{Number of Training~\glspl{pde}.}

\begin{figure}
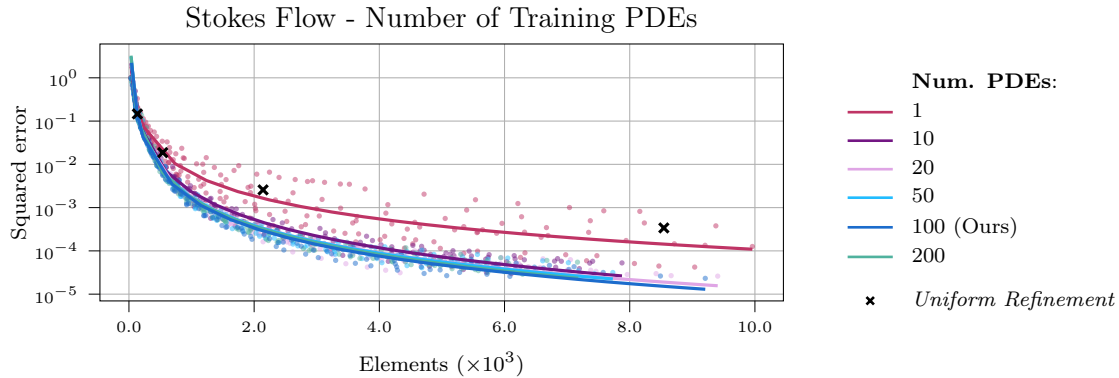

    \centering 
    \begin{minipage}{0.62\textwidth}
        \tikzsetnextfilename{app_abl_npde}
        \input{appendix/figures/quantitative/ablations/tikz/stokes_npde__squared_error}
    \end{minipage}%
    \begin{minipage}{0.01\textwidth}
    ~
    \end{minipage}%
    \begin{minipage}{0.33\textwidth}
        \tikzsetnextfilename{app_abl_npde_legend}
        \begin{tikzpicture}
\tikzstyle{every node}'=[font=\scriptsize]
\input{tikz_colors}

\begin{axis}[%
                    hide axis,
                    xmin=10,
                    xmax=50,
                    ymin=0,
                    ymax=0.1,
                    legend style={
                        draw=white!15!black,
                        legend cell align=left,
                        legend columns=1,
                        legend style={
                            draw=none,
                            column sep=1ex,
                            line width=1pt
                        }
                    },
                    ]
\addlegendimage{empty legend}
\addlegendentry{\textbf{Num. PDEs}:}
\addlegendimage{indianred19152101}
\addlegendentry{$1$}
\addlegendimage{purple11522131}
\addlegendentry{$10$}
\addlegendimage{plum223165229}
\addlegendentry{$20$}
\addlegendimage{deepskyblue33188255}
\addlegendentry{$50$}
\addlegendimage{royalblue28108204}
\addlegendentry{$100$ (Ours)}
\addlegendimage{cadetblue80178158}
\addlegendentry{$200$}
\addlegendimage{empty legend}
\addlegendentry{~}
\addlegendimage{only marks, mark=x, black}
\addlegendentry{\textit{Uniform Refinement}}
\end{axis}
\end{tikzpicture}    
    \end{minipage}
    \caption{
        Pareto plot of normalized squared errors and number of final mesh elements on the Stokes flow task for different numbers of training systems of equations.
        \gls{method} performs better than uniform on novel evaluation settings when trained on a single system of equations, and improves for up to $100$ systems of equations in the training set.
    }
    \label{app_fig:abl_npde}
\end{figure}

We generally train~\gls{method} on $100$ systems of equations, each of which consists of a randomly sampled domain and random process conditions.
Figure~\ref{app_fig:abl_npde} explores how fewer or more systems of equations during training affect performance, finding that performance improves for up to $100$ points of data.
\rebuttal{
Beyond $100$ training systems of equations, performance gains diminish. 
During training, the reward function requires solving a fine-grained reference mesh $\Omega^*$ for each \glspl{pde}, which is computationally expensive.
We thus use $100$ training systems of equations for our main experiments.
}
Albeit noisy and inconsistent, \gls{method} performs better than uniform refinements on the unseen evaluation set when trained on a single system of equations, likely due to the mostly local optimization objective and network architecture.

\subsubsection{Network Architecture}

\begin{figure}
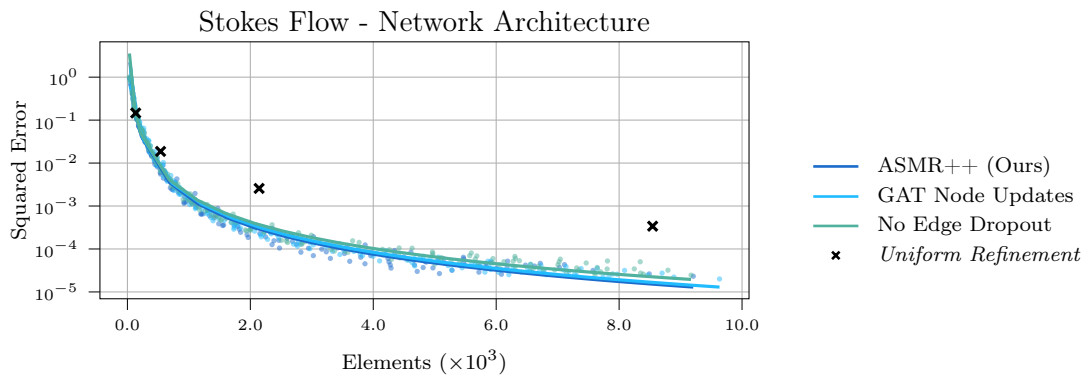

    \centering
    \begin{minipage}{0.62\textwidth}
        \tikzsetnextfilename{app_abl_architecture}
        \input{appendix/figures/quantitative/ablations/tikz/stokes_architecture__squared_error}
    \end{minipage}%
    \begin{minipage}{0.01\textwidth}
    ~
    \end{minipage}%
    \begin{minipage}{0.33\textwidth}
        \tikzsetnextfilename{app_abl_architecture_legend}
        \begin{tikzpicture}
\tikzstyle{every node}'=[font=\scriptsize]
\input{tikz_colors}
\begin{axis}[%
                    hide axis,
                    xmin=10,
                    xmax=50,
                    ymin=0,
                    ymax=0.1,
                    legend style={
                        draw=white!15!black,
                        legend cell align=left,
                        legend columns=1,
                        legend style={
                            draw=none,
                            column sep=1ex,
                            line width=1pt
                        }
                    },
                    ]

\addlegendimage{royalblue28108204}
\addlegendentry{ASMR++ (Ours)}
\addlegendimage{deepskyblue33188255}
\addlegendentry{GAT Node Updates}
\addlegendimage{cadetblue80178158}
\addlegendentry{No Edge Dropout}
\addlegendimage{only marks, mark=x, black}
\addlegendentry{\textit{Uniform Refinement}}

\end{axis}
\end{tikzpicture}
    \end{minipage}

    \caption{
        Pareto plot of normalized squared errors and number of final mesh elements on the Stokes flow task for different network architectures.
        Performing message passing updates with Graph Attention Networks with edge features instead of~\glspl{mpn} does not significantly impact performance.
        Omitting Edge Dropout leads to slightly worse refinements.
    }
    \label{app_fig:abl_architecture}
    
\end{figure}

We ablate our choice of network architecture on the right side of Figure~\ref{app_fig:abl_architecture}.
Using Graph Attention Networks~\citep{velickovic2018graph} with edge features instead of our~\glsfirst{mpn} does not significantly impact performance.
\gls{method} uses Edge Dropout~\citep{rong2019dropedge} of $0.1$, i.e., it randomly removes every tenth edge of the observation graph, regularizing the training data and incentivizing the model to learn general patterns instead of spurious correlations.
Omitting this dropout leads to slightly worse refinements.

\subsection{Mean and Maximum Error Metrics}
\label{app_ssec:alternate_metrics}

\begin{figure}
    \centering
    \includegraphics{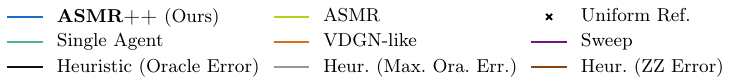}
    \includegraphics{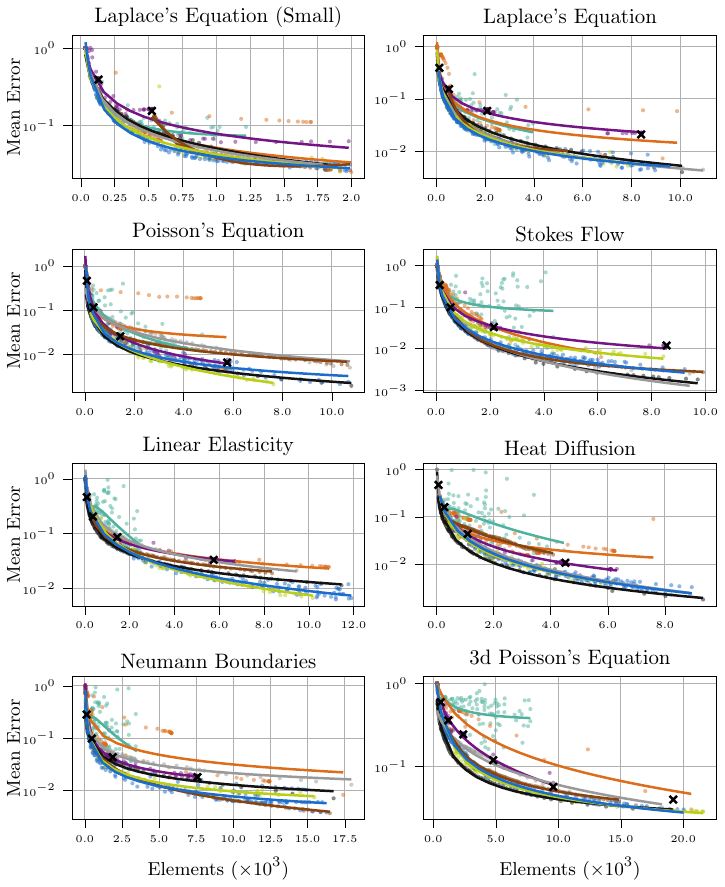}
    
    \caption{
    Pareto plot of normalized mean errors and number of final mesh elements for all tasks.
    \gls{method} performs on par with or better than~\gls{asmr} on all tasks, and both methods significantly outperform all~\gls{rlamr} baselines.
    }
    \label{app_fig:quantitative_additional_metrics_mean}
    \vspace{-0.1cm}
\end{figure}

\begin{figure}
    \centering
    \includegraphics{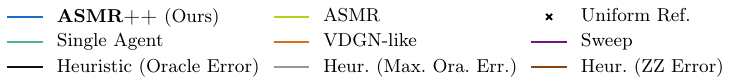}
    \includegraphics{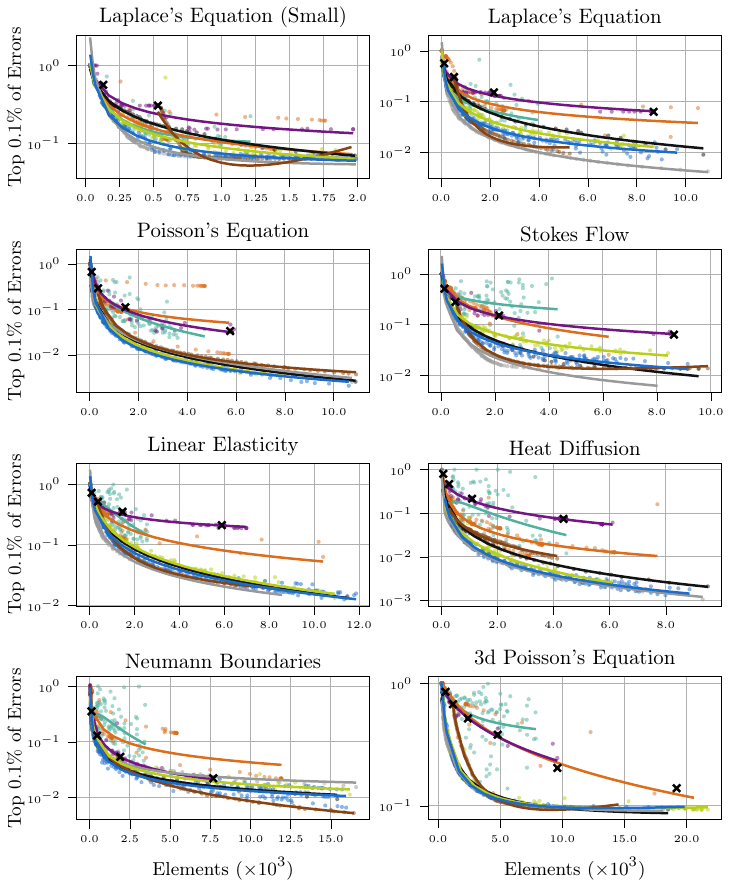}
    
    \caption{
    Pareto plot of normalized maximum element error and number of final mesh elements for all tasks.
    \gls{method} directly minimizes the error of the maximum element during training.
    In contrast,~\gls{asmr} optimizes a scaled variant of the average error.
    Consequently,~\gls{method} surpasses~\gls{asmr} on all tasks.
    Both methods outperform all other~\gls{rlamr} baselines.
    }
    \label{app_fig:quantitative_additional_metrics_top}
    \vspace{-0.1cm}
\end{figure}

In addition to the squared error evaluated in Section~\ref{sec:results}, we present results for a mean and maximum mesh error metric.
We approximate the maximum error as the average of the Top $0.1\,\%$ of errors of all integration points $p_{\Omega^*_m}$ to make it robust to outliers.
We normalize both metrics by the error of the initial mesh for each~\gls{pde}.

Figure~\ref{app_fig:quantitative_additional_metrics_mean} shows pareto plots for all tasks for a normalized mean mesh error. 
The \textit{Oracle Error Heuristic} outperforms its maximum error counterpart, likely because it selects elements with a high integrated error rather than elements with a high maximum error for refinement, thus targeting areas with a high mean error.
\gls{method} performs on par with or better than~\gls{asmr} on average, even though it explicitly optimizes a decrease in maximum mesh error rather than a decrease in its mean error, likely due to its local optimization objective.
Since the mean mesh error is less sensitive to outliers, more uniform refinements such as those produced by~\textit{Sweep} perform better than on other metrics. 
The mean error directly quantifies the difference between the solution calculated on the fine ground truth mesh and that produced by the different methods. 
For most tasks,~\gls{method} produces meshes that reduce the error between the initial mesh and the ground truth by more than $99\,\%$ with only a few thousand elements.

Figure~\ref{app_fig:quantitative_additional_metrics_top} visualizes pareto plots for all tasks for the Top $0.1\,\%$ mesh error. 
Here, the \textit{Maximum Oracle Error Heuristic} excels when compared to the regular \textit{Oracle Error Heuristic}, likely because it refines elements that have the highest maximum error.
Similarly,~\gls{method} outperforms~\gls{asmr} on all tasks, as it directly minimizes the maximum error instead of a scaled version of the average mesh error.
Methods that produce relatively uniform meshes, such as \textit{Sweep}, perform considerable worse on this approximated maximum error than on the other metrics, likely because uniform refinements produce a lot of elements on mesh areas that do not participate in the maximum mesh error.
Notably, the maximum error for the $3$d Poisson's equation is bound at around $0.1$. 
This is likely the case because we use longest edge bisection for the adaptive refinements, which produces refined meshes with local elements that significantly differ from that of the fine-grained uniform ground truth reference. 
As bounding the maximum prediction error is important in many applications, these uniform meshes can potentially waste a lot of computational resources or lead to worse results than the highly adaptive meshes produced by, e.g.~\gls{method}.
Overall, both metrics are consistent with the results in Figures~\ref{fig:quantitative_laplace} and ~\ref{fig:quantitative_others}.

\section{Inference-Time Generalization}
\label{app_sec:generalization_capabilities_runtime}

\begin{figure}[ht]
    \centering
    \input{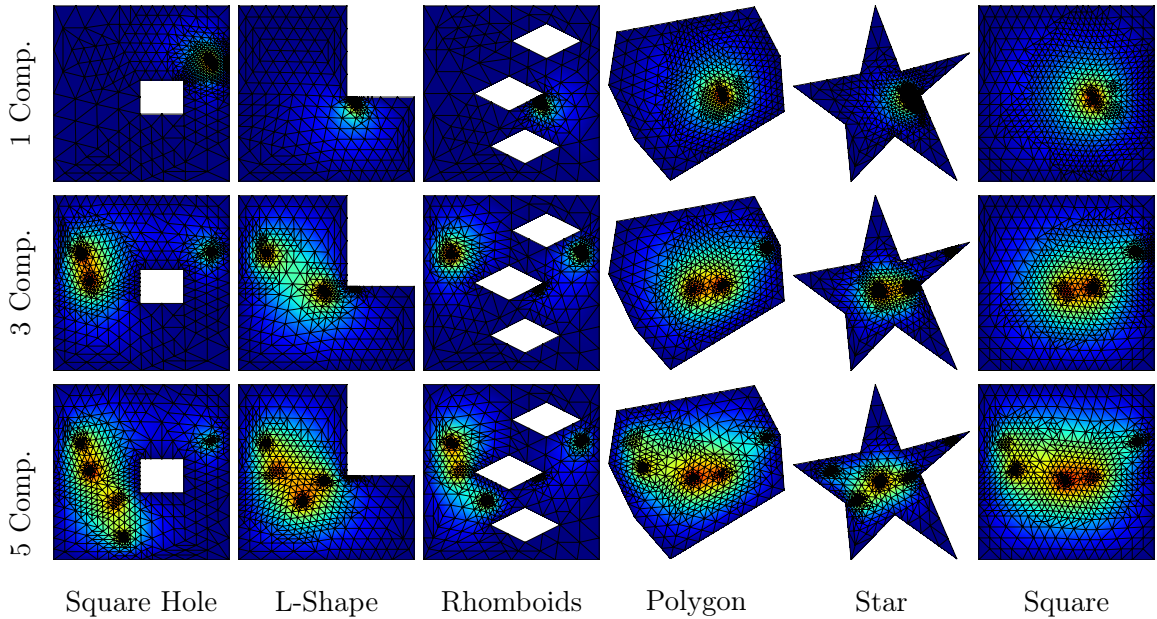}
    \caption{
        \gls{method} refinements for $6$ different domains in $(0,1)^2$ and Poisson's equation with a Gaussian mixture model load function with $1$, $3$ and $5$ components.
        Even though the policy is only trained on L-shaped domains with $3$ components, is easily generalizes to different domains and more components in the applied load function.
    }
    \label{app_fig:poisson_generalizing_1x1}
\end{figure}

\subsection{Same-size Generalization Capabilities}
\label{app_ssec:poisson_generalizing_1x1}
Figure \ref{app_fig:poisson_generalizing_1x1} visualizes final~\gls{method} meshes on Poisson's equation on different $1\times1$ domains and load functions that are not seen during training.
We experiment on the $5$ different domain classes that are used for the different tasks, plus a rectangular $\Omega=(0,1)^2$ domain.
For each class, we randomly sample $3$ domains and apply a Gaussian mixture model load function with $1$, $3$, and $5$ components, respectively.
We find that our approach generalizes well to novel domains and load functions, likely due to the strong local inductive bias of the~Swarm~\gls{rl} objective and the \gls{mpn} policy.

\subsection{Domain Size Up-Scaling}
\label{app_ssec:domain_size_generalization}

\begin{figure}[ht]
    \centering

    \begin{minipage}{0.153\textwidth}
            \centering
            \includegraphics[width=\textwidth]{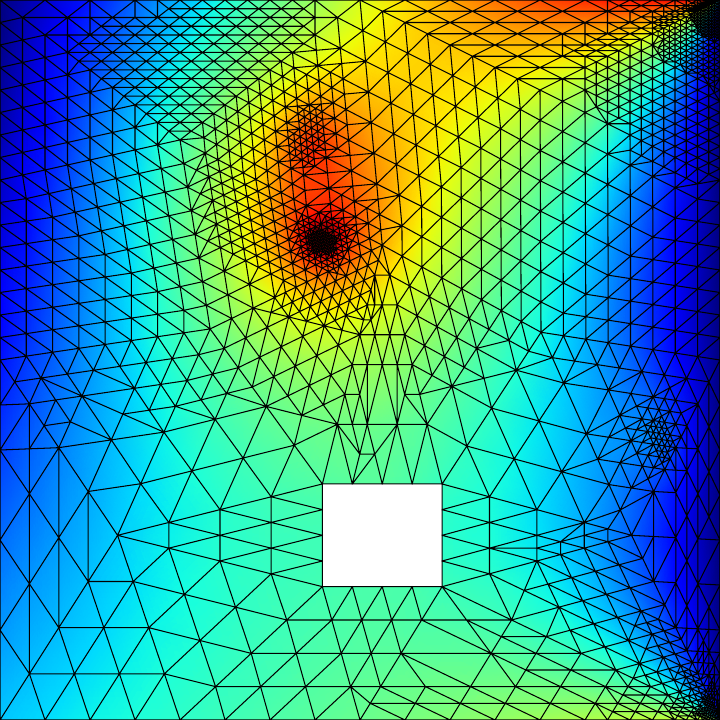}
    \end{minipage}
    \begin{minipage}{0.153\textwidth}
            \centering
            \includegraphics[width=\textwidth]{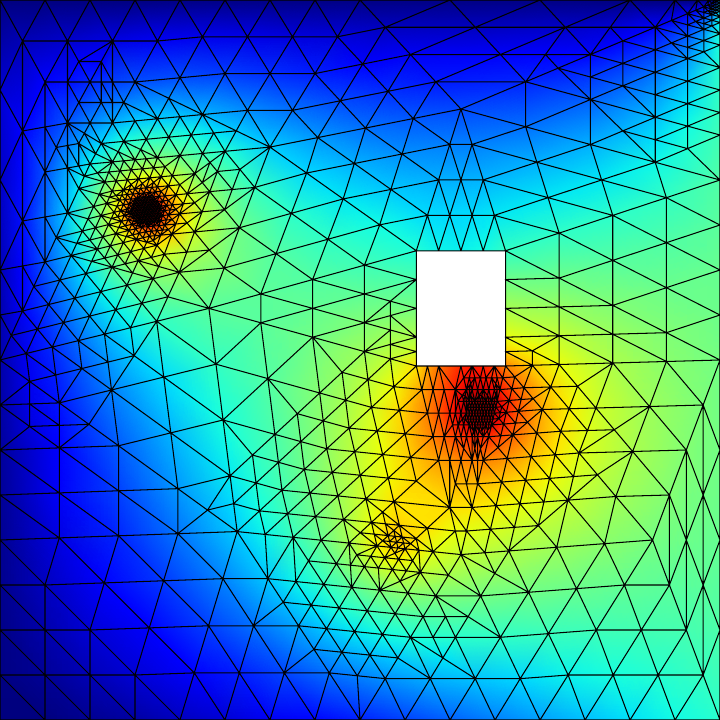}
    \end{minipage}
    \begin{minipage}{0.153\textwidth}
            \centering
            \includegraphics[width=\textwidth]{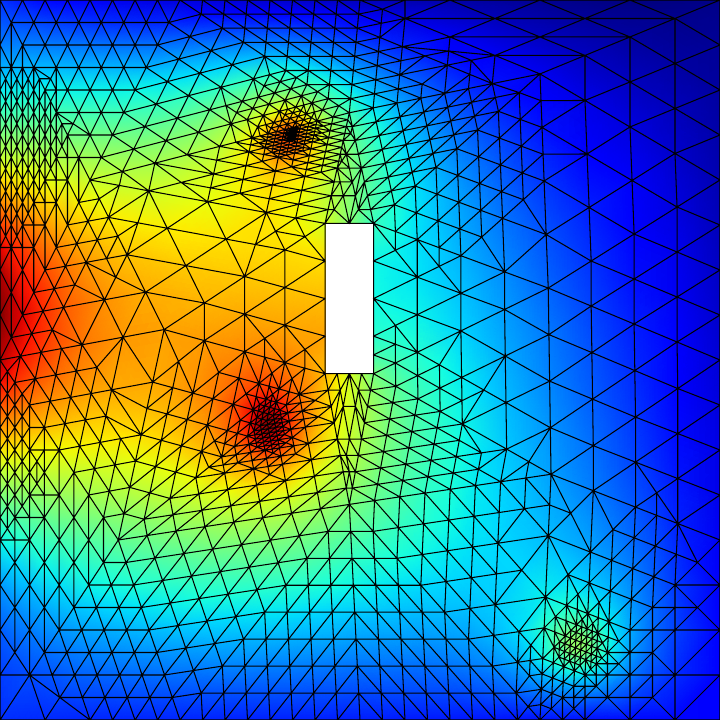}
    \end{minipage}
    \begin{minipage}{0.153\textwidth}
            \centering
            \includegraphics[width=\textwidth]{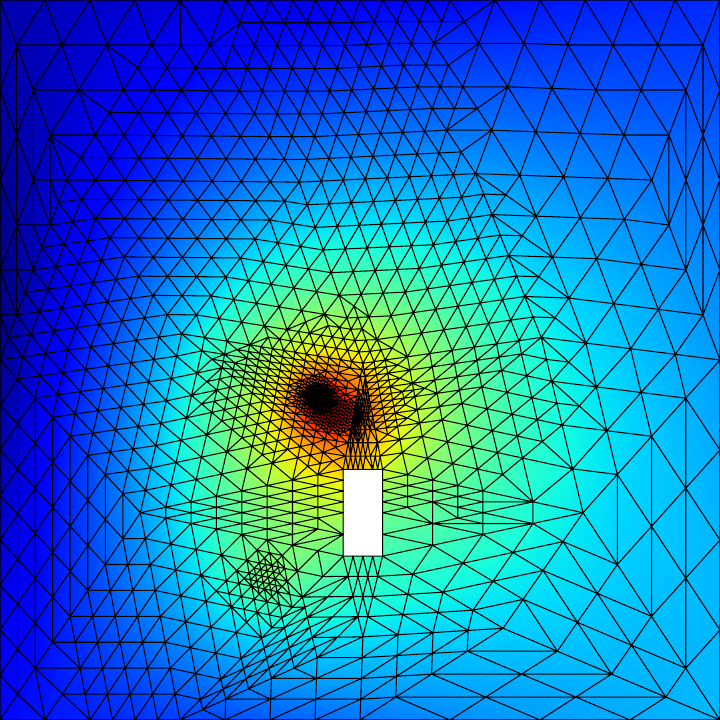}
    \end{minipage}
    \begin{minipage}{0.153\textwidth}
            \centering
            \includegraphics[width=\textwidth]{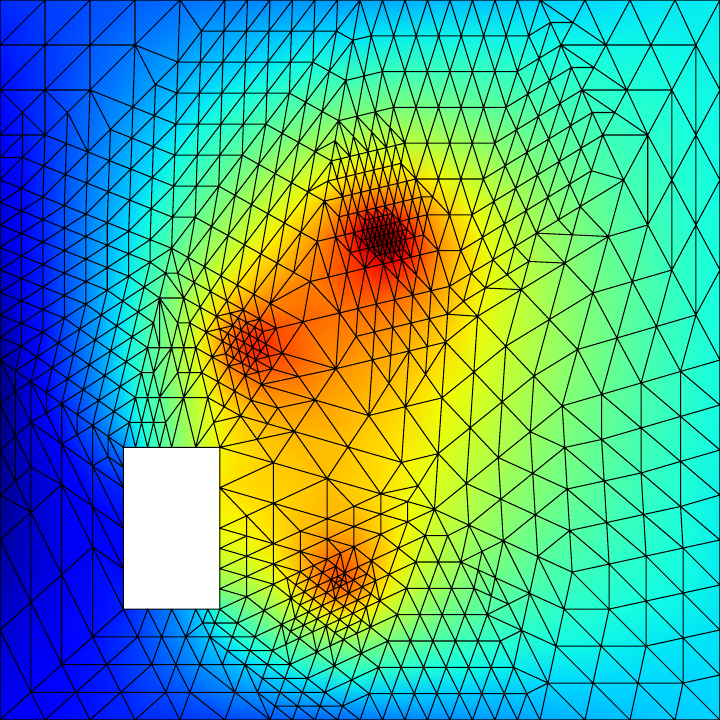}
    \end{minipage}
    \begin{minipage}{0.153\textwidth}
            \centering
            \includegraphics[width=\textwidth]{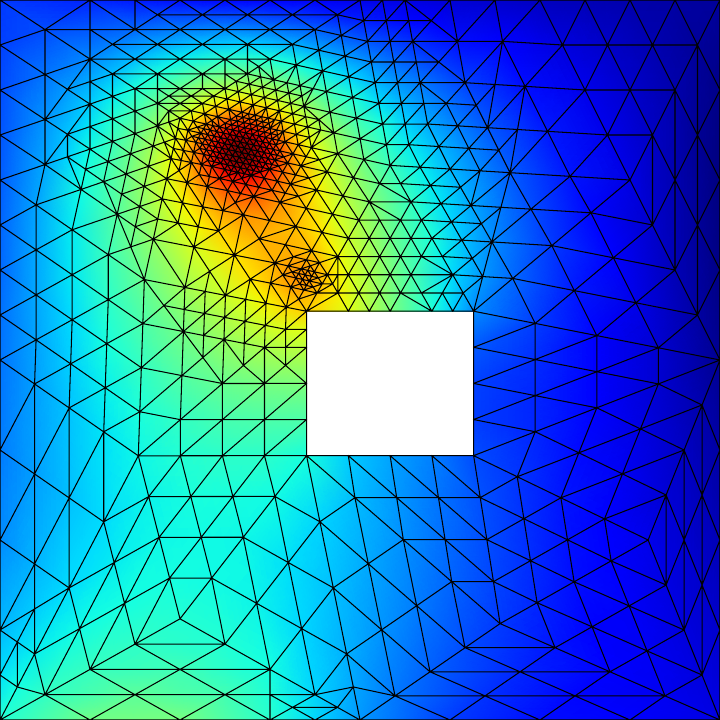}
    \end{minipage}%
    \vspace{0.01\textwidth}%

    \caption{
    Random augmented training \glspl{pde} and \gls{method} refinements for the domain size up-scaling experiments on Poisson's Equation.
    }
    \label{app_fig:poisson_generalization_training}
\end{figure}

We additionally experiment with the abilities of \gls{asmr} to up-scale to larger domains during inference.
This up-scaling is crucial for practical scenarios, as it allows applying policies that are trained on small and comparatively cheap domains to apply to complex, large-scale domains where training would otherwise be too expensive.

Using Poisson's equation as an example, we augment the training \glspl{pde} to cover a larger space of potential simulations and mimic larger mesh segments by altering boundary conditions and load functions. 
We train our policies on the square hole domains of, e.g., Figure~\ref{app_fig:poisson_generalizing_1x1}.
We sample the means of the load function from a centered Gaussian with a standard deviation of $0.2$, allowing components outside the mesh, and additionally apply random Gaussian loads to selected boundary edges to emulate a larger domain outside of the simulated mesh.
As these changes lead to training~\glspl{pde} of varying complexity, we use $10\cdot \hat{\text{err}}(\Omega)$ instead of $\text{err}(\Omega)$ in Equation~\ref{eq:asmr_local_reward}, i.e., we omit the per-domain normalization, and use an unlimited number of training~\glspl{pde} to further increase variety.
These changes only affect the training environments, and do not infer with our training algorithm.
Figure~\ref{app_fig:poisson_generalization_training} visualizes examplary training \glspl{pde} and policy refinements.

\begin{figure}[ht]
    \centering 
    \begin{minipage}{0.62\textwidth}
        \tikzsetnextfilename{app_poisson_generalization_quantitative}
        \input{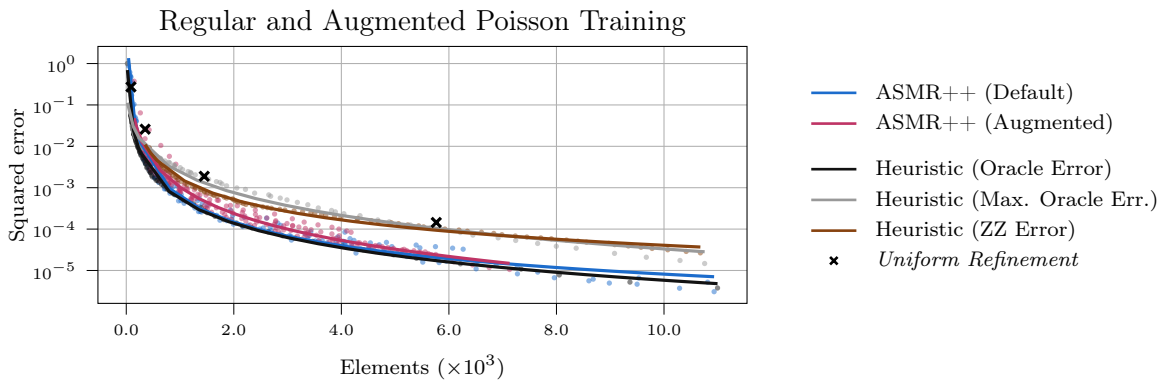}  
    \end{minipage}%
    \begin{minipage}{0.01\textwidth}
    ~
    \end{minipage}%
    \begin{minipage}{0.33\textwidth}
        \tikzsetnextfilename{app_poisson_generalization_quantitative_legend}
        \begin{tikzpicture}
\tikzstyle{every node}'=[font=\scriptsize]\definecolor{black18}{RGB}{18,18,18}
\definecolor{darkgray153}{RGB}{153,153,153}
\definecolor{darkgray176}{RGB}{176,176,176}
\definecolor{indianred19152101}{RGB}{191,52,101}
\definecolor{lightgray204}{RGB}{204,204,204}
\definecolor{royalblue28108204}{RGB}{28,108,204}
\definecolor{saddlebrown1356716}{RGB}{135,67,16}
\begin{axis}[%
                    hide axis,
                    xmin=10,
                    xmax=50,
                    ymin=0,
                    ymax=0.1,
                    legend style={
                        draw=white!15!black,
                        legend cell align=left,
                        legend columns=1,
                        legend style={
                            draw=none,
                            column sep=1ex,
                            line width=1pt
                        }
                    },
                    ]
\addlegendimage{empty legend}
\addlegendentry{\textbf{ASMR++:}}
\addlegendimage{royalblue28108204}
\addlegendentry{Default}
\addlegendimage{indianred19152101}
\addlegendentry{Augmented}

\addlegendimage{empty legend}
\addlegendentry{~}
\addlegendimage{empty legend}
\addlegendentry{\textbf{Heuristic:}}
\addlegendimage{black18}
\addlegendentry{Oracle Error}
\addlegendimage{darkgray153}
\addlegendentry{Max. Oracle Err.}
\addlegendimage{saddlebrown1356716}
\addlegendentry{ZZ Error}
\addlegendimage{only marks, mark=x, black}
\addlegendentry{\textit{Uniform Refinement}}
\end{axis}
\end{tikzpicture}     
    \end{minipage}
    \caption{
        Pareto plot of the normalized squared error for \gls{method} trained on regular and augmented environments evaluated on the regular evaluation environments.
        Augmenting the training environments facilitates inference-time up-scaling at the cost of slightly decreased refinement quality.
    }
    \label{app_fig:poisson_generalization_quantitative}
\end{figure}

Figure~\ref{app_fig:poisson_generalization_quantitative} compares \gls{method} trained on these augmented training~\glspl{pde} with the setup used throughout the paper.
The results show that the augmented training setup leads to a slight decrease in performance relative to~\gls{method} trained under standard conditions.
This decrease in performance is likely caused by the more challenging optimization problem.
Yet, the augmented training yields high-quality refinements.

Once trained, we evaluate the policy on spiral-shaped domains of increasing sizes, ensuring the volume of the initial mesh elements remains constant. 
To align the complexity of the load function with the enlarged mesh sizes, we adapt the number of Gaussian mixture model modes, using more modes for larger domains. 
Specifically, we use $16$ components which are first placed on a uniform grid and then randomly perturbed, allowing for them to be positioned outside of the mesh.
Figure~\ref{app_fig:generalization5_15} shows an exemplary~\gls{method} refinement for a $5\times5$ domain.
Figure~\ref{fig:size20_generalization} shows a final refined mesh on a $20\times20$ domain with $100$ components, including detailed views of specific sections.
Both figures use the same policy with a fixed element penalty.

{
\captionsetup{skip=-0.01cm} %
\begin{figure}[ht!]
    \centering
    \begin{minipage}[b]{0.9\textwidth}
            \centering
            \includegraphics[width=\textwidth]
            {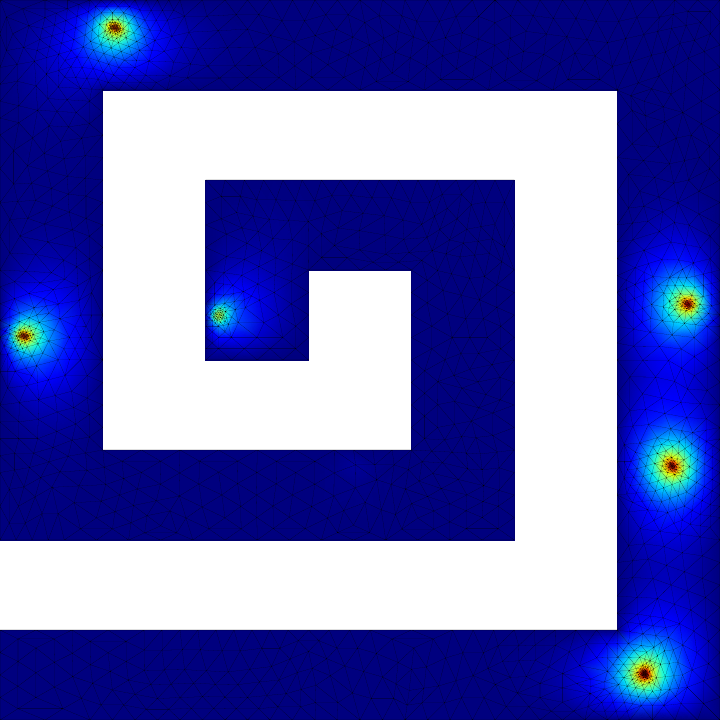}
            \caption*{$5\times5$ Domain}
    \end{minipage}
    \vspace{0.01\textwidth}%
    \caption{
        Visualization of a final~\gls{method} mesh on a $5\times5$ spiral domain with a randomly sampled load function.
        Our approach consistently provides high-quality refinements for both larger domains and more complex load functions during inference.
    }
    \label{app_fig:generalization5_15}
\end{figure}
}

\section{Hyperparameters}
\label{app_sec:hyperparameters}

\subsection{General Hyperparameters}
\label{app_ssec:general_hyperparameters}
The following section lists important hyperparameters used for the training of all methods.
Hyperparameters are kept consistent across all tasks and~\gls{rlamr} methods unless mentioned otherwise.

\subsubsection{Neural Networks.}

We implement all neural networks in PyTorch~\citep{paszke2019pytorch} and optimize them using ADAM~\citep{kingma2014adam} with a learning rate of $3.0$e-$4$ and batch size of $32$ unless mentioned otherwise.
All \glspl{mlp} use $2$ hidden layers and a latent dimension of $64$.
Each \gls{mpn} consists of $2$ message passing steps, where each update function is represented as an \gls{mlp} with \textit{LeakyReLU} activation functions. 
We apply Layer Normalization~\citep{ba2016layer} and Residual Connections~\citep{he2016deep} independently after each node and edge feature update, and use Edge Dropout~\citep{rong2019dropedge} of $0.1$ during training.
The edge feature aggregations~$\bigoplus$ are mean aggregations.
We use separate architectures for the policy and the value function, i.e., we do not share weights between the policy and value function. 
The policy and value function heads are \glspl{mlp} with \textit{tanh} activation functions acting on the final latent node features of the \gls{mpn}.

\subsubsection[PPO.]{\gls{ppo}.}

For~\gls{ppo}, we follow previous work~\citep{andrychowicz2021what} to select important hyperparameters and code-level optimizations.
We train each \gls{ppo} policy for a total of $400$ iterations, except for the $3$d Poisson task, where we only use $200$ iterations.
In each iteration, the algorithm samples $256$ environment transitions and performs $5$ epochs of optimization.
The value function loss is multiplied with a factor of $0.5$.
We clip the gradient norm to $0.5$ and choose clip ranges of $0.2$ for the policy and value function.
We normalize the observations with a running mean and standard deviation.
Advantages are estimated via Generalized Advantage Estimate~\citep{schulman2015high} using $\lambda=0.95$.
We compute an agent's advantage by subtracting the agent-wise value estimates from the return in Equation~\ref{eq:half_half_return}.

\subsubsection[DQN.]{\gls{dqn}.}

For \gls{dqn}-based approaches, we draw $500$ initial samples with a random policy for the replay buffer and then train for $24*400=9600$ steps, where each step consists of executing and storing an environment transition and then drawing a random mini-batch with replacement from the buffer for a single gradient update.
We experimented with more training steps in preliminary experiments, finding that they do no significantly improve performance or stabilize training, but may lead to increased memory footprint and longer runtimes.
We update the target networks using Polyak averaging at a rate of $0.99$ per step.
We select training actions using a Boltzmann distribution over the predicted Q-values per agent, where we linearly decrease the temperature of the distribution from $1$ to $0.01$ in the first half of training. 
This action selection strategy favors more correlated actions when compared to an epsilon greedy action sampling, which empirically stabilizes the training for our problem setting of iterative mesh refinement.
Further, we follow previous work~\citep{hessel2018rainbow} and combine a number of common improvements for~\glspl{dqn}, namely double Q-learning~\citep{van2016deep}, dueling Q-networks~\citep{wang2016dueling} and prioritized experience replay~\citep{schaul2015prioritized}.

\subsection{Baseline-Specific Parameters}
\label{app_ssec:baseline_hyperparameters}

\subsubsection{Single Agent.}
We use a maximum refinement depth of $10$ refinements per element to avoid numerical instabilities during simulation, skipping actions that try to refine elements that have been refined too often.
We consider environment sequences of up to $T=400$ steps since the method marks only one element at a time, but train a separate policy for every value of $T$.

\subsubsection{Sweep.}
The \textit{Sweep} baseline moves a single agent to a random mesh element after each training step. The agent then decides if this element should be refined or not.
We follow the proposed hyperparameters and have training rollouts of $200$ steps~\citep{foucart2023deep}.
Since \textit{Sweep} uses a local agent living on a single mesh element, we use an~\gls{mlp} instead of the~\gls{mpn} for both the policy and value function.
Correspondingly, we adapt the input features, using our regular node features and the global resource budget proposed by the authors.
We additionally add aggregated neighborhood information in the form of a mean solution and area of the element's neighbors and the average distance to them.
The global budget is controlled via a maximum number of elements $N_{\text{max}}$, allowing to train policies that produce refinements of different granularity.
We increase the number of environment transitions per \gls{ppo} step to $512$, and the number of \gls{dqn} steps to $96*400=38400$.
This change is intended to compensate for decreased number of refined elements of each environment step while roughly equating for the computation time of the other methods.

\subsubsection{VDGN-like.}
We use a learning rate of $1.0$e-$5$ instead of $3.0$e-$4$ for the \gls{dqn} variant of the \gls{vdgn}-like baseline to stabilize its training.

\subsubsection{ASMR.}

We use the reward of Equation~\ref{eq:volume_reward}, an infinite horizon~\gls{rl} setting with $\gamma=0.99$ and omit the edge dropout in the observation graph for~\gls{asmr}, and additionally use an agent mapping without the normalization factor $\frac{|\Omega^t|}{|\Omega^{t+1}|}$ in Equation~\ref{eq:agent_mapping}.

\subsection{Mesh Resolution Parameters}
\label{app_ssec:refinement_hyperparameters}
All~\gls{amr} methods in this work allow for some control over the desired refinement level of the final mesh.
\gls{method}, \gls{asmr} and \gls{vdgn} use an element penalty $\alpha$ that trades off the cost of adding new elements with the benefit of a refinement.
\textit{Sweep} considers an element budget $N_{\text{max}}$, attempting to minimize the error of the mesh while staying within this budget.
\textit{Single Agent} varies the number of rollout steps $T$, refining once for every step.

Both~\gls{method} and~\gls{vdgn} train on a range of $\alpha$ values and condition the policy on it, allowing for adaptive mesh resolutions during inference.
Here, we evaluate $20$ penalties that are log-uniformly sampled over the training penalty range to cover different final mesh resolutions.
As the other methods do not directly support an adaptive penalty during inference, we train $10$ separate policies with $10$ repetitions each for different refinement parameters.
The \textit{Oracle}, \textit{Maximum Oracle}, and \gls{zz} \textit{Heuristics} refine based on estimated errors per mesh element. 
Here, we evaluate $100$ values for the error threshold $\theta$, which specifies which elements to refine based on the ratio between their error and the maximum element error.
Tables~\ref{app_tab:resolutions_rl} and~\ref{app_tab:resolutions_heuristic} list the minimum and maximum mesh resolution parameters for all tasks and \gls{rlamr} methods and heuristics, respectively.
\begin{table}[ht!]
    \centering
    \resizebox{\textwidth}{!}{
    \begin{tabular}{l|lllll}
    \toprule
    & \textbf{\gls{method}} & \textbf{\gls{asmr}} & \textbf{\gls{vdgn}-like} & \textbf{\textit{Sweep}} & \textbf{\textit{Single Agent}} \\
    \textbf{Parameter} & $\alpha$ &$\alpha$ & $\alpha$ & $N_{\text{max}}$ & $T$ \\
    \midrule
    \textbf{Laplace (Small)} & $[0.001, 0.1] $ & $[0.01, 0.3]$ & $[1\text{e}-5, 5\text{e}-2]$ & $[50, 1000]$ & $[5, 100]$ \\
    \textbf{Laplace} & $[0.001, 0.03]$ & $[0.01, 0.5]$ & $[1\text{e}-5, 1\text{e}-2]$ & $[200, 3000]$ & $[25, 400]$ \\
    \textbf{Poisson} & $[0.0001, 0.03]$ & $[0.002, 0.1]$ & $[2\text{e}-5, 5\text{e}-2]$ & $[200, 3000]$ & $[25, 400]$ \\
    \textbf{Stokes Flow} & $[0.0005, 0.05]$ & $[0.015, 0.3]$ & $[3\text{e}-4, 2\text{e}-2]$ & $[150, 2500]$ & $[25, 400]$ \\
    \textbf{Linear Elasticity} & $[0.00015, 0.03]$ & $[0.01, 0.15]$ & $[1\text{e}-5, 1\text{e}-2]$ & $[500, 6000]$ & $[25, 400]$ \\
    \textbf{Heat Diffusion} & $[0.00005, 0.01]$ & $[0.003, 0.3]$ & $[1\text{e}-5, 1\text{e}-2]$ & $[400, 5000]$ & $[25, 400]$ \\
    \textbf{Neumann Boundaries} & $[0.00005, 0.03]$ & $[0.0075, 0.5]$ & $[1\text{e}-5, 1\text{e}-1]$ & $[200, 3000]$ & $[25, 400]$ \\
    \textbf{Poisson $3$d} & $[0.0001, 0.1]$ & $[0.1, 20.0]$ & $[1\text{e}-5, 1\text{e}-1]$ & $[500, 5000]$ & $[25, 400]$ \\
    \bottomrule
    \end{tabular}
    }
    \caption{
        Ranges for the different refinement hyperparameters for all task and \gls{rlamr} methods.
        \gls{method}, \gls{asmr} and \gls{vdgn} apply a penalty $\alpha$ for each added element. 
        \textit{Sweep} uses an element budget $N_{\text{max}}$. 
        \textit{Single Agent} varies the number of rollout steps $T$.
    }
    \label{app_tab:resolutions_rl}
\end{table}

\begin{table}[ht!]
    \centering
    \resizebox{\textwidth}{!}{
    \begin{tabular}{l|lll}
        \toprule
        & \textbf{\textit{Oracle Error}} & \textbf{\textit{Max. Oracle Err.}} & \textbf{\gls{zz}} \\
        \textbf{Parameter} & $\theta$ & $\theta$ & $\theta$ \\
        \midrule
        \textbf{Laplace (Small)} & $[0.22, 1.0]$ & $[0.16, 1.0]$ & $[0.002, 1.0]$ \\
        \textbf{Laplace} & $[0.20, 1.0]$ & $[0.12, 1.0]$ & $[0.001, 1.0]$ \\
        \textbf{Poisson} & $[0.12, 1.0]$ & $[0.15, 1.0]$ & $[0.002, 1.0]$ \\
        \textbf{Stokes Flow} & $[0.12, 1.0]$ & $[0.04, 1.0]$ & $[0.001, 1.0]$ \\
        \textbf{Linear Elasticity} & $[0.06, 1.0]$ & $[0.02, 1.0]$ & $[0.001, 1.0]$ \\
        \textbf{Heat Diffusion} & $[0.04, 1.0]$ & $[0.02, 1.0]$ & $[0.001, 1.0]$ \\
        \textbf{Neumann Boundaries} & $[0.14, 1.0]$ & $[0.24, 1.0]$ & $[0.001, 1.0]$ \\
        \textbf{Poisson $3$d} & $[0.02, 1.0]$ & $[0.02, 1.0]$ & $[0.001, 1.0]$ \\
        \bottomrule
    \end{tabular}
    }
    \caption{
        Ranges for the threshold $\theta$ of the error-based refinement strategy of the heuristic baselines.
        All ranges are chosen to facilitate direct comparison to the \gls{rlamr} methods in the quantitative evaluation, i.e., to produce meshes with a comparable number of elements.
    }
    \label{app_tab:resolutions_heuristic}
\end{table}

\section{Baseline Visualizations}
\label{app_sec:baseline_visualizations}

\subsection{RL-AMR visualizations}
\begin{figure}[ht]
    \centering
    \input{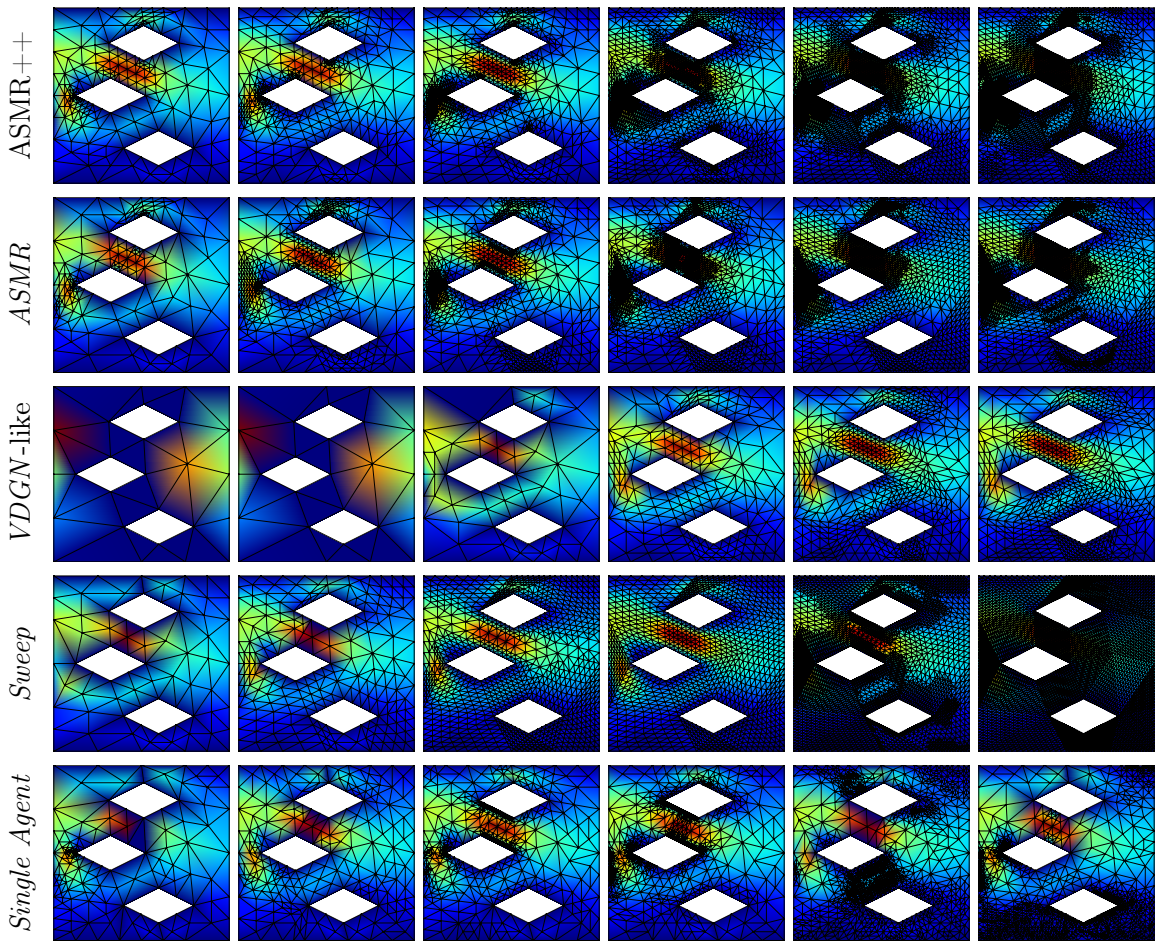}
    \caption{
    Comparison of the different~\gls{rlamr} methods for the Stokes Flow task for the same \gls{pde} and different target mesh resolutions.
    The target mesh resolution increases from the left to the right for each row.
    \gls{method} and the~\gls{vdgn}-like baseline use the same model queried with different $\alpha$ values for each row. 
    The other methods use a different policy for each produced mesh.
    }
    \label{app_fig:visualization_matrix_stokes_flow_all_rl_amr}
\end{figure}

We visualize the final meshes and corresponding~\gls{pde} solutions of all \gls{rlamr} methods on the Stokes Flow task for $6$ different refinement granularities on the same randomly selected system of equations.
For~\gls{method} and~\gls{vdgn}, which condition the policy on the element penalty parameter, we use the same trained model for the $6$ mesh refinement granularities.
Figure~\ref{app_fig:visualization_matrix_stokes_flow_all_rl_amr} visualizes the results.

We find that~\gls{method} provides accurate refinements for different values of~$\alpha$ using a single policy.
The produced meshes are of higher quality than those created by~\gls{asmr}, especially near the corners of the rhomboid obstacles in the domain.
All other~\gls{rl}-\gls{amr} methods are unable to produce consistent meshes.
\textit{Single Agent} tends to over-refine uninteresting regions, likely due to its iterative refinement procedure. 
In contrast, \textit{Sweep} often collapses to uniform or mostly uniform refinements, which may be a by-product of the difference in training on individual elements and sweeping over all elements at once during inference.
\gls{vdgn} produces high-quality refinements in some cases, but is inconsistent across seeds and in some cases fails to optimize for different element penalties with the same model, which could be because its value decomposition objective may scale poorly to large numbers of agents.
These qualitative findings are consistent with the error measurements of Figure~\ref{fig:quantitative_others}.

\subsection{Heuristic visualizations}
\begin{figure}[ht]
    \centering
    \input{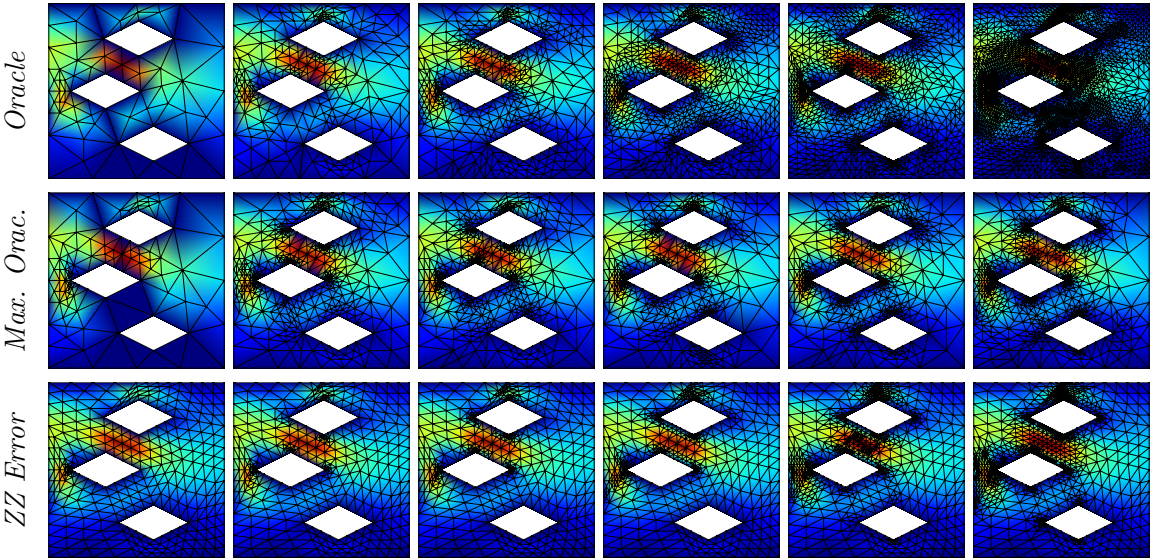}
    \caption{
    Comparison of the different \textit{Heuristic} methods for the same Stokes Flow task \gls{pde} and different error thresholds $\theta$.
    The target mesh resolution increases from the left to the right for each row.
    }
    \label{app_fig:visualization_matrix_stokes_flow_all_heuristics}
\end{figure}

We visualize exemplary refinements for the error estimation-based threshold \textit{Heuristics} in Figure~\ref{app_fig:visualization_matrix_stokes_flow_all_heuristics}.
All \textit{Heuristics} greedily refine the elements with their respective largest error estimates, regardless of the resulting decrease in error.
This behavior is fully local, which may cause issues for global dependencies~\citep{strauss2007partial} and conforming refinements.
The \gls{zz} \textit{Heuristic} uses a smoother error estimate than the oracle heuristics, which leads to potentially sub-optimal but generally more consistent refinements.

\section{Additional ASMR++ Visualizations}
\label{app_sec:asmr_visualizations}

\begin{figure}[ht]
    \centering
    \input{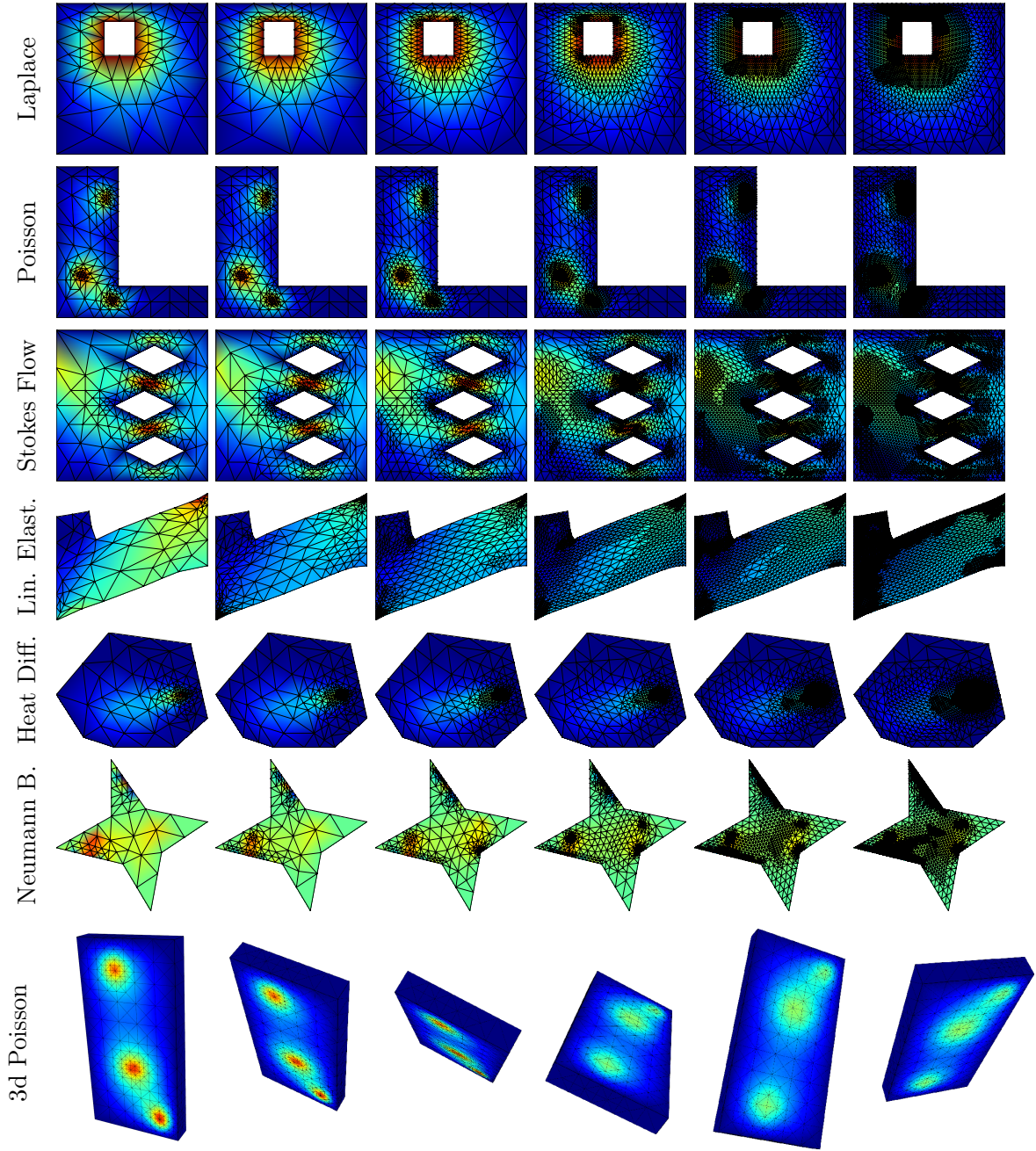}
    \caption{
        \gls{method} meshes for the different tasks and varying element penalties $\alpha$. 
        Each row uses the same policy conditioned on a range of low (left) to high (right) $\alpha$ values for the $2$-dimensional tasks.
        For the $3$-dimensional Poisson task, we use the same \gls{pde} and $\alpha$ value, and show different camera angles.
    }
    \label{app_fig:visualization_matrix_all_tasks_asmr}
\end{figure}

We visualize meshes created by~\gls{method} policies for all considered tasks.
For the $2$-dimension tasks, we follow Section~\ref{app_sec:baseline_visualizations} and use a fixed policy to create refinements with $6$ different granularities for a random~\gls{pde} for each task, only varying the element penalty $\alpha$.
For the $3$-dimensional variant of Poisson's equation, we instead visualize the same random~\gls{pde} and $\alpha$ value for $6$ different camera angles.
Figure~\ref{app_fig:visualization_matrix_all_tasks_asmr} displays the final refined meshes, as well as the solutions of the~\gls{pde} on the meshes.
On all tasks, \gls{method} provides highly accurate refinements across target mesh refinement levels.

\end{appendices}

\clearpage

\bibliography{bibliography}%

\end{document}